\PassOptionsToPackage{hyphens}{url}
\documentclass[r1]{rax}  
\setcitestyle{sort}

\usepackage{amsmath,amsfonts,bm}

\def\eqref#1{equation~\ref{#1}}

\def\1{\bm{1}}

\DeclareMathAlphabet{\mathsfit}{\encodingdefault}{\sfdefault}{m}{sl}
\SetMathAlphabet{\mathsfit}{bold}{\encodingdefault}{\sfdefault}{bx}{n}

\usepackage{amssymb}
\usepackage{tabularx}
\usepackage{algorithm}
\usepackage{algorithmic}
\usepackage{xspace}
\usepackage{xcolor}

\newcommand{\rax}{RAX\xspace}

\newenvironment{raxquote}{%
    \begin{list}{}{%
        \setlength{\leftmargin}{1.25em}\setlength{\rightmargin}{1.25em}%
        \setlength{\topsep}{0.9em}\setlength{\partopsep}{0pt}\setlength{\parsep}{0pt}}%
    \item\relax\itshape
}{%
    \end{list}\vspace{0.15em}
}

\title{Scaling Online Exploration for Robot Agents}

\author[1,\ast]{Runhan Huang}
\author[1,\ast,\dagger]{Xiangcheng Zhang}
\author[1,\ddagger]{Haoyu Chen}
\author[2,\ddagger]{Yitong Li}
\author[3,\ddagger]{Yen-Jen Wang}
\author[3,\ddagger]{Bohan Lyu}
\author[5]{Haotian Ye}
\author[3]{Chuye Hong}
\author[2]{Yixuan Wang}
\author[1]{Chloe Su}
\author[4]{Haoyu Xiong}
\author[3]{Yangzhen Wu}
\author[2]{Enze Xie}
\author[4]{Travis Manderson}
\author[5]{Stefano Ermon}
\author[3]{Pieter Abbeel}
\author[4]{Antonio Torralba}
\author[3]{Dawn Song}
\author[4]{Weirui Ye}
\author[1]{Yilun Du}
\affiliation[1]{Harvard University}
\affiliation[2]{NVIDIA}
\affiliation[3]{UC Berkeley}
\affiliation[4]{MIT}
\affiliation[5]{Stanford University}
\contribution[\ast]{Equal contribution}
\contribution[\dagger]{Project lead}
\contribution[\ddagger]{Core contributor}
\logo[1.16cm]{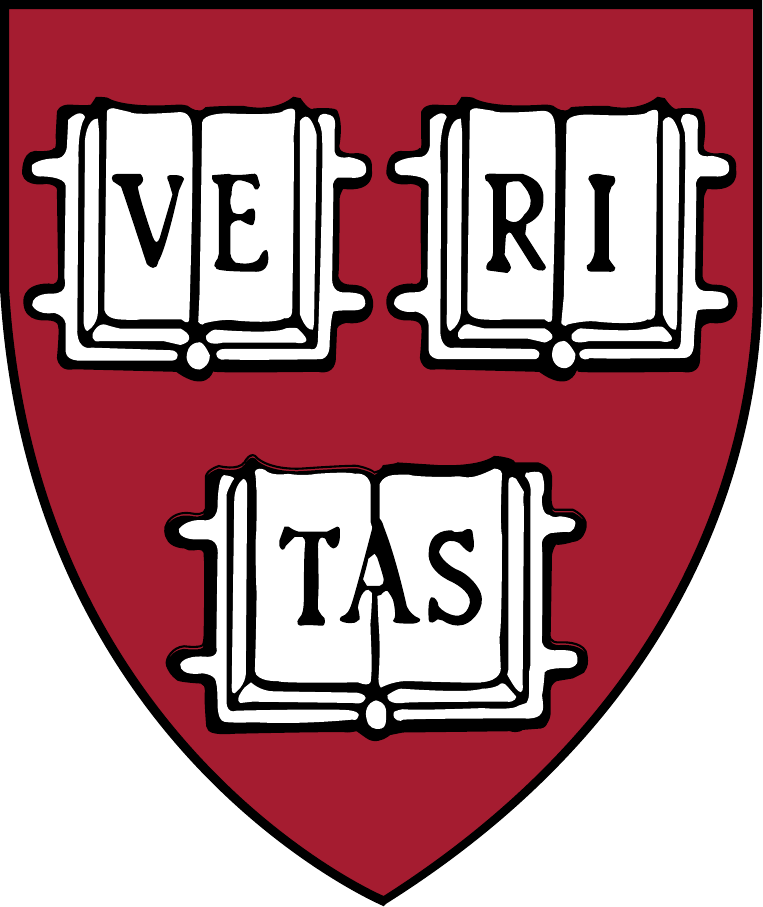}
\logo[0.62cm]{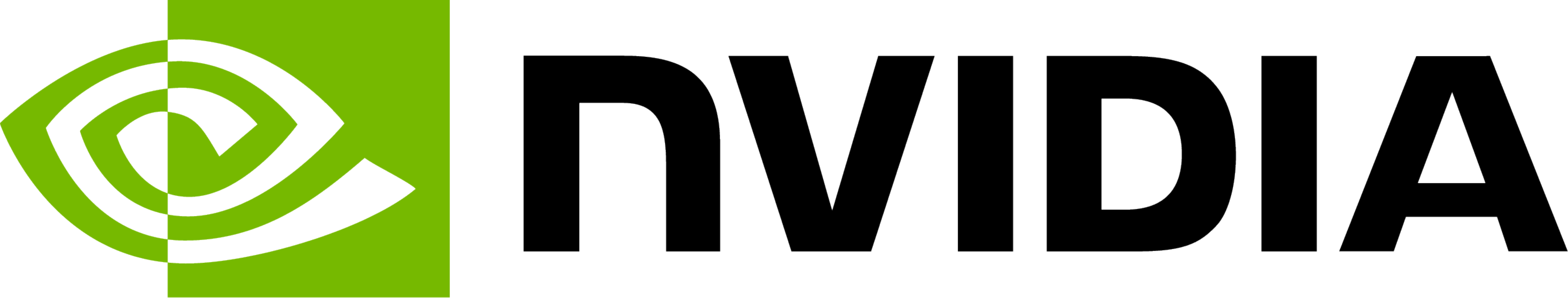}
\logo[1.08cm]{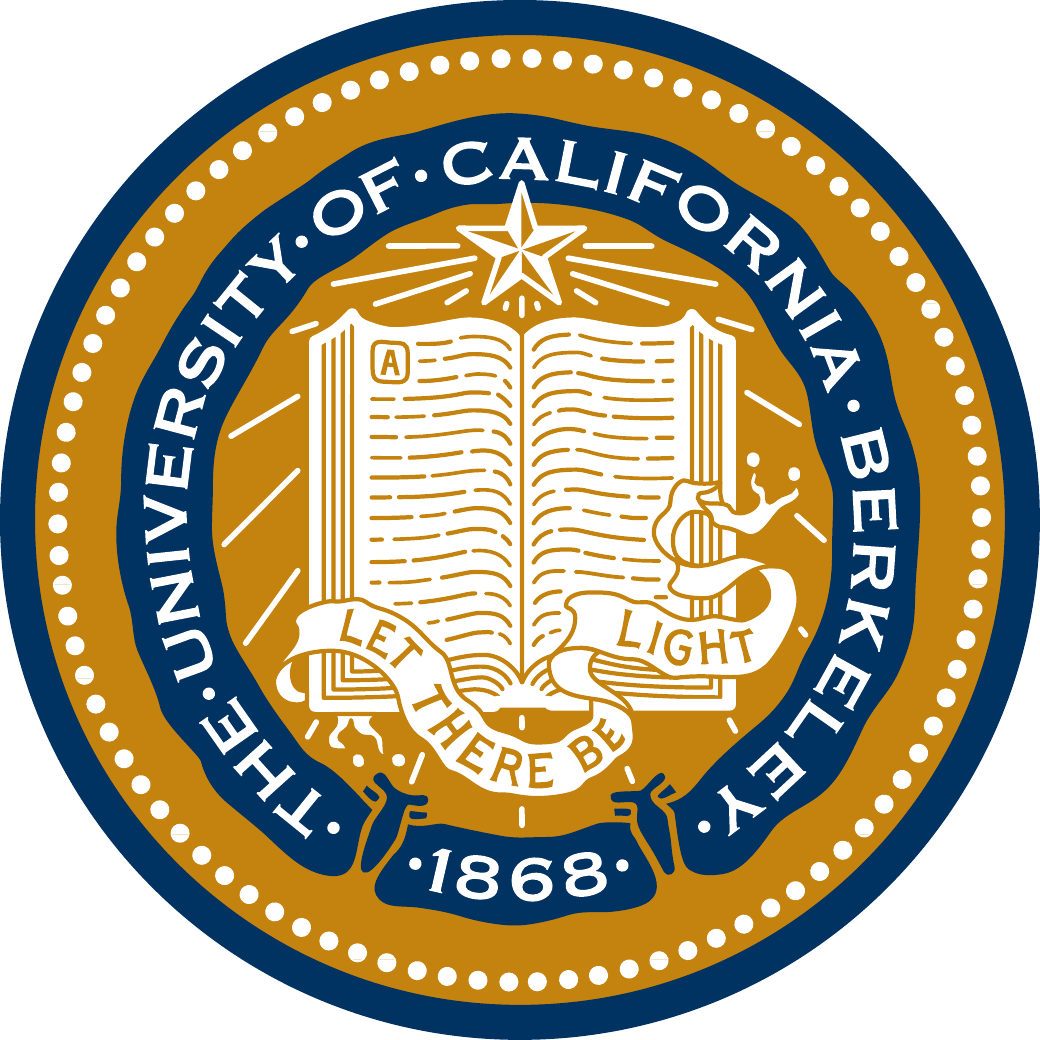}
\logo[0.66cm]{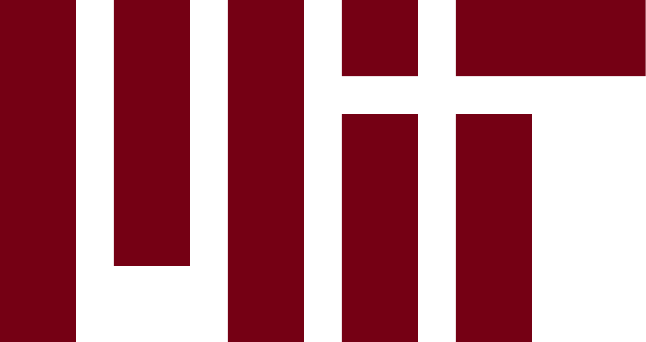}
\logo[1.12cm]{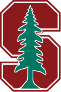}

\abstract{%
Robot learning has advanced significantly by scaling demonstration datasets and model capacity; however, deploying a new robot embodiment or task typically necessitates a dedicated teleoperation corpus. Furthermore, resulting policies remain bounded by their imitation distributions, limiting generalization and continual self-improvement. We introduce \textbf{\rax (Robot Agent eXploration)}, an open-source framework for robot agents centered on \textbf{scaling exploration}: enabling agents to improve by acquiring online experience and synthesizing executable programs. In \rax, a general-purpose coding agent generates robot control code directly, operating without action demonstration data, hand-engineered action primitives, or value-function guidance. Through online exploration, the agent identifies decision sequences that achieve task goals and compiles them into modular programs encapsulating the full perception-planning-control loop for fast inference. Using \rax, we present a systematic multi-embodiment study on scaling online exploration across fine-grained precision and long-horizon manipulation tasks using single-arm, bimanual, dexterous, mobile, and humanoid systems in simulation and on real robots. Generalization performance on unseen environments improves consistently as exploration scales. In simulation, \rax outperforms top baseline models by $+37.4\%$ on average. On real robots, \rax autonomously learns memory-dependent, articulated, high-precision, deformable, and long-horizon mobile manipulation tasks without human demonstrations, continually improving performance by composing verified skills from prior exploration sessions. As exploration scales, we observe emergent system capabilities including active perception, error recovery, safety guardrails, and human-robot agent collaboration. We open-source \rax as a generalist robot-agent platform that allows new robot embodiments to be integrated solely via API interfaces, bypassing teleoperated data collection and policy training entirely. More details are available on our project page at \href{https://robot-agent-exploration.com/}{\texttt{robot-agent-exploration.com}}.
}

\project{\href{https://robot-agent-exploration.com/}{\texttt{https://robot-agent-exploration.com}}}

\teaser{%
    {\centering
    \includegraphics[width=\linewidth]{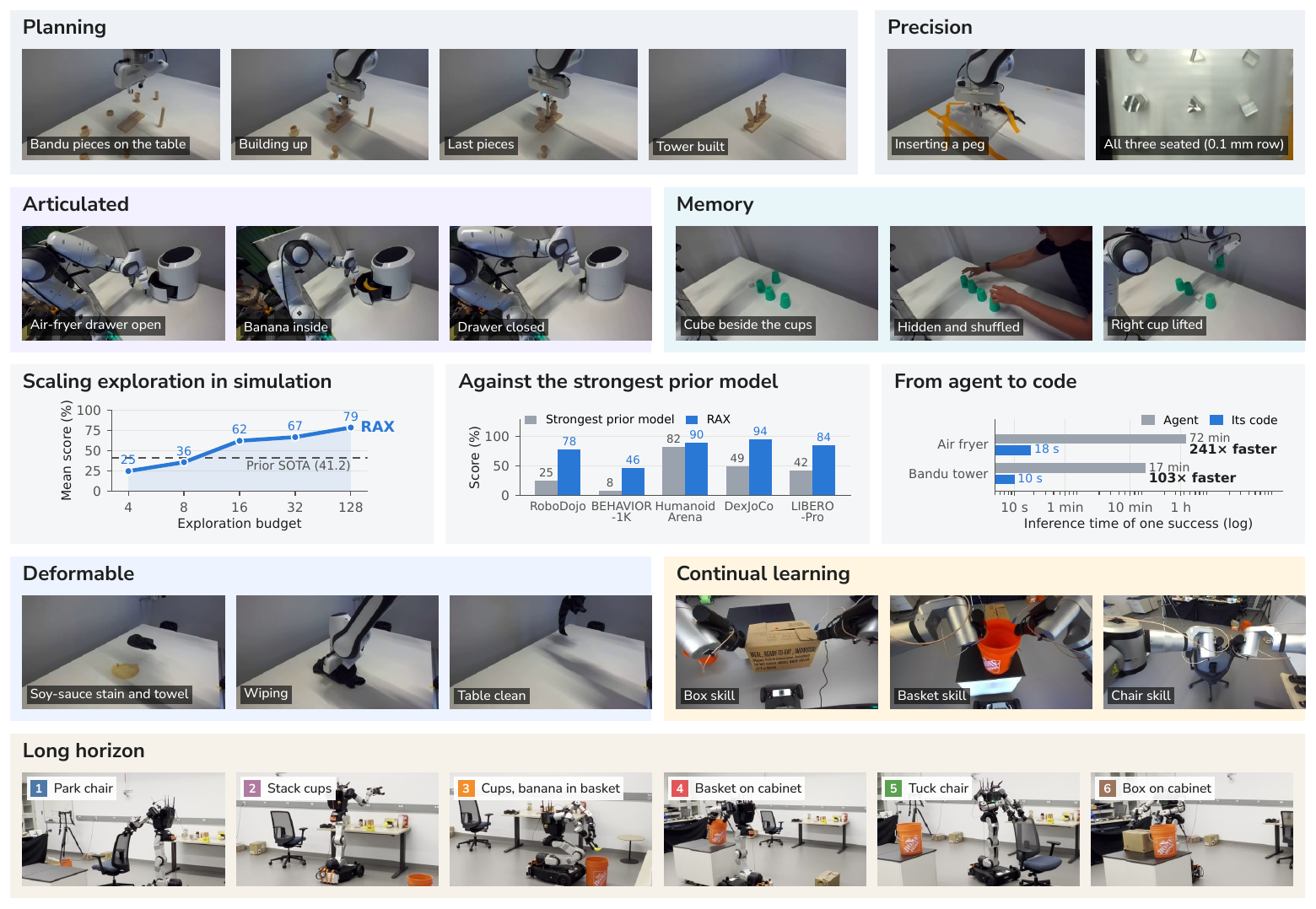}\par}
    {\captionsetup{type=figure,skip=4pt}\caption{\textbf{Overview of \rax.} By scaling online exploration, a coding agent masters
    robot tasks across embodiments without any action demonstrations, from precise insertion and
    articulated, deformable, and memory tasks to long-horizon mobile manipulation, and surpasses
    the state of the art on unseen scenes.}\label{fig:teaser}}}

\begin{document}

\maketitle
\clearpage
\raxtableofcontents
\clearpage

\section{Introduction}
\label{sec:intro}

\begin{raxquote}
    \textquotedblleft We want AI agents that can discover like we can, not which contain what
    we have discovered. Building in our discoveries only makes it harder to see how the
    discovering process can be done.\textquotedblright\par
    \vspace{0.5em}
    \raggedleft --- \textnormal{Richard S. Sutton, \emph{The Bitter Lesson}~\citeyearpar{sutton2019bitter}}\par
\end{raxquote}
\vspace{-0.6em}

Robot learning has made substantial progress by scaling demonstration data and model capacity with imitation learning, aiming to replicate the training recipe that transformed language modeling~\citep{kaplan2020scaling}. Vision-language-action (VLA) models and World-Action models (WAMs) fine-tune pretrained VLMs and video generation models, respectively, using large teleoperation corpora~\citep{brohan2023rt,kim2024openvla,black2024pi_0}. However, deploying a new robot or task still requires a dedicated teleoperation corpus of its own~\citep{o2024open,team2024octo,black2024pi_0,intelligence2025pi_,bjorck2025gr00t,liu2026g0,wang2026openwam,yuan2026fast,kim2026cosmos,bi2026motus}. Furthermore, the knowledge of such policy models is inherently bounded by the demonstrations they imitate. Consider a mobile robot tasked with loading a car in an unfamiliar house: while opening doors, grasping bags, and placing objects may each be familiar skills, the robot must adapt to novel scenes, visual appearances, and task compositions while discovering new approaches. This adaptation and discovery process is absent during imitation learning on static teleoperation data. Consequently, these policies struggle to generalize and self-improve, even when provided with environment interactions and test-time compute. This limitation motivates a central question: \emph{can robots self-improve and learn to discover new skills through autonomous online experience and active exploration?}

Humans learn physical skills by attempting a task, observing the outcome, diagnosing failures, and practicing until their behavior becomes reliable. LLM agents exhibit a similar pattern in software domains: they explore unfamiliar systems by executing programs, inspecting outputs, and revising their work~\citep{yang2024swe}. Instead of training policies exclusively on teleoperated demonstration data, we let \textit{agents program robots directly}~\citep{1456949} without hand-designed action primitives. The agent explores online to discover decisions that drive task success, then automates these decision-making principles into modular robot programs that integrate perception, planning, and control. It continuously debugs and tests these programs through online execution. These programs encode the agent's underlying principles for \textit{how to} generate successful actions based on environment perception, rather than merely mimicking the actions themselves, thereby enabling more robust generalization. Furthermore, the agent can compose, modify, and update these programs when confronting increasingly complex scenes and tasks, continually improving through online experience.

We introduce \textbf{\rax (Robot Agent eXploration)}, a cognitive architecture for robot agents. Its core organizing principle is \textbf{scaling exploration}: robot agents should improve by gathering their own online experience and optimizing their programs. In \rax, a general-purpose coding agent receives a natural-language instruction, at most a few goal images or visual demonstrations, and a budget of exploration episodes. It interacts directly through the robot's perception and control interfaces without requiring human teleoperation or value function guidance. The underlying language model parameters remain fixed and no policies are trained; instead, learning updates a persistent workspace of programs, notes, and verified skills. The agent converts successful interactions into autonomous code programs and stores reusable knowledge for subsequent episodes and tasks.

As exploration scales, we observe several emergent system capabilities: the agent constructs task-relevant 3D perception models, plans actions via pose search in 3D joint space, generates fully automated programs with closed-loop visual and force feedback, and adapts its programs when encountering novel task conditions. Furthermore, we observe agentic behaviors on real robots, including active perception, error recovery, safety guardrails, and active human-robot interaction. For example, the robot executes end-effector contact probing for precise localization when depth measurements are inaccurate. In mobile manipulation experiments, following an initial collision with a cabinet, the agent automatically instantiates a safety check to verify workspace clearance before every base rotation. To test code robustness, the agent actively requests the human operator to vary object layouts, systematically testing the spatial generalization of the autonomous program.

In \rax, we equip the agent with rich multimodal sensing by exposing metric depth and force signals through the robot interface. A unified architecture supports tabletop bimanual arms, dexterous hands, mobile household robots, and humanoids through interfaces to each robot's sensors, controllers, and kinematic models (\Cref{sec:system}). Unlike standard VLAs or WAMs operating on RGB images alone, the agent autonomously leverages these rich sensing signals to construct 3D point clouds and fit object models directly from perception. It then searches for actions to manipulate these 3D models while accounting for their unique shapes and affordances. Force feedback further enables the agent to resolve spatial uncertainty through contact and enable close-loop control on force signals in insertion tasks. Such multimodal world modeling have been largely absent from generalist policy models, yet naturally align with human cognitive processes.

To enable high-frequency closed-loop control, the agent compiles its decision-making process into executable Python programs that generate actions directly from sensor signals. Unlike prior approaches that rely on human-designed action primitives or policies trained on demonstrations \citep{lu2026aspire, xiao2026enpire}, our approach directly generates robot control programs that encapsulate the full perception-planning-control loop, whereas standard policies discard the underlying decision-making process essential for generalization. In practice, we observe the agent writing programs that automatically segment and identify objects in 3D using point-cloud fitting and hue-based identification, then plan actions grounded in this 3D world model, with retry loops and exception handling naturally incorporated into the code. Because actions are planned directly on perceived 3D objects rather than learned through imitation, this approach generalizes to far more diverse object layouts and domain variations than standard VLAs or WAMs. For semantically rich perception in open-world environments, the agent can program direct API calls to LLMs (such as Gemini) within the code to assist with object identification, while incorporating checkpoints and exception handling to step in and verify progress. Over the course of exploration, both action inference latency and token counts decrease as an increasing fraction of decisions becomes automated through code.

As experience scales with exploration, \rax maintains an intra-task memory file system across episodes to record failure lessons and successful strategies, allowing the agent to continually refine its programs. Across multiple tasks and sessions, we maintain a repository of modular programs as verified skills for various tasks and manipulation primitives. Each skill includes the executable code for a specific task, accompanied by execution logs from successful runs, clear preconditions, and verification checkpoints, allowing skills to be composed and reused for complex downstream tasks. In real-world mobile manipulation experiments in household settings, the agent practices individual tasks, each in its own exploration session, such as picking and placing boxes or tucking in office chairs, and commits them as verified skills. When subsequently evaluated on a composite, long-horizon task involving more than six sequential object manipulations, the agent achieves substantial automation and efficiency gains by reusing these stored skills. We envision agents practicing skills in a distributed, parallel fashion to build a rich repository of diverse capabilities, thereby enabling continual lifelong learning.

Using \rax, we conduct \textbf{the first large-scale, multi-embodiment, systematic study of
scaling exploration for robot agents}, spanning fine-grained precision and long-horizon tasks
across LIBERO, LIBERO-Pro, RoboDojo, DexJoCo, BEHAVIOR-1K, and HumanoidArena in simulation and
on two real robots. We evaluate the resulting policies on unseen scenes, separating
exploration from evaluation. Performance grows steadily with exploration, and on
fine-grained and long-horizon tasks on which state-of-the-art methods rarely succeed, \rax
achieves high success rates, surpassing the strongest prior models by +37.4 on average over
five benchmarks (\Cref{fig:teaser}, \Cref{app:teaser}). Without any expert action trajectories, \rax sets a new
state of the art on RoboDojo (+53.4), BEHAVIOR-1K (+38.3), DexJoCo (+31.4), and
LIBERO-Pro (+21.2). On HumanoidArena, it sets a new state of the art (+7.9) with three
teleoperated demonstrations per task, where every baseline uses 100. On real Franka and Rainbow RB-Y1
robots, \rax masters five complex tabletop manipulation tasks and a long-horizon mobile
manipulation task with zero demonstrations, and verified skills accumulated across sessions let
later sessions build on earlier ones.

We summarize our contributions below:
\begin{enumerate}
    \item We introduce \textbf{\rax} and establish \textbf{scaling online exploration} as a foundational principle for agentic robotics. In \rax, the agent operates directly through perception and robot control interfaces without policy training, compiling successful decision sequences into executable code programs that run autonomous execution loops.

    \item Using \rax, we conduct the first systematic, multi-embodiment study of scaling online exploration across simulation and real robot environments. \rax acquires diverse manipulation capabilities solely through autonomous exploration, outperforming state-of-the-art baselines by $+37.4$ points on average across five simulation benchmarks, while solving high-precision, deformable, articulated, and long-horizon manipulation tasks on real robots without human demonstrations.

    \item We demonstrate rapid adaptation and continual learning on real robots, showing that generated programs can be selectively updated in response to environmental and task shifts while retaining verified skills in a persistent repository for composition and reuse in complex tasks.
\end{enumerate}
We release the full architecture of \rax, together with its agent instructions, skills,
benchmark integrations, and evaluation protocols, as a generalist robot-agent platform, on
which a new robot can be brought up through an interface alone, without data collection or
policy training.

\section{Related Work}
\label{sec:related}

\paragraph{Language Models for Robot Control.}
A substantial body of work uses pretrained language models as high-level planners or program
generators for robots. SayCan grounds language-model plans in learned
affordances~\citep{ahn2022can}, Inner Monologue folds environment feedback into iterative
replanning~\citep{huang2022inner}, and Code as Policies, ProgPrompt, and VoxPoser synthesize
executable programs, situated task plans, or 3D value maps directly from
instructions~\citep{liang2023code,singh2022progprompt,huang2023voxposer}. These methods
typically invoke the model zero-shot or with a few in-context examples, producing a plan or a
program that is executed with little opportunity to improve from experience. \textbf{\rax}
instead treats the language model as a \emph{learner}: it interacts with the robot over many
episodes and revises its perception and control programs according to what it observes.

\paragraph{Agentic Robot Learning.}
More recently, several works let coding agents improve robot behavior through interaction.
ENPIRE runs agents in a closed loop of automatic reset, rollout, and policy training on a
fleet of bimanual arms, and accelerates discovery by scaling the fleet~\citep{xiao2026enpire};
FAEA and Encore refine policy programs from execution feedback, the latter starting from a few
teleoperated demonstrations~\citep{tsui2026demonstrationfreeroboticcontrolllm,kang2026encore}; Baton explores long-horizon
tasks subtask by subtask with transition-aware memory~\citep{xu2026don}; SimEX develops robot
skills in a simulator reconstructed from the physical workspace before a few trials on the
real robot~\citep{hu2026simex}; RPG practices in simulation on tasks derived from an offline
trajectory dataset and refines a library of reusable skills~\citep{wang2026reconstruct}; EmbodiedSWE
turns coding-agent solutions into training data for vision-language-action
models~\citep{you2026embodiedswe}; and CaP-X benchmarks coding agents on
manipulation~\citep{fu2026cap}. These works establish that agents can improve robot behavior
through interaction, and have so far demonstrated it mainly in tabletop manipulation with
parallel-jaw
grippers~\citep{xiao2026enpire,tsui2026demonstrationfreeroboticcontrolllm,kang2026encore,xu2026don,hu2026simex,wang2026reconstruct},
often on a single platform~\citep{xiao2026enpire,hu2026simex,wang2026reconstruct}, with humanoid and
mobile settings explored through pretrained locomotion~\citep{you2026embodiedswe} or short
pick-up tasks~\citep{fu2026cap}. Several also draw on demonstrations~\citep{kang2026encore},
offline datasets~\citep{wang2026reconstruct}, reconstructed simulators~\citep{hu2026simex}, or neural
policy training~\citep{xiao2026enpire}. \textbf{\rax} complements these efforts: it learns each
task on the robot itself, from its own exploration, and does so across single-arm and bimanual
manipulators, dexterous hands, mobile manipulators performing long-horizon household
activities, and legged humanoids whose locomotion the agent constructs itself, on established
benchmarks and on real robots. On this basis, it studies systematically how performance scales
with exploration, with 3D perception, and with accumulated memory.

\paragraph{Scaling Robot Learning.}
Vision-language-action models~\citep{brohan2023rt,kim2024openvla,black2024pi_0} and imitation
learning methods~\citep{zhao2023learning,chi2025diffusion} improve robot capability by scaling
demonstration data and model size, and world action models extend this recipe by learning to
predict how the scene evolves together with the actions that produce
it~\citep{wang2026openwam,yuan2026fast,kim2026cosmos,bi2026motus}. To cover multiple embodiments, generalist
policies pool demonstrations across robots into a single training
mixture~\citep{o2024open,team2024octo,doshi2024scaling,intelligence2025pi_,bjorck2025gr00t},
and a robot outside this mixture requires data of its own and further fine-tuning. In language
modeling, computation spent at inference has emerged as a further scaling
axis~\citep{guo2025deepseek,openai2024learning}. \textbf{\rax} instead studies how robot
capability grows with online experience acquired and organized by the agent itself, without
additional demonstrations or parameter updates. Because this experience is gathered on the
robot at hand, \rax extends to a new embodiment without pooled data, and adapts to it within a
few episodes of exploration.

\paragraph{Memory and Cognitive Architectures.}
Language agents can retain experience as explicit memory: Voyager grows a library of code
skills through open-ended exploration in a game~\citep{wang2023voyager}, and Reflexion and
ExpeL distill feedback into textual lessons that improve later
attempts~\citep{shinn2023reflexion,zhao2024expel}. At a broader level, classical cognitive
architectures such as Soar and ACT-R structure intelligent behavior around perception,
memory, and deliberation~\citep{LAIRD19871,Anderson2004AnIT}, CoALA revisits this structure
for language agents~\citep{sumers2023cognitive}, and Sutton's common model of the intelligent
agent~\citep{sutton2022quest} and OaK architecture~\citep{sutton2022alberta} hold that all
task-specific knowledge should be acquired at runtime from the agent's own experience, a view
developed further as the era of experience~\citep{SilverWelcomeTT}. \textbf{\rax} brings this
view to physical robots, where every skill is executable control code that enters memory only
after it succeeds without supervision.

\section{The \rax Cognitive Architecture}
\label{sec:architecture}

\begin{figure}[t]
    \centering
    \includegraphics[width=\linewidth]{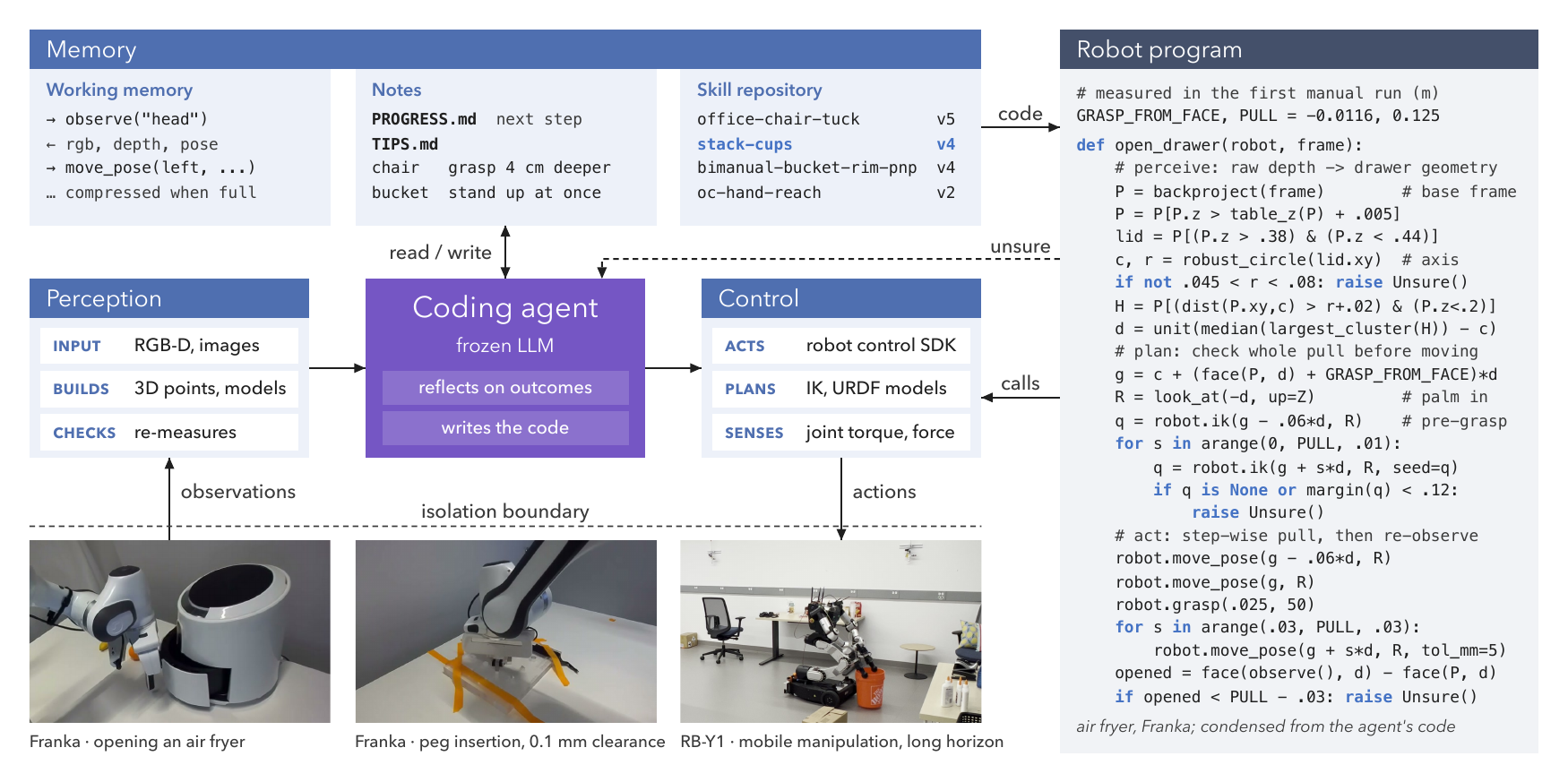}
   \caption{\textbf{System architecture and self-improving loop of \rax.} The coding agent, powered by a frozen LLM, interacts with physical hardware across an isolation boundary using multimodal perception and control SDK interfaces. By reflecting on execution outcomes, the agent continually updates a persistent memory system containing working context, task notes, and a versioned repository of verified skills. During exploration, the agent compiles its perception and planning logic into standalone Python control programs that execute high-frequency loops directly on the robot. When encountering perception or planning failures due to environment variations, the program raises an exception (\texttt{raise Unsure()}), delegating control back to the agent to make adjustments and refine the codebase.}
    \label{fig:architecture}
\end{figure}

Inspired by Sutton's OaK architecture~\citep{sutton2022alberta}, \rax acquires all task-specific knowledge
at runtime from the agent's own experience. Building on task-agnostic robot control interfaces and low-level controllers, the agent autonomously develops task-specific perception, manipulation strategies, and executable skills through exploration.  The following modules surround the language-model core: \textbf{perception and control},
\textbf{autonomous robot program}, and \textbf{memory for continual learning} (\Cref{fig:architecture}).

\subsection{Multimodal Perception} To build a physically accurate model of the world, we equip the agent with multimodal sensory inputs, including RGB images, depth maps, camera parameters, and force signals. The agent can thus construct 3D world models using depth data while regulating contact via force feedback. Notably, the 3D perception and object model fitting behaviors completely emerges from the agent exploration process, while we provide no skills, instructions or examples regarding specific usage of these multimodal sensing signals.

\begin{figure}[htb]
    \centering
    \includegraphics[width=\linewidth]{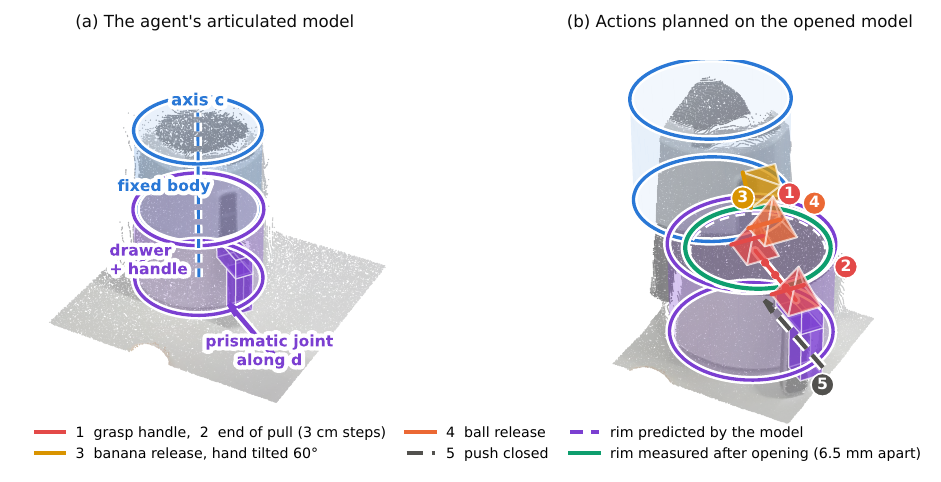}
    \caption{\textbf{Agentic 3D perception and motion planning on a real robot.} \textbf{(a)} The articulated air fryer model constructed programmatically by the agent from point-cloud measurements and fit to RGB-D observations, consisting of a fixed chassis and a drawer articulated via a prismatic joint along pull direction $d$ relative to central axis $c$. \textbf{(b)} Action trajectories planned on the instantiated model with the drawer opened by $13.1\,\text{cm}$. End-effector poses are visualized as pyramids whose bases face the approach vector and whose thick edges indicate the finger closure axis. The underlying geometric estimator maintains high fidelity, yielding only a $6.5\,\text{mm}$ displacement between the predicted basket rim and the post-opening physical measurement. \Cref{fig:airfry-stills} illustrates the corresponding execution sequence on the physical robot.}
    \label{fig:perception}
\end{figure}

A real-robot task illustrates this process: opening an air fryer's drawer, placing a banana in its basket, and closing the drawer (\Cref{fig:perception}). The agent lifts RGB-D observations into a point cloud in the robot's base frame and constructs a geometric model of the fryer from its own measurements, localizing it in each new scene layout by fitting the fryer's axis and handle direction with robust estimators. Because the drawer moves, the model is articulated, consisting of a fixed body and a drawer that slides along a prismatic joint. The agent plans every action, including grasping, pulling, and releasing, on this model and checks each motion for collisions. \Cref{app:perception} details the model, the estimators, and how the program uses them. On a humanoid, the same process yields a gravity-aligned frame from the floor plane, landmarks for self-localization, and tracking updates after every gait cycle.

\subsection{Planning and Control}
Given the accurate world model, the agent then directly plans the robot actions in 3D world using robot control SDK and kinematic solvers. As the interface between the agent and the physical robot,
we expose the robot's control SDK as MCP tools, together with helpers such as an inverse
kinematics (IK) solver and the robot's URDF model. With these tools, the agent plans each motion
and searches for feasible trajectories, accounting for the robot's posture, collisions, and
joint limits. In the air fryer task (\Cref{app:real-airfry}), the drawer's handle has a short horizontal arm on top and
a grip hanging down from its end. Grasping the grip from the front, as a person would, leaves
the arm almost no joint margin near its base for the pull, so the agent grasps the short arm from
above and pulls the drawer open in 3\,cm steps (actions 1 and 2 in \Cref{fig:perception}(b)).
The 18\,cm banana barely fits the 19\,cm basket: released from a level hand, it touched the top
of the fryer, and the opening fingers would hit the fryer's body. The agent therefore tilts the
hand by $60^\circ$ about the banana's axis, with the banana's concave side toward the fryer, and
releases it with its middle just beyond the basket's center (action 3). The agent found these
motions itself, by searching through the control tools during exploration, without human advice
or demonstrations; \Cref{fig:airfry-stills} shows them on the real robot.

\subsection{From Exploration to Autonomous Robot Program}
\label{sec:code}

Generating each action with LLM-reasoning is expensive and slow. To enable high-frequency close-loop planning, the agent automates the decision process into a autonomous robot program with code that can directly execute to complete the task. Unlike prior work on coding agents for robotics \citep{lu2026aspire, xiao2026enpire}, we do not rely on hand-designed primitive skills or train policies by imitating human demonstrations, which inherently restricts scaling beyond human capability \citep{sutton2019bitter}. Instead, \rax agents directly control the robot without pre-trained policies or hand-coded primitives, progressively transferring control from high-level deliberation to autonomous code programs, resulting in \textbf{autonomous decision-making through agent-native coding}. Initially, the agent controls the robot through tool calls while perceiving the scene via the core LLM. Upon achieving task success, it compiles these decisions into code programs that run independently of main-agent deliberation. For example, after identifying object shapes and geometries, the agent generates a 3D object model and automatically estimates its position and pose using depth-derived point clouds when encountering new layouts. This eliminates the need for the main agent to repeatedly perceive and process raw images, which is slow and costly.

The agent decomposes tasks into stages and implements modular skills with programs that check preconditions, execute motions, and verify outcomes, operating much like a software development process through testing, debugging, and code versioning. Empirical measurements from exploration, such as gait-cycle distance or reliable grasp orientation, become program parameters. The agent tests successive code versions during exploration and uses stage-level failures to identify which specific skill requires revision. The hammer-delivery program on the HumanoidArena DoubleDesk task (\Cref{app:doubledesk}), for instance, decouples approach, grasp, transit, obstacle crossing, and placement so that every failure traces directly to a specific stage (\Cref{fig:dd-strip}).

Once reliable, the program is tested unattended in a new scene with exception handling and agent monitoring the generated logs in case of emergency situations. If perception or planning fails due to significant open-world variations, the program throws an exception, enabling the agent to step in, perceive the environment, and make necessary adjustments. Through scaled program testing and debugging during exploration, the result is a \textbf{fully automated program} that perceives, decides, and acts in unseen scenes with minimal or no agent intervention. Automated perception and planning allow it to adapt to new layouts in ways that replayed trajectories cannot. Deployment thus spans a continuum from the agent acting alone to a fully automated program, along which the agent advances as it accumulates experience.

\subsection{Continual Learning with Memory}
\label{sec:continual}
As online experience scales and decision-making becomes automated, \rax continually expands its capabilities by constructing persistent memory systems for reuse across episodes and tasks.  Persistent memory carries experience at multiple timescales (\Cref{fig:architecture}, top). \emph{Within a session}, notes
retain the current strategy, confirmed results, failed approaches, and useful parameters
through context compression, which lives in the agent scratch pad in workspace. \emph{Across sessions and tasks}, a skill \textit{repository} stores each skill with its code, the conditions under which it
applies, and evidence of its execution. It is continually contributed and merged by agents across different sessions and tasks, evolving with more experience.

A skill enters the repository only after it executes successfully without supervision, together
with the evidence of that execution. Later sessions retrieve and compose skills that are
known to work instead of rediscovering the same behavior. The benefit is largest on real
robots, where every episode costs operator time: a grasp or a drawer-opening skill becomes a
building block for longer tasks, and each session leaves the library more capable than it
found it.

\paragraph{Learning via Practice} Much like humans, who repeatedly practice individual skills to gain proficiency before rehearsing complex, long-horizon tasks end-to-end, robot agents follow a similar cognitive progression. Here we depart from the conventional per-task training and evaluation protocol in robotics and frame robot learning as a continual process, where it generates and refines skills through practice and gradually moves on to more complex tasks by reusing the skills in the repository. Taking household tasks as an example, we let the agent practice individual subtasks, such as picking and placing (PnP) boxes and buckets or sorting tabletop items; each practice session is an exploration session on a single sub-task. During this process, the agent develops autonomous programs, rehearses them end-to-end to improve robustness, and eventually commits them to a multi-task skill library. During deployment on long-horizon compositional tasks, the agent reuses these subtask skills directly while verifying execution evidence and preconditions to handle environmental variations beyond those encountered during practice. Such skill acquisition and practice can also occur in parallel across multiple robots and tasks, enabling collective learning across multi-robot systems.

\subsection{Parallel Multi-Agent and Multi-Environment Exploration}
\label{sec:parallel}
Online exploration can be time-consuming due to the latency of agent reasoning and environment execution. To accelerate this process, we parallelize exploration across multiple agents and environments, enabling agents to evaluate distinct strategies concurrently rather than sequentially within a single environment. In our simulation experiments, leveraging parallel environments and agents reduces the wall-clock time required to reach target performance by more than $2\times$, highlighting the scalability of parallelized exploration for robotic systems.
\Cref{sec:exp-parallel} measures how parallel exploration changes the time to consume the
exploration budget and the score of the resulting program.

\section{The \rax System}
\label{sec:system}

\begin{figure}[t]
    \centering
    \includegraphics[width=\linewidth]{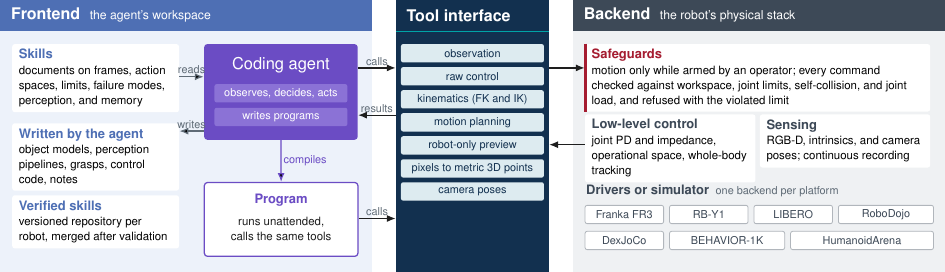}
    \caption{\textbf{Frontend--backend separation.} The agent and its programs act only through the
    tool interface; the backend owns control, sensing, and safeguards on every platform.}
    \label{fig:frontend-backend}
\end{figure}

\rax supports a wide range of embodiments in simulation and in the real world, including
single-arm and bimanual manipulators, dexterous hands, mobile manipulators, and humanoids, with
one agent architecture. To do so, it adopts a separation familiar from software engineering, in
which a frontend is kept apart from the backend that serves it. In \rax, the backend is the robot's physical stack: the
hardware drivers or the simulator, the low-level controllers, the sensor pipelines, and the
safeguards around them. The frontend is the agent's workspace, where the agent observes,
decides, and acts, and where it writes the programs that encode its decision logic. The two meet
only at a tool interface (\Cref{fig:frontend-backend}), and the same frontend connects to the backend of every platform, from
the real Franka and Rainbow RB-Y1 robots to each simulated benchmark. The agent never touches the
backend: it cannot modify controllers, drivers, or robot infrastructure. Because the agent acts only through these tools, its programs contain the
decision logic of a task, namely perception, planning, and the checks between stages, while
low-level control, from joint-level PD control to whole-body motion tracking on humanoids, and
the enforcement of safety limits remain in the backend. The remainder of this
section describes the separation and what each side provides (\Cref{sec:separation}), and the
control interface through which one frontend drives arms, dexterous hands, mobile manipulators,
and humanoids (\Cref{sec:control}).

\subsection{Frontend--Backend Separation}
\label{sec:separation}

On a real robot, the backend owns everything that keeps the hardware calibrated and safe. It runs
the vendor's drivers and real-time controllers, serves calibrated RGB-D images together with the
intrinsics and poses of their cameras, and records the robot's sensors and state continuously,
so that the agent can review the outcome of an action after the fact. It also enforces the
safeguards. Motion is enabled only while an operator has armed the robot, and disarming halts it
immediately; every motion command is validated before execution against workspace bounds, joint
limits, and self-collision, and on the RB-Y1 mobile manipulator also against joint load. A command that violates
a limit is refused before the robot moves, together with the limit it violated, so that the agent
can change its approach.

The frontend receives exactly two things from the system, tools and skills. The \emph{tools} expose as much of the backend as
possible: observation, raw control, and model-based helpers in the form of forward and inverse
kinematics, motion planning where the platform offers it, and a preview that executes a candidate
motion on the robot model alone, without the scene. They also provide the geometric primitives on
which perception is built, namely the back-projection of pixels to metric 3D points, the poses of
the cameras in the robot's frame, and the kinematic chain from each camera to the end-effectors.
The \emph{skills} are documents that describe how to work with these tools: the robot's frames,
action spaces, rates, limits, and failure modes, the conventions of perception, and the workflow
of exploration and memory. They are organized along four axes, namely task strategy,
coordination, perception, and robot-specific control, and are composed into a single workspace
for each session according to its embodiment and observation mode. Everything that solves a task,
from object models and perception pipelines to grasps and control code, is written by the agent,
which works in an isolated container with a persistent workspace and standard programming tools.

\begin{figure}[htb]
    \centering
    \includegraphics[width=\linewidth]{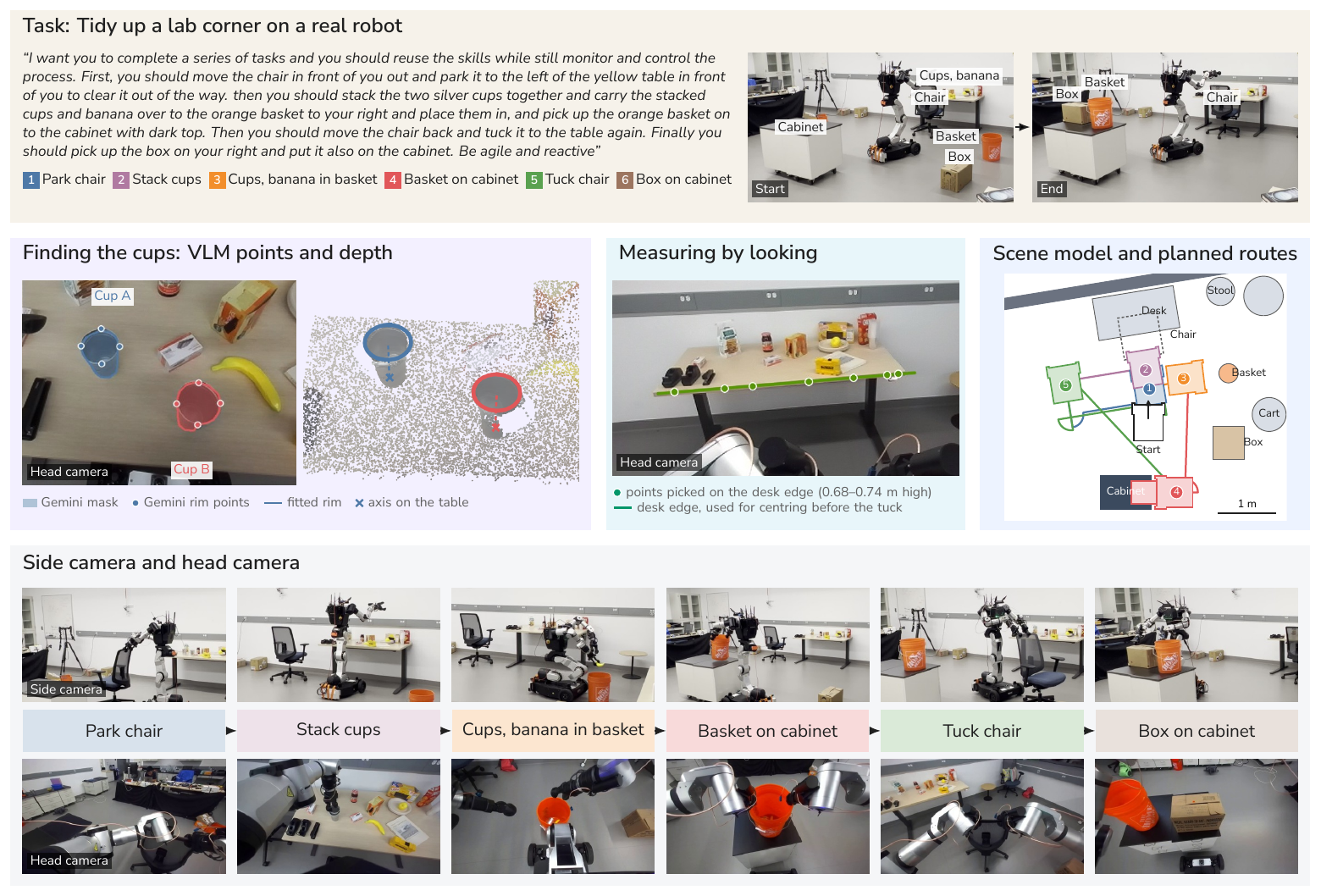}
    \caption{\textbf{A long-horizon task on the real RB-Y1.} The agent composes six sub-tasks from skills
    learned in earlier sessions. It finds the cups from Gemini's rim points and depth, measures the desk
    edge by picking points in the head image, and plans the base routes on a model of the scene.}
    \label{fig:rby1}
\end{figure}

This division has three consequences. First, programs written by the agent call exactly
the tools that the agent itself calls, so code developed interactively runs unchanged as
a fully automated program. Second, the interface returns compact results, consisting of
structured state and the final image of each motion, while the complete images and actions of
every motion are stored as episodic evidence that the agent retrieves on demand, which keeps the
agent's context small over sessions that last many hours. Third, on real robots, the skills that
agents verify are kept in a versioned repository for each robot, in which every session works on
its own branch, and a contribution is merged only after its structure, its entry point, and the
evidence of its unattended success have been validated.

Most importantly, the separation makes \rax multi-embodiment by construction. Bringing up a new
robot requires only a backend that exposes it through the tool interface, together with documentation of its control and
sensing interfaces, while the frontend, the organization of
memory, and the evaluation protocol are
shared across all platforms. No expert teleoperation data is collected and no policy is trained to bring a robot into
\rax. The single-arm, bimanual, dexterous, mobile, and humanoid embodiments of our experiments,
in simulation and in the real world, differ only in their backends and in the
documentation of their interfaces, and the agent learns to control each of them through exploration.

\subsection{Multi-Embodiment Control Interface}
\label{sec:control}

\rax adopts a layered control design. Low-level controllers, from joint position and impedance controllers to whole-body motion
trackers, run in the backend and are responsible for high-rate tracking and
stability, while the agent makes decisions at the level of short action chunks and motion goals.
Each platform exposes raw control for full expressiveness, together with model-based helpers in
the form of forward and inverse kinematics, collision-aware motion
planning~\citep{sundaralingam2023curobo}, and a robot-only preview that executes a candidate
motion on the robot model without the scene. The agent thus chooses its own level of
abstraction, from individual joint targets to planned free-space motions.
\Cref{tab:control} summarizes this construction for each platform.

\begin{table}[htb]
    \centering
    \small
    \caption{\textbf{Low-level control on each platform.} EEF: end effector; OSC: operational-space
    control; PD: proportional-derivative.}
    \label{tab:control}
    \begin{tabularx}{\linewidth}{@{}>{\raggedright\arraybackslash}p{2.5cm}>{\raggedright\arraybackslash}p{3.0cm}>{\raggedright\arraybackslash}X>{\raggedright\arraybackslash}X@{}}
        \toprule
        \textbf{Platform} & \textbf{Low-level control} & \textbf{Agent action space}
            & \textbf{Model-based helpers} \\
        \midrule
        LIBERO, LIBERO-Pro \newline \footnotesize single-arm Franka Panda
            & EEF pose control (OSC)
            & EEF position and rotation deltas; gripper command
            & Pixel-to-3D projection from depth \\
        \addlinespace
        RoboDojo \newline \footnotesize dual-arm ARX X5
            & Joint position control (PD)
            & Joint-target chunks for both arms and grippers; bounded EEF steps
            & Forward kinematics, differential IK, pose algebra, collision-aware motion
              planning, robot-only preview \\
        \addlinespace
        DexJoCo \newline \footnotesize Franka Panda + Allegro hands
            & Arm: EEF pose control (OSC); hand: joint position control
            & Flange pose and sixteen hand joint targets per arm, single- or dual-arm
            & Pixel-to-3D projection from depth, robot-only preview \\
        \addlinespace
        BEHAVIOR-1K \newline \footnotesize Galaxea R1Pro
            & Base velocity control; trunk and arm joint position control
            & Base velocity, trunk and arm joint targets, grippers; closed-loop base moves
              from odometry
            & Closed-loop arm IK, arm and trunk motion planning, robot-only preview \\
        \addlinespace
        HumanoidArena \newline \footnotesize Unitree G1 + Dex3
            & Whole-body motion tracker
            & Whole-body reference frames: heading, 29 joint targets, hand open or close
            & Arm and waist motion planning to hand poses; kinematic model for forward
              kinematics and image projection \\
        \addlinespace
        Franka FR3 \newline \footnotesize real
            & Joint impedance control
            & Straight-line EEF moves, joint moves, force-controlled grasps
            & Inverse kinematics; path checks against workspace and table \\
        \addlinespace
        Rainbow RB-Y1 \newline \footnotesize real, mobile manipulator
            & Joint position control; Cartesian IK commands for torso and arms; wheel velocity
              control
            & EEF poses; torso, arm, and head joint targets; coordinated torso and arm motions; base
              motion; head gaze
            & Forward and inverse kinematics; dry-run checks of collision, joint load, and
              center of mass \\
        \bottomrule
    \end{tabularx}
\end{table}

On the real robots, the backend wraps each vendor's own control stack. The Franka FR3 arm runs joint impedance control, and the agent commands straight-line Cartesian
moves of the tool center point, joint-space moves whose paths are checked against the workspace
and the table before the arm moves, and force-controlled grasps that close the gripper to a
target width with a set force. The Rainbow RB-Y1, a wheeled mobile manipulator with a torso, two arms, and
a two-joint head that carries an RGB-D camera, runs joint position control on its robot controller,
which also accepts Cartesian IK commands for the torso and arms. The agent commands end-effector
poses, joint-space postures of the torso, arms, and head, coordinated motions of the torso and arms, motions
of the differential-drive base, and the gaze of the head camera. It can query the vendor's
kinematic model for the collisions, joint load, and center of mass of any posture, and dry-run any
motion command to learn whether it would be accepted. Neither robot offers a motion planner, so
the agent composes long motions from short checked moves through clear intermediate postures.

On the HumanoidArena
benchmark~\citep{wang2026humanoidarena}, the agent commands the Unitree G1 through whole-body
kinematic references rather than joint torques: each reference specifies a desired heading, a
full joint configuration, and hand commands, and a whole-body motion tracker converts these references into balanced whole-body motion. Because the robot moves through space only by
stepping, locomotion is not provided as a primitive but must be constructed by the agent from leg
references, and agents do so by measuring their own gait, calibrating step length, and composing
walking, turning, and stepping over obstacles from these measurements.

On DexJoCo~\citep{wang2026dexjoco}, the official
operational-space controller tracks the flange pose of each arm while position actuators drive
the sixteen joints of each Allegro hand, and no motion planner is
provided, so the agent plans the motion of each arm and the configuration of each hand itself. The remaining simulated
platforms span the same range of abstraction. The single arm of LIBERO and
LIBERO-Pro~\citep{liu2023libero,zhou2025libero} is driven through the benchmark's own
operational-space controller by relative end-effector commands alone, whereas on
RoboDojo~\citep{chen2026robodojo} and BEHAVIOR-1K~\citep{li2023behavior} the native controllers
execute joint targets under each benchmark's own dynamics, and inverse kinematics and motion
planning are offered as helpers alongside raw control rather than as replacements for it. Motion
planning accounts for the robot and static structures but never for task objects, which the agent
must perceive itself.

\section{Experiments}
\label{sec:experiments}

In this section we report our simulation benchmark results as well as case studies on both simulation and real robot experiments.
\subsection{Setup}
\label{sec:exp-setup}

We comprehensively evaluate \rax in both simulated environments and on real-world robotic hardware. In simulation, the evaluation spans six benchmarks covering single-arm, bimanual, dexterous, mobile, and humanoid embodiments: RoboDojo, BEHAVIOR-1K, HumanoidArena, DexJoCo, LIBERO, and LIBERO-Pro. In each session, a coding agent starts in a localized workspace containing only the platform's baseline skills. Guided by a task description and RGB-D observations, the agent explores training scenes within a strictly defined episode budget. The submitted code is then executed as a fully automated program under each benchmark's official evaluation protocol (\Cref{app:tasks}).

On physical robots, \rax is deployed on a Franka FR3 manipulator to master five distinct tabletop tasks: (1) \emph{memory}, tracking and finding a hidden cube randomly shuffled under four cups (\Cref{app:real-cube}); (2) \emph{articulated objects}, opening an air fryer drawer, inserting a banana, and closing it (\Cref{app:real-airfry}); (3) \emph{deformable objects}, wiping a stain off a table using a wet towel (\Cref{app:real-wipe}); (4) \emph{precision}, inserting pegs into a ManipulationNet board with a stringent 0.1,mm clearance (\Cref{app:real-peg}); and (5) \emph{precision and long horizon}, dismantling and rebuilding a Bandu tower of irregular pieces (\Cref{app:real-bandu}).

Additionally, we evaluate \rax on a Rainbow RB-Y1 mobile manipulator to solve a long-horizon logistical task. This involves pulling out a chair, stacking cups, picking up a banana, loading the items into a basket, carrying the basket to a cabinet, tucking the chair back, and subsequently transferring a box to the cabinet (\Cref{app:real-rby1}). Crucially, the agent receives only natural-language instructions and zero teleoperated demonstrations. It is designed to autonomously evaluate its own success criteria, only prompting a human operator to reset the physical scene between trials.

\subsection{Simulation Results}
\label{sec:exp-main}
Here we report the simulation benchmark results across RoboDojo, BEHAVIOR-1K, HumanoidArena, DexJoCo, LIBERO, and LIBERO-Pro.
\subsubsection{RoboDojo}\label{sec:exp-scaling}

\paragraph{Setup.} We measure exploration scaling on seven RoboDojo tasks with the dual-arm ARX X5 (\Cref{app:robodojo}), spanning domain-randomized variants (\textit{make toast random}, \textit{hang mugs random}, \textit{sweep blocks random}), as well as precision (\textit{plug in charger}), memory (\textit{swap blocks}), long-horizon (\textit{make kong}), and open-instruction (\textit{pour by language}) tasks. For each task, a single agent explores a single environment from an empty workspace, given the task's public scoring rubric with a short clarification of its conditions, RGB-D observations, and one image of the task's completed state. Sessions differ only in the exploration budget $N$ (i.e., the number of exploration episodes), with each episode conducted in a new exploration scene. At the end of exploration, the agent submits its rehearsed bundle. The resulting fully automated program is then evaluated without the agent on 30 to 50 unseen scenes, scored by the native evaluator with partial credit.

\paragraph{Baselines.} We compare our approach with the five strongest learned policies on the RoboDojo-Sim leaderboard~\citep{chen2026robodojo}, which are trained on demonstrations, and with PhysicalRSI, an agent-based system listed separately on the same leaderboard (\Cref{tab:exploration-scaling}).

\paragraph{Results.} With its largest budget on each task, \rax exceeds every baseline on all seven tasks, averaging 78.3\% compared to 24.9\% for VPP2, the strongest policy on the leaderboard (\Cref{tab:exploration-scaling}). The margins are largest where learned policies make little progress: on \textit{sweep blocks random}, \rax reaches 71.7\% where the best baseline achieves only 8.0\%; and on \textit{pour by language}, which every learned policy fails, it reaches 100.0\%. Performance consistently grows with the exploration budget. \textit{Plug in charger} rises from 26.0\% with 4 exploration episodes to 94.0\% with 8; \textit{swap blocks} improves from 34.0\% to 100.0\% with 16; and \textit{make kong}---for which the agent submitted no program within 8 episodes---increases from 64.0\% with 16 episodes to 96.0\% with 32. The domain-randomized tasks require the largest budgets: \textit{make toast random} rises from 2.5\% with 16 episodes to 31.7\% with 32 and 47.2\% with 64, and \textit{sweep blocks random} improves from 53.3\% to 71.7\%, while \textit{hang mugs random} stays between 33.3\% and 39.2\% across all budgets.

\begin{table}[htb]
    \centering
    \small
    \setlength{\tabcolsep}{4pt}
    \caption{\textbf{Scaling exploration on RoboDojo.} Score (\%) on unseen layouts;
    baselines from the top five policies of the RoboDojo-Sim leaderboard~\citep{chen2026robodojo}. Cells
    with more than one run report the mean. --: not run. $^\dagger$No program submitted.
    $^\ddagger$Agent + VLA; listed among agents and unranked on the leaderboard.}
    \label{tab:exploration-scaling}
    \begin{tabular}{@{}lcccccccc@{}}
        \toprule
        & & Make toast & Hang mugs & Sweep blocks & Plug & Swap & Make & Pour by \\
        \textbf{Method} & \textbf{Budget}
            & random & random & random & charger & blocks & kong
            & language \\
        \midrule
        VPP2           & -- & 38.0 & 13.5 & 1.3 & 16.7 & 26.0 & 78.7 & 0.0 \\
        Awomo-0.5      & -- & 7.0  & 9.6  & 0.0 & 7.3  & 4.7  & 30.0 & 0.0 \\
        Simate-beta    & -- & 7.7  & 11.4 & 0.0 & 29.3 & 38.7 & 76.0 & 0.0 \\
        Rex-M1 Preview & -- & 14.3 & 20.2 & 8.0 & 3.3  & 13.3 & 33.3 & 0.0 \\
        VPP2-Preview   & -- & 13.7 & 14.5 & 2.7 & 6.0  & 22.7 & 34.7 & 0.0 \\
        PhysicalRSI$^\ddagger$ & -- & 4.7 & 2.0 & 0.0 & 30.7 & 0.0 & 82.0 & 10.1 \\
        \midrule
        \multirow{5}{*}{\rax}
                       & 4  & --   & --   & --   & 26.0 & 34.0 & 0.0 & 94.4 \\
                       & 8  & --   & --   & --   & \textbf{94.0} & 74.0 & 0.0$^\dagger$ & \textbf{100.0} \\
                       & 16 & 2.5  & 36.9 & 53.3 & -- & \textbf{100.0} & 64.0 & -- \\
                       & 32 & 31.7 & 33.3 & 68.3 & -- & --   & \textbf{96.0} & -- \\
                       & 64 & \textbf{47.2} & \textbf{39.2} & \textbf{71.7} & -- & -- & -- & -- \\
        \bottomrule
    \end{tabular}
\end{table}


\subsubsection{BEHAVIOR-1K}\label{sec:exp-behavior}

\paragraph{Setup.} We evaluate \rax on six long-horizon household tasks from the 2025 BEHAVIOR Challenge~\citep{li2023behavior}, where previous challenge entries have struggled to make progress. For each task, cooperating agents explore parallel environments using the Galaxea R1Pro for 128 exploration episodes, given the task description, RGB-D observations, and an image sequence of a successful completion (keyframes of official demonstrations, without actions or states). We evaluate the agent-in-the-loop program: a fresh copy of the agent starts from the frozen workspace and attempts each of the ten public test instances of the challenge---the same instances used to report public results.

\paragraph{Review reports.} Throughout exploration, after each run of the agent's code, a separate review model provides detailed feedback to the agent regarding the outcome. This report outlines any missed goal conditions and their likely causes, suggests the most valuable code changes, and includes the run's partial score alongside key camera frames (\Cref{app:behavior}).

\begin{figure}[htb]
    \centering
    \includegraphics[width=\linewidth]{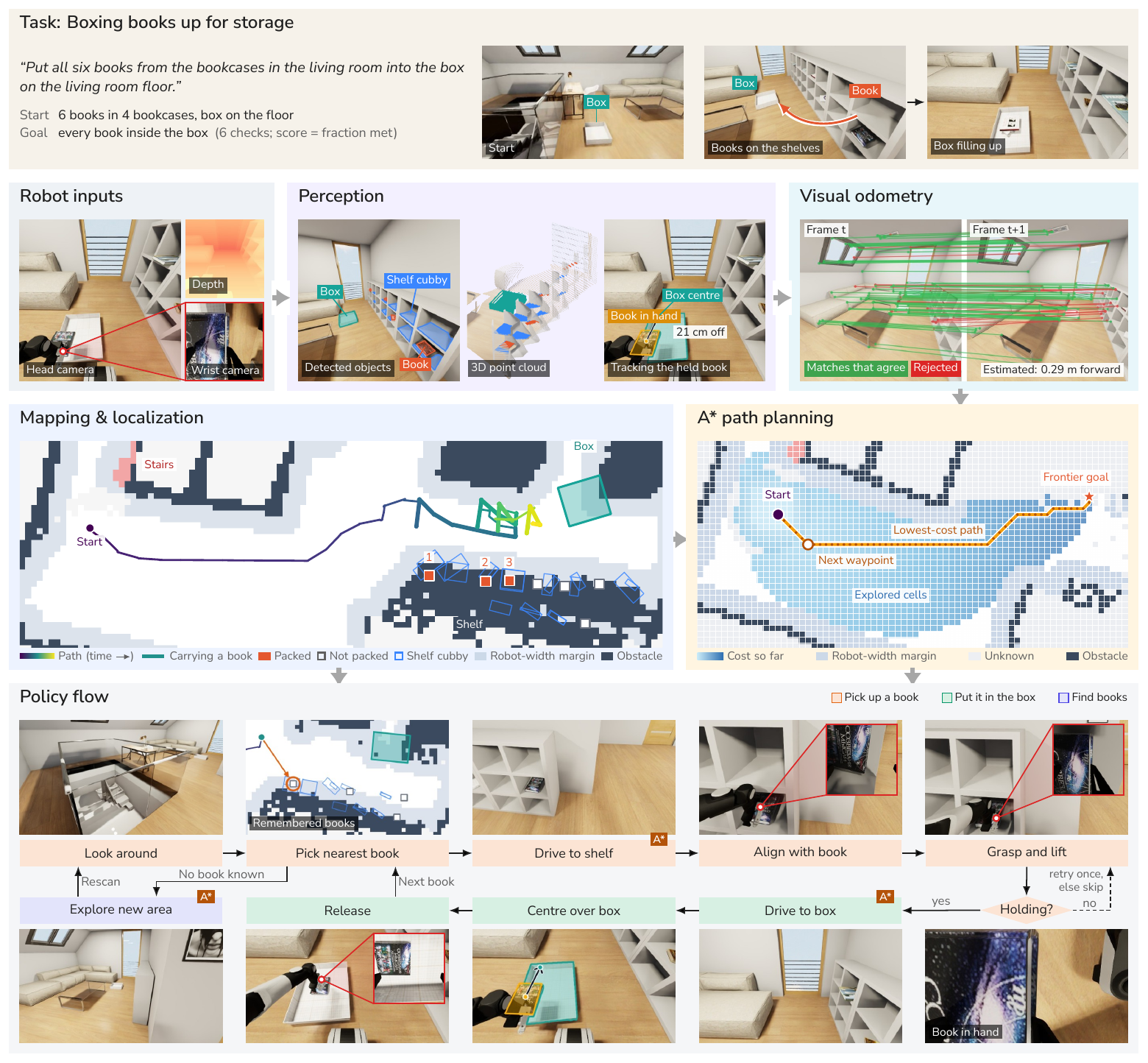}
    \caption{\textbf{The program \rax writes for boxing books on BEHAVIOR-1K} (\Cref{tab:behavior}).
    It detects the books, shelves and box in the head camera, tracks the robot's motion by visual
    odometry, maps the scene, and plans paths with A*; the policy flow shows the camera view at each step.}
    \label{fig:behavior-boxing}
\end{figure}

\paragraph{Scoring and baselines.} Episodes are evaluated using the challenge's Q-score, which measures the fraction of satisfied goal conditions and awards full credit for a complete success. We compare our approach against the five highest-ranked entries from the challenge (\Cref{tab:behavior}). \Cref{fig:behavior-boxing} illustrates the program \rax generates for the \textit{boxing books} task.

\begin{table}[htb]
    \centering
    \small
    \setlength{\tabcolsep}{4pt}
    \caption{\textbf{BEHAVIOR-1K.} Q-score (\%) on the ten public test instances; baselines from
    the five highest-ranked entries of the 2025 BEHAVIOR Challenge~\citep{li2023behavior}. $^\dagger$Fully automated program.
    $^\ddagger$Privileged track.}
    \label{tab:behavior}
    \begin{tabular}{@{}lccccccc@{}}
        \toprule
        & Boxing & Getting & Sorting & Clearing & Clean up & Loading & \\
        \textbf{Method} & books & organized & items & food & desk & the car & \textbf{Avg.} \\
        \midrule
        Robot Learning Collective & 0.0 & 2.0 & 6.2 & 8.0 & 12.7 & 20.0 & 8.2 \\
        Comet                     & 0.0 & 0.0 & 0.0 & 0.0 & 0.0  & 0.0  & 0.0 \\
        SimpleAI Robot            & 0.0 & 0.0 & 0.0 & 0.0 & 0.0  & 0.0  & 0.0 \\
        The North Star            & 0.0 & 0.0 & 0.0 & 0.0 & 0.0  & 20.0 & 3.3 \\
        Embodied Intelligence$^\ddagger$ & 0.0 & 0.0 & 0.0 & 0.0 & 0.0 & 0.0 & 0.0 \\
        \midrule
        \rax & \textbf{26.7}$^\dagger$ & \textbf{9.0} & \textbf{75.0} & \textbf{40.0}
             & \textbf{28.2} & \textbf{100.0} & \textbf{46.5} \\
        \bottomrule
    \end{tabular}
\end{table}

\subsubsection{HumanoidArena}\label{sec:exp-humanoidarena}

\paragraph{Setup.} We evaluate \rax on all seven tasks of
HumanoidArena~\citep{wang2026humanoidarena} that have official test episodes: three
human--object interaction tasks (Football, DoubleDesk, and P\&PBox) and four human--scene
interaction tasks (OpenDoor, SitSofa, Boxing, and VisNavi), on a Unitree G1 with Dex3 hands
(\Cref{app:humanoidarena}).
The agent commands whole-body reference poses, which the SONIC motion
tracker~\citep{luo2026sonic} turns into balanced whole-body motion. The agent perceives the scene through the
egocentric RGB-D camera and receives the task description together with three
teleoperated demonstrations, which are available only to the exploring agent. It explores
with a budget of up to 128 episodes on DoubleDesk and P\&PBox and up to 32 episodes on each of
the other tasks, on
scenes drawn from the training distribution and never on a test episode, and the resulting
fully automated program is evaluated alone, without the agent, on unseen scenes, the 60
official test episodes of each task.

\paragraph{Baselines.} We compare with the baselines of the benchmark, ACT~\citep{zhao2023learning}, Diffusion
Policy~\citep{chi2025diffusion}, a flow-matching policy~\citep{bdtri2025lbm}, and the VLA
$\pi_{0.5}$~\citep{intelligence2025pi_}, each trained on 100 teleoperated demonstrations per
task, and with WB-WAM~\citep{qin2026wb}, a world action model pretrained on
1,880 hours of video and motion data and fine-tuned on the same 100 demonstrations per task, all with
the same tracker (\Cref{tab:humanoidarena}).

\paragraph{Results.} \rax reaches an average success rate of 89.8\% (\Cref{tab:humanoidarena}), 7.9 points above WB-WAM and 29.8 points above
the strongest baseline of the benchmark, from three demonstrations instead of one hundred. It
succeeds in every test episode of OpenDoor and Boxing, and on Football and VisNavi it improves
on WB-WAM by 28.3 and 16.6 points. Performance grows with the exploration budget: averaged over
the seven tasks, the success rate rises from 52.1\% with 2 exploration episodes to 80.7\% with 32.
Tasks governed by whole-body posture and locomotion, such as OpenDoor and SitSofa, are solved
within a few episodes, Football and VisNavi rise from 40.0\% and 71.7\% to 98.3\% and 93.3\%,
and the two
whole-body loco-manipulation tasks keep improving with further exploration, P\&PBox from 0.0\% to 83.3\% and DoubleDesk from 0.0\% to 56.7\% at 128
episodes. We also test \rax with the three demonstrations without their actions (\rax$^\star$,
\Cref{sec:exp-noaction}): without any action data, it still
matches the 81.9\% average of WB-WAM, which is fine-tuned on 100 action demonstrations per task.

\begin{table}[htb]
    \centering
    \small
    \setlength{\tabcolsep}{3pt}
    \caption{\textbf{HumanoidArena.} Success rate (\%) on the 60 official test episodes of
    each task, all methods with the SONIC tracker. Baselines from \citet{wang2026humanoidarena}
    and \citet{qin2026wb}. Max: each task at its largest budget. $^\star$Three demonstrations without
    actions. --: not run.}
    \label{tab:humanoidarena}
    \begin{tabular}{@{}lcccccccccc@{}}
        \toprule
        & & & \multicolumn{3}{c}{Human--object interaction} & \multicolumn{4}{c}{Human--scene interaction} & \\
        \cmidrule(lr){4-6}\cmidrule(lr){7-10}
        \textbf{Method} & \textbf{Demos} & \textbf{Budget} & Football & DoubleDesk & P\&PBox & OpenDoor
            & SitSofa & Boxing & VisNavi & \textbf{Avg.} \\
        \midrule
        ACT              & 100 & -- & 16.7 & 18.3 & 56.7 & 78.3 & 73.3 & 56.7 & 33.3 & 47.6 \\
        Diffusion Policy & 100 & -- & 45.0 & 36.7 & 75.0 & 85.0 & 78.3 & 76.7 & 23.3 & 60.0 \\
        Flow matching    & 100 & -- & 13.3 & 38.3 & 73.3 & 70.0 & 15.0 & 70.0 & 38.3 & 45.5 \\
        $\pi_{0.5}$      & 100 & -- & 10.0 & 43.3 & 71.7 & 66.7 & 73.3 & 70.0 & 23.3 & 51.2 \\
        WB-WAM           & 100 & -- & 70.0 & \textbf{65.0} & \textbf{86.7} & 98.3 & 95.0 & 81.7 & 76.7 & 81.9 \\
        \midrule
        \multirow{8}{*}{\rax} & \multirow{8}{*}{3}
                         & 2   & 40.0 & 0.0  & 0.0  & 98.3 & 96.7 & 58.3 & 71.7 & 52.1 \\
        &                & 4   & 51.7 & 1.7  & 16.7 & 93.3 & 98.3 & 95.0 & 80.0 & 62.4 \\
        &                & 8   & 76.7 & 5.0  & 23.3 & 93.3 & \textbf{100.0} & 80.0 & 80.0 & 65.5 \\
        &                & 16  & 96.7 & 28.3 & 20.0 & 93.3 & 96.7 & 80.0 & 83.3 & 71.2 \\
        &                & 32  & \textbf{98.3} & 25.0 & 51.7 & \textbf{100.0} & 96.7 & \textbf{100.0} & \textbf{93.3} & 80.7 \\
        &                & 64  & --   & 25.0 & 46.7 & --   & --   & --   & --   & -- \\
        &                & 128 & --   & 56.7 & 83.3 & --   & --   & --   & --   & -- \\
        \cmidrule(lr){3-11}
        &                & Max & \textbf{98.3} & 56.7 & 83.3 & \textbf{100.0} & 96.7 & \textbf{100.0} & \textbf{93.3} & \textbf{89.8} \\
        \midrule
        \rax$^\star$ & 3 & Max & \textbf{98.3} & 46.7 & 70.0 & 95.0 & \textbf{100.0} & 86.7 & 76.7 & 81.9 \\
        \bottomrule
    \end{tabular}
\end{table}

\subsubsection{DexJoCo}\label{sec:exp-dexjoco}

\paragraph{Setup.} DexJoCo~\citep{wang2026dexjoco} evaluates task-oriented dexterous
manipulation with Franka arms that carry 16-degree-of-freedom Allegro hands, on six
single-arm and five bimanual tasks. The agent commands the end-effector pose of each arm and
the joint positions of each hand. It perceives the scene through
RGB-D observations and receives the task description and
three completion images of expert episodes as a goal reference, but no action demonstrations or
expert videos. For each task, a single agent
explores a single environment from an empty workspace, with the exploration episodes drawn from
the training distribution of the rand-obj setting and never on an evaluation layout. Sessions
differ only in the exploration budget, from 2 to 32 exploration episodes, and the resulting fully
automated program is evaluated alone, without the agent, on unseen scenes, the 150 official
evaluation episodes of each task (\Cref{app:dexjoco}).

\paragraph{Baselines.} We compare with the baselines of
the benchmark, Diffusion Policy with transformer and CNN backbones~\citep{chi2025diffusion},
ACT~\citep{zhao2023learning}, and the VLAs $\pi_{0.5}$~\citep{intelligence2025pi_} and GR00T
N1.5~\citep{bjorck2025gr00t}, each trained on 100 teleoperated demonstrations per task, and
with subsequent results on the same benchmark: GR00T N1.6, the world action models
Fast-WAM~\citep{yuan2026fast}, PointWAM~\citep{park2026pointwam}, which pretrains on 1.15
million human videos, and ME-Dex~\citep{zhang2026me}, and the diffusion transformer
DECO~\citep{li2026deco} (\Cref{tab:dexjoco}).

\paragraph{Results.} Performance grows rapidly with exploration (\Cref{tab:dexjoco}):
averaged over the eleven tasks, the success rate rises from 31.9\% with 2 exploration episodes to
86.9\% with 8 and 96.7\% with 32, and with 32 exploration episodes \rax exceeds the best published
result on every task. Hammer, Mouse, and iPad are solved from two exploration episodes, most
other tasks within four to eight, and the bimanual Assembly within sixteen. Taking the best
session of each task, \rax succeeds in every evaluation episode of eight of the eleven tasks and
reaches an average success rate of 99.5\%, 45.4 points above the strongest baseline of the
benchmark and 34.2 points above the best published average, without a single action demonstration.
The margins are largest on the bimanual tasks that existing methods rarely complete: \rax
solves Hanoi (\Cref{app:hanoi}) and iPad in every episode, where the best published results are 42.0 and
52.0, and reaches 96.0 on Assembly, where the best published result is 28.0.

\begin{table}[htb]
    \centering
    \footnotesize
    \setlength{\tabcolsep}{2.6pt}
    \caption{\textbf{DexJoCo, rand-obj setting.} Success rate (\%) on the 150 official evaluation
    episodes of each task. Baselines from \citet{wang2026dexjoco},
    \citet{park2026pointwam} (GR00T N1.6, Fast-WAM, PointWAM), \citet{zhang2026me}, and
    \citet{li2026deco}. DP-T and DP-C: Diffusion Policy with transformer and CNN backbones.
    $^\dagger$One policy for all tasks. $^\ddagger$Mean over the ten tasks other
    than iPad. $^\S$With simulated tactile input. --: not reported.}
    \label{tab:dexjoco}
    \begin{tabular}{@{}lccccccccccccc@{}}
        \toprule
        & & \multicolumn{6}{c}{Single-arm} & \multicolumn{5}{c}{Bimanual} & \\
        \cmidrule(lr){3-8}\cmidrule(lr){9-13}
        \textbf{Method} & \textbf{Budget} & Hammer & Mouse & Bucket & Tongs & Glasses & Water & iPad & Hanoi
            & Assembly & Microwave & Photo & \textbf{Avg.} \\
        \midrule
        DP-T  & -- & 81.3 & 62.0 & 83.3 & 22.7 & 53.3 & 84.0 & 8.0  & 24.7 & 4.7  & 73.3 & 56.7 & 50.4 \\
        DP-C  & -- & 58.7 & 74.0 & 70.0 & 57.3 & 54.0 & 63.3 & 52.0 & 12.7 & 3.3  & 54.0 & 24.0 & 47.6 \\
        ACT        & -- & 50.0 & 61.3 & 64.0 & 31.3 & 47.3 & 47.3 & 9.3  & 6.0  & 0.0  & 66.0 & 7.3  & 35.5 \\
        $\pi_{0.5}$ & -- & 84.7 & 64.7 & 84.0 & 24.0 & 72.0 & 88.7 & 36.0 & 23.3 & 2.7 & 60.0 & 54.7 & 54.1 \\
        GR00T N1.5 & -- & 67.3 & 85.3 & 72.0 & 12.7 & 27.3 & 72.7 & 12.7 & 0.7 & 0.7 & 50.7 & 40.7 & 40.2 \\
        GR00T N1.6 & -- & 58.0 & 66.0 & 94.0 & 33.3 & 51.3 & 77.3 & -- & 24.7 & 4.0 & 76.7 & 30.0 & 51.5$^\ddagger$ \\
        DECO        & -- & 68.0 & 90.0 & 80.0 & 46.0 & 50.0 & 88.0 & 40.0 & 18.0 & 0.0  & 64.0 & 76.0 & 56.4 \\
        \midrule
        Fast-WAM$^\dagger$ & -- & 43.3 & 77.3 & 81.3 & 47.3 & 62.7 & 42.7 & -- & 42.0 & 2.7 & 88.7 & 40.7 & 52.9$^\ddagger$ \\
        PointWAM & -- & 82.0 & 89.3 & 96.7 & 50.0 & 62.0 & 94.7 & -- & 36.0 & 11.3 & 72.0 & 24.0 & 61.8$^\ddagger$ \\
        ME-Dex$^{\dagger\S}$ & -- & 86.0 & 76.0 & 92.0 & 92.0 & 56.0 & 72.0 & 48.0 & 32.0 & 28.0 & 88.0 & 48.0 & 65.3 \\
        \midrule
        \multirow{5}{*}{\rax}
            & 2  & 98.7 & \textbf{100.0} & 0.0 & 36.7 & 0.7 & 0.0 & \textbf{100.0} & 0.0 & 14.7 & 0.0 & 0.0 & 31.9 \\
            & 4  & 98.7 & 97.3 & 93.3 & 2.7 & 0.0 & 0.0 & \textbf{100.0} & \textbf{100.0} & 0.7 & 0.0 & 0.0 & 44.8 \\
            & 8  & 94.0 & \textbf{100.0} & \textbf{100.0} & \textbf{100.0} & \textbf{100.0} & 91.3 & \textbf{100.0} & \textbf{100.0} & 0.0 & 84.0 & 86.7 & 86.9 \\
            & 16 & 98.0 & \textbf{100.0} & 96.7 & 99.3 & 55.3 & 98.7 & \textbf{100.0} & 98.7 & \textbf{96.0} & 93.3 & 68.7 & 91.3 \\
            & 32 & \textbf{100.0} & \textbf{100.0} & \textbf{100.0} & \textbf{100.0} & 93.3 & \textbf{99.3} & \textbf{100.0} & \textbf{100.0} & 72.0 & \textbf{99.3} & \textbf{100.0} & \textbf{96.7} \\
        \bottomrule
    \end{tabular}
\end{table}

\subsubsection{LIBERO and LIBERO-Pro}\label{sec:exp-libero}

\paragraph{Setup.} LIBERO~\citep{liu2023libero} evaluates a single Franka
Panda arm with a parallel-jaw gripper on four ten-task suites, Spatial, Object, Goal, and Long
(\Cref{app:libero}),
and LIBERO-Pro~\citep{zhou2025libero} perturbs the layouts of the same tasks. The agent
commands delta end-effector poses and the gripper. It perceives the scene
through the agent-view and wrist RGB-D cameras and receives the task description and three
images of the completed task, but no action demonstrations. It explores with a budget of up to 20 episodes on fresh layouts drawn from the official
placement distribution and may continue with up to 5 episodes of adaptation on layouts drawn from
the position-perturbed distribution of LIBERO-Pro, in which objects exchange their initial regions
(\Cref{app:adaptation}), never on an evaluation state. The resulting fully automated program is
evaluated alone, without the agent, on unseen scenes, the 50 official initial states of each task
and the 50 position-perturbed states of LIBERO-Pro.

\paragraph{Baselines.} We
compare with the VLAs OpenVLA~\citep{kim2024openvla}, OpenVLA-OFT~\citep{kim2025fine},
$\pi_0$~\citep{black2024pi_0}, $\pi_{0.5}$~\citep{intelligence2025pi_}, and
X-VLA~\citep{zheng2026x}, and with the world action models Cosmos
Policy~\citep{kim2026cosmos}, Fast-WAM~\citep{yuan2026fast}, PatchWAM~\citep{wang2026action},
and VPP2~\citep{guo2026video}, all of which fine-tune on the benchmark's demonstrations
(\Cref{tab:libero}).

\paragraph{Results.} Without a single action demonstration, \rax achieves one of the highest average
success rates among these models on the original suites, 98.5\% (\Cref{tab:libero}). Under position perturbation the
learned policies collapse: OpenVLA, $\pi_0$, and X-VLA succeed in at most 2\% of the episodes
of any suite, $\pi_{0.5}$ in at most 38\%, and the world action models PatchWAM and VPP2 in
8.8\% and 42.3\% on average. Without a single episode on perturbed layouts, the programs of
\rax, which locate every object anew in each episode, already retain 76.0\% on average and
exceed every learned policy on every suite, 33.7 points above VPP2; they also exceed by 12.8
points the coding agent ASPIRE~\citep{lu2026aspire}, which develops a program for each task and
perturbation with up to 75 rollouts and evolutionary search on instances from the evaluation distribution
(\Cref{app:aspire}). With at most 5 episodes of adaptation on perturbed layouts, \rax raises
its average by a further 8.4 points to 84.4\%, from 71.4\% on Long to 100.0\% on Object, and
exceeds ASPIRE on every suite. Exploration thus yields programs that transfer to new layouts,
and a few further episodes of exploration repair those that do not.

\begin{table}[htb]
    \centering
    \small
    \setlength{\tabcolsep}{3pt}
    \caption{\textbf{LIBERO and LIBERO-Pro.} Success rate (\%) on the 50 official initial states of
    each task and on the 50 position-perturbed states of LIBERO-Pro. Baselines from their papers,
    openpi ($\pi_0$, $\pi_{0.5}$), and the LIBERO-Pro leaderboard~\citep{zhou2025libero};
    PatchWAM reports only the mean, and VPP2 uses 10 initial states per task. $^\dagger$Developed
    on the perturbed setting under evaluation (\Cref{app:aspire}). $^\ddagger$Zero-shot with a
    skill library built on LIBERO-90; included in the average. --: not reported.}
    \label{tab:libero}
    \begin{tabular}{@{}lccccccccccc@{}}
        \toprule
        & & \multicolumn{5}{c}{LIBERO} & \multicolumn{5}{c}{LIBERO-Pro, position perturbation} \\
        \cmidrule(lr){3-7}\cmidrule(lr){8-12}
        \textbf{Method} & \textbf{Demos} & Spatial & Object & Goal & Long & \textbf{Avg.} & Spatial & Object & Goal & Long & \textbf{Avg.} \\
        \midrule
        OpenVLA      & 50 & 84.7 & 88.4 & 79.2 & 53.7 & 76.5 & 0.0  & 0.0  & 0.0  & 0.0 & 0.0 \\
        OpenVLA-OFT  & 50 & 97.6 & 98.4 & 97.9 & 94.5 & 97.1 & --   & --   & --   & --  & -- \\
        $\pi_0$      & 50 & 96.8 & 98.8 & 95.8 & 85.2 & 94.2 & 0.0  & 0.0  & 0.0  & 0.0 & 0.0 \\
        $\pi_{0.5}$  & 50 & 98.8 & 98.2 & 98.0 & 92.4 & 96.9 & 20.0 & 17.0 & 38.0 & 8.0 & 20.8 \\
        X-VLA        & 50 & 98.2 & 98.6 & 97.8 & 97.6 & 98.1 & 0.0  & 2.0  & 1.0  & 0.0 & 0.8 \\
        \midrule
        Cosmos Policy & 50 & 98.1 & \textbf{100.0} & 98.2 & 97.6 & 98.5 & -- & -- & -- & -- & -- \\
        Fast-WAM      & 50 & 98.2 & \textbf{100.0} & 97.0 & 95.2 & 97.6 & -- & -- & -- & -- & -- \\
        PatchWAM      & 50 & 96.5 & \textbf{100.0} & \textbf{99.0} & \textbf{98.0} & 98.4 & -- & -- & -- & -- & 8.8 \\
        VPP2          & 50 & \textbf{99.2} & 99.8 & 98.2 & \textbf{98.0} & \textbf{98.8} & 54.0 & 58.0 & 37.0 & 20.0 & 42.3 \\
        \midrule
        ASPIRE$^\dagger$ & 0 & -- & -- & -- & -- & -- & 51.0 & 98.0 & 81.0 & 22.6$^\ddagger$ & 63.2$^\ddagger$ \\
        \midrule
        \rax, no adaptation & 0  & \textbf{99.2} & \textbf{100.0} & 98.6 & 96.2 & 98.5 & \textbf{79.2} & \textbf{100.0}
            & 73.4 & 51.2 & 76.0 \\
        \rax, adaptation    & 0  & \textbf{99.2} & \textbf{100.0} & 98.6 & 96.2 & 98.5 & \textbf{79.2} & \textbf{100.0}
            & \textbf{87.0} & \textbf{71.4} & \textbf{84.4} \\
        \bottomrule
    \end{tabular}
\end{table}


\subsection{Ablations}
\label{sec:exp-ablations}
In this section we ablate the key design choices of our system, including parallel exploration with multi-agent and multiple environments, the effect of depth sensing over RGB-only observations, and the evolve of inference tokens and inference time over exploration.
\subsubsection{Parallelization}\label{sec:exp-parallel}

Exploration can also be parallelized
(\Cref{sec:parallel}): the same budget of
exploration episodes is distributed across $K$ parallel environments, explored by a team of
cooperating agents that share a common workspace. In the same way that multiple agents
scale test-time compute in parallel rather than purely serially, trading some efficiency for
wall-clock time, parallel environments let a team of agents consume an exploration budget
concurrently. We parallelize exploration on \textit{make
toast}. All sessions use RGB-D observations, an image sequence of one successful completion,
and instructions that summarize the official task code, including its numeric success
thresholds; the fully automated program is evaluated on the same 32 unseen scenes. Parallelization
reduces the time needed to consume the exploration budget for both agents
(\Cref{tab:parallel}). With four agents in four environments, Claude Code reaches a score of
92.2 in 4.9 hours, against 63.3 in 13.7 hours for a single agent with a larger budget; with
two agents in two environments, Codex matches its single-agent score in 9.2 hours instead of
14.2.

\begin{table}[htb]
    \centering
    \small
    \caption{\textbf{Parallel exploration on make toast.} Agent time until submission, and success
    rate and score (\%) of the fully automated program on 32 unseen scenes.}
    \label{tab:parallel}
    \begin{tabular}{@{}lcccccc@{}}
        \toprule
        \textbf{Agent} & \textbf{Agents} & \textbf{Envs} & \textbf{Budget} & \textbf{Time (h)}
            & \textbf{Success} & \textbf{Score} \\
        \midrule
        Codex       & 1 & 1 & 32 & 14.2 & 40.6 & 66.4 \\
        Codex       & 2 & 2 & 32 & 9.2  & 56.3 & 65.6 \\
        \midrule
        Claude Code & 1 & 1 & 48 & 13.7 & 50.0 & 63.3 \\
        Claude Code & 4 & 4 & 32 & 4.9  & 87.5 & 92.2 \\
        \bottomrule
    \end{tabular}
\end{table}

\subsubsection{Depth versus RGB-Only Perception}\label{sec:exp-depth}

To measure the contribution of depth, we compare RGB-D and RGB-only sessions on the three
domain-randomized RoboDojo tasks. In the RGB-only profile, depth
and the geometric primitives built on it are removed, and the agent must perceive through
motion, relating image coordinates to robot motion by moving the robot by known amounts.
Each pair shares the task, the agent, and the exploration budget $N$, and its
fully automated program is evaluated on the same unseen scenes as in \Cref{sec:exp-scaling}.
With depth, the agent outperforms the RGB-only agent in every pair, by 17.3 points on
average (\Cref{tab:depth}).

\begin{table}[htb]
    \centering
    \small
    \caption{\textbf{Depth versus RGB-only perception.} Score (\%) on unseen layouts at matched
    exploration budgets. Cells with more than one run report the mean.}
    \label{tab:depth}
    \begin{tabular}{@{}lcccc@{}}
        \toprule
        \textbf{Task} & \textbf{Budget} & \textbf{RGB-D} & \textbf{RGB-only} & $\Delta$ \\
        \midrule
        Make toast random   & 16 & \textbf{2.5}  & 0.0  & $+$2.5  \\
        Make toast random   & 32 & \textbf{31.7} & 7.8  & $+$23.9 \\
        Hang mugs random    & 16 & \textbf{36.9} & 13.9 & $+$23.0 \\
        Sweep blocks random & 16 & \textbf{53.3} & 33.3 & $+$20.0 \\
        \midrule
        Average      &    & \textbf{31.1} & 13.8 & $+$17.3 \\
        \bottomrule
    \end{tabular}
\end{table}

\paragraph{Case study: perception with and without depth.}\label{sec:exp-perception-case}
To understand where this gap comes from, we inspect the fully automated programs submitted
by the RGB-D and RGB-only agents on make toast random with a budget of 32 episodes. Both agents use the
vision-language model in the same role, to point at semantic image locations such as the
toaster slots, the top crust of a bread slice, and the lever. They differ in how they turn
these image points into motion.

\textit{With depth.} The RGB-D agent lifts every image point into a metric 3D point and
builds the task's geometry from these points. It first refines each slot opening by
fitting a line to its visible contour, then measures both rim endpoints of each slot in 3D,
from which it derives the slot centers and their common heading. On the bread, it measures
two points along the top crust and one on the near side, and recovers the slice's full 3D
orientation from the plane through these points. After the grasp, it measures the slice
again in the wrist camera to obtain its pose in the hand, and repeats this measurement above
the slot to correct any slip before release. The agent also uses metric geometry to verify
its perception: a measurement is rejected when a rim point does not lie within a narrow band
at the toaster's roof height, when the two rim endpoints are not 9 to 20\,cm apart, or when
the two slots are not 2.6 to 7.5\,cm apart, and a press of the lever counts only when the
lever is measured to have travelled 2.5 to 13\,cm downward.

\textit{Without depth.} The RGB-only agent does not give up on 3D reasoning; it
reconstructs a coarse camera model of its own. It maps overhead-camera pixels to table
coordinates through a linear calibration at an assumed height, about 1.8\,mm per pixel
across the image and 1.4\,mm per pixel along it, and then refines every target by visual
servoing in the wrist camera: it moves the gripper until the target is centered, with
hand-tuned gains of 0.4 to 0.8\,mm per pixel, and rotates it until the slot or the bread is
aligned in the image. The quantities that depth would provide are replaced by constants,
including the grasp and release heights, the 3.8\,cm spacing between the slots, and the
height to which the lever is pressed, and success is judged by asking the vision-language
model whether a slice is held or seated. These approximations hold only where the scene
matches the agent's assumptions. In evaluation, 36 of its 45 episodes ended in the lever
stage without success, the step at which a fixed press height replaces a measured one,
against 8 successes for the RGB-D agent.


\subsubsection{Inference Time and Tokens}\label{sec:exp-code}
\begin{figure}[htb]
    \centering
    \includegraphics[width=\linewidth]{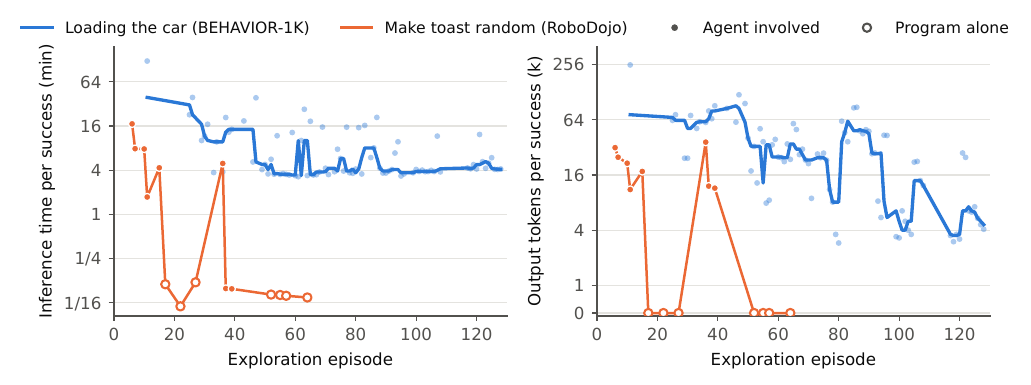}
    \caption{\textbf{From exploration to code.} Inference time (left) and output tokens
    (right) of each successful exploration episode, on base-2 logarithmic vertical axes; for
    rehearsals, inference time is the program's computation time as measured by the harness,
    without the overhead of running in isolation. The token
    axis is linear below one thousand tokens so that zero can be shown. Filled dots mark
    episodes in which the agent takes part, alone or together with its program; hollow circles
    mark rehearsals, in which the program runs alone. On loading the car, raw points are
    faint and the line is a
    running median over five successes; output tokens of parallel episodes are attributed to the environment an agent was assigned, or split evenly across the
    episodes active at that moment.}
    \label{fig:inference}
\end{figure}

Here we demonstrate the decrease of inference wall time and output tokens with exploration as more and more decisions are being automated with programs.
\paragraph{Simulation} As the agent compiles its experience into autonomous code programs, the computational overhead of action generation decreases substantially. We evaluate this efficiency gain across two benchmark tasks: \textit{loading the car} in BEHAVIOR-1K (128 exploration episodes across two parallel environments) and \textit{make toast random} in RoboDojo (64 exploration episodes). We define episode \emph{inference time} as the cumulative wall-clock latency dedicated to action generation---excluding simulator and hardware step times---regardless of whether actions stem from direct LLM deliberation, code program execution, or a hybrid mode. \Cref{fig:inference} tracks inference latency and output token consumption across all successful exploration episodes. On \textit{loading the car}, the initial successful episode required 123 minutes of inference time and 254k output tokens. By the final exploration episode, inference overhead dropped to 4 minutes and 4k tokens, while environment execution time remained over 1 hour per episode. Similarly, on \textit{make toast random}, initial task completion required 17 minutes of inference latency. Once execution transitioned to the compiled program, inference latency for the final agent-monitored successes fell to 6 seconds per episode---matching the 3--7 seconds observed during standalone program rehearsals (which accounted for 7 of the 15 total successes).

\paragraph{Real robots} Similar efficiency trends hold on real robots using the Franka manipulator (\Cref{tab:inference-real}). We evaluate air fryer manipulation (\Cref{app:real-airfry}) and Bandu tower assembly (\Cref{app:real-bandu}), comparing an initial trial in which the agent generates each action via active deliberation against a subsequent execution of the compiled autonomous program. In both paradigms, the system actuates the robot through identical tool interfaces, with robot time defined as the duration spent executing arm, gripper, and visual perception calls. For the air fryer task, initial task completion (including online exploration) required 72 minutes of inference latency and 262k output tokens, against 3 minutes of robot time. Upon compilation, the initial autonomous run required only 18 seconds of inference latency (attributable entirely to local programmatic execution), with four subsequent unattended trials consistently yielding latencies between 18 and 23 seconds. Similarly, on the Bandu tower, step-by-step LLM deliberation to assemble six blocks required 17 minutes of inference latency; once compiled, autonomous assembly of seven blocks completed with 10 seconds of program overhead alongside 11 minutes of continuous robot execution. During these autonomous runs, control proceeded without human intervention or online LLM action synthesis, while background monitoring generated at most 2.2k tokens per run in parallel without imposing latency on robot execution.

\begin{table}[htb]
    \centering
    \small
    \caption{\textbf{From the agent's actions to a fully automated program on the real Franka.} The
    first success with actions generated by the agent and a later unattended success of its program.
    Inference time: wall-clock time outside robot calls. The agent's Bandu rebuild places six pieces,
    the program seven.}
    \label{tab:inference-real}
    \begin{tabular}{@{}llrrr@{}}
        \toprule
        \textbf{Task} & \textbf{Actions from} & \textbf{Inference} & \textbf{Robot time}
            & \textbf{Output tokens} \\
        \midrule
        Air fryer   & Agent   & 72.3\,min & 2.9\,min  & 262.4\,k \\
                    & Program & 18\,s     & 2.5\,min  & 2.0\,k   \\
        \midrule
        Bandu tower & Agent   & 17.2\,min & 9.3\,min  & 88.4\,k  \\
                    & Program & 10\,s     & 10.9\,min & 2.2\,k   \\
        \bottomrule
    \end{tabular}
\end{table}


\subsubsection{Demonstrations Without Actions}\label{sec:exp-noaction}

On HumanoidArena, the exploring agent reads three teleoperated demonstrations of each task, and
its programs may reuse motion segments cut from them, such as a gait cycle
(\Cref{app:doubledesk}). We remove the actions from these demonstrations and keep everything
else unchanged except the budget (\rax$^\star$), which is 256 exploration episodes on DoubleDesk, 128 on
P\&PBox, and 64 on each of the other five tasks, and evaluate the program on the same 60 official
test episodes. Without any action data, \rax$^\star$ reaches 81.9\% on average and matches
WB-WAM, which is fine-tuned on 100 action demonstrations per task
(\Cref{tab:humanoidarena-noaction}). On Football, OpenDoor, and SitSofa, it stays within 5 points
of \rax with action demonstrations. The more dynamic and contact-rich whole-body tasks, which
require sustained locomotion or whole-body contact, VisNavi, Boxing, P\&PBox, and DoubleDesk,
remain 10.0 to 16.6 points lower despite larger budgets. When the agent has to discover a
suitable gait and whole-body control on its own, whole-body loco-manipulation requires more
exploration.

\begin{table}[htb]
    \centering
    \small
    \setlength{\tabcolsep}{3pt}
    \caption{\textbf{HumanoidArena without action demonstrations.} Success rate (\%) on the 60
    official test episodes of each task. \rax: largest budgets of \Cref{tab:humanoidarena}.
    $^\star$Three demonstrations without actions.}
    \label{tab:humanoidarena-noaction}
    \begin{tabular}{@{}lcccccccccc@{}}
        \toprule
        \textbf{Method} & \textbf{Demos} & \textbf{Budget} & Football & DoubleDesk & P\&PBox & OpenDoor
            & SitSofa & Boxing & VisNavi & \textbf{Avg.} \\
        \midrule
        WB-WAM        & 100 & --         & 70.0 & \textbf{65.0} & \textbf{86.7} & 98.3 & 95.0 & 81.7 & 76.7 & 81.9 \\
        \rax          & 3   & 32--128    & \textbf{98.3} & 56.7 & 83.3 & \textbf{100.0} & 96.7 & \textbf{100.0} & \textbf{93.3} & \textbf{89.8} \\
        \rax$^\star$  & 3   & 64--256    & \textbf{98.3} & 46.7 & 70.0 & 95.0 & \textbf{100.0} & 86.7 & 76.7 & 81.9 \\
        \bottomrule
    \end{tabular}
\end{table}

\subsection{Intriguing Emerging Agentic Behaviors with Scaling Exploration}
In this section, we analyze agentic behaviors that emerge as exploration scales, in simulation and on
real robots; each task is detailed in \Cref{app:sim-cases,app:real}.

\subsubsection{Active Perception}
\label{sec:exp-active}
\begin{figure}[htb]
    \centering
    \includegraphics[width=\linewidth]{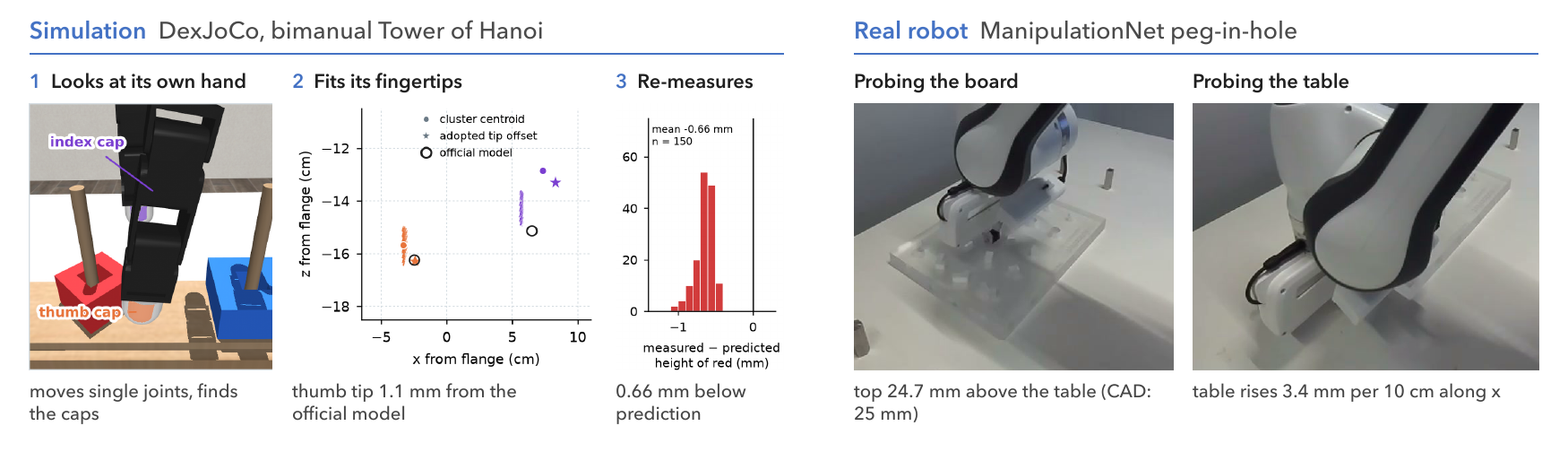}
    \caption{\textbf{Active perception.} Simulation, DexJoCo bimanual Tower of Hanoi
    (\Cref{app:hanoi}): (1) in its first episode, the agent moves single joints of the Allegro hand and
    finds the white fingertip caps in the wrist camera; (2) from the caps' points it fits the
    fingertip offsets relative to the flange, and, without a model of the hand, its thumb tip lies
    1.1\,mm from the official model; (3) after each placement, the program measures the disk again,
    on average 0.66\,mm below the height it predicted over 150 evaluation episodes. Real robot,
    ManipulationNet peg-in-hole (\Cref{app:real-peg}): the agent probes with closed fingers to
    measure the height of the transparent board and the tilt of the table.}
    \label{fig:active}
\end{figure}

Active perception~\citep{xiong2025vision} emerges naturally as the agent does not treat perception as a single look at the scene (\Cref{fig:active}). It moves its cameras and its
hand to obtain the measurements that a step needs, measures again before it acts, and checks the
outcome of each action with a new measurement.

\paragraph{Simulation.} On the bimanual Tower of Hanoi of DexJoCo (\Cref{app:hanoi}), the agent
received no model of its Allegro hands. It spent its first episode on its own hand: it moved single
joints, lifted the wrist cameras' depth into points to see what each joint moves, segmented the
white fingertip caps, and fitted the fingertip offsets on which every grasp of its program is built;
its thumb offset lies within 1.1\,mm of the official model it never saw. On the LIBERO-Long task of putting two cans into a basket
(\Cref{app:libero-basket}), the program perceives the scene anew in every round; before each
grasp, it hovers 7\,cm above the can, refits the lid from the wrist camera, and rejects a top that
is not can-shaped. Because the agent-view camera cannot see the floor of the basket, the program
maps the basket from the wrist camera while it holds the first can, and maps it again after the
first release. On the bimanual Tower of Hanoi of DexJoCo (\Cref{app:hanoi}), the program verifies
every lift by measuring the held disk from three cameras and pinches again when the disk rose less
than half of the lift; after placing the red disk, it measures the disk's top again and finds it on
average 0.66\,mm below the height it predicted. On the humanoid that carries a hammer over a bar
(\Cref{app:doubledesk}), the program perceives anew at every stage and before every step at the
bar, and it confirms a grasp by measuring the height of the handle again. Without depth, the
RGB-only agent recovers metric relations by moving the robot by known amounts and servoing on the
wrist camera (\Cref{sec:exp-depth}).

\paragraph{Real robot.} On the ManipulationNet peg-in-hole challenge~\citep{chen2026manipulationnet}
(\Cref{app:real-peg}), the board is transparent acrylic, and the task text warned that the wrist
camera's depth is inaccurate at close range. The agent turned this warning into a measurement
procedure that uses the arm itself as a probe: it closes the gripper, descends in 1.5\,mm steps,
and records the height at which the vertical force drops by 3 to 4\,N. Touching the table at three
points, it found that the table rises by about 3.4\,mm per 10\,cm along the robot's $x$ axis and
that the top of the board lies 24.7\,mm above the table, against 25\,mm in the board's CAD model,
and it models the height of the board's top as a plane that rises along $x$. Before each insertion
into the row with 0.1\,mm clearance, it touches the board next to the hole with the held peg, and
these reference heights agree within 0.5\,mm. Probing also calibrates the camera: from a hole whose
centre it had located by touch, the agent found that the wrist camera's absolute positions were
offset by 6.4 and $-0.8$\,mm, corrected every hole by this offset, and the next peg dropped into its
hole on the first probe.

\begin{figure}[htb]
    \centering
    \includegraphics[width=\linewidth]{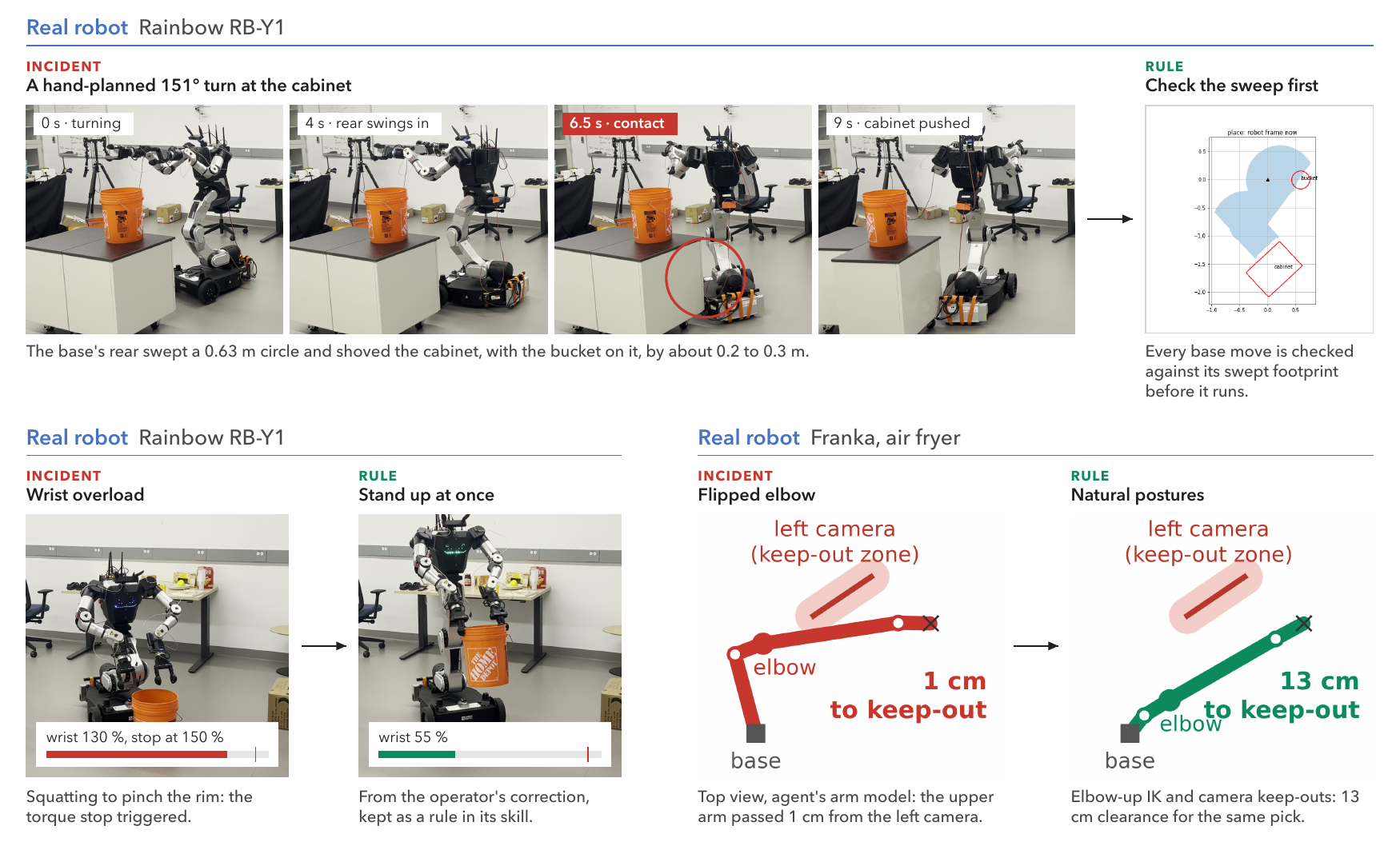}
    \caption{\textbf{Error recovery and safety on real robots.} Each incident and the rule the agent
    derived from it. RB-Y1: a turn planned by hand pushes a cabinet, after which every base move is
    checked against the area the base sweeps; squatting to lift a bucket overloads a wrist, and
    standing up first resolves it. Franka: an unusual arm posture hits a camera, after which the
    agent plans only natural postures that keep the arm away from the cameras.}
    \label{fig:safety}

\end{figure}

\begin{figure}[htb]
    \centering
    \includegraphics[width=0.92\linewidth]{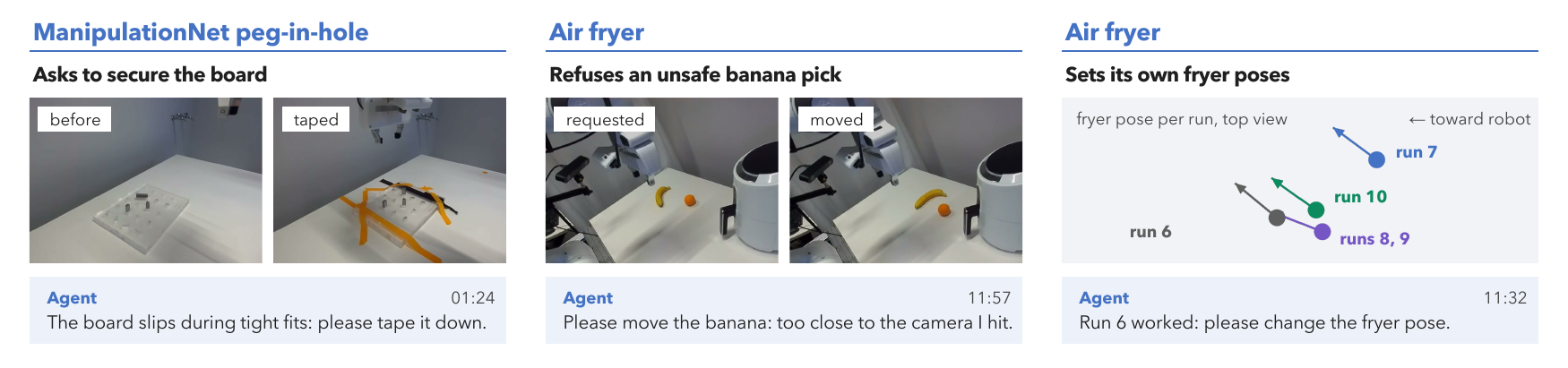}
    \caption{\textbf{Human--robot interaction.} Requests that the agent sent to the operator, summarized,
    with the scene before and after. Right: the fryer poses
    that the agent requested across runs 6 to 10, from its own measurements of the fryer's axis and
    handle direction.}
    \label{fig:hri}
\end{figure}

\subsubsection{Error Recovery and Safety}
\label{sec:exp-safety}
On real robots, the agent's mistakes have physical consequences, and it turns each one into a rule
that its later programs follow (\Cref{fig:safety}). On the RB-Y1, an in-place turn that the agent
planned by hand swung the rear of the base into a cabinet; from then on, the agent checked every
base move against the area the base sweeps. Lifting a bucket while squatting overloaded a wrist and
stopped the robot; standing up before the lift, a correction from the operator, resolved it, and
the agent kept it as a rule in its skill. On the Franka, an unusual arm posture drove the elbow
into a camera; the agent then restricted itself to natural arm postures, kept the arm away from
the cameras, and recalibrated the bumped camera itself (\Cref{app:perception}). In simulation, the
humanoid crosses a bar only from foot placements that its own falls showed to be safe
(\Cref{app:doubledesk}).

\begin{figure}[htb]
    \centering
    \includegraphics[width=\linewidth]{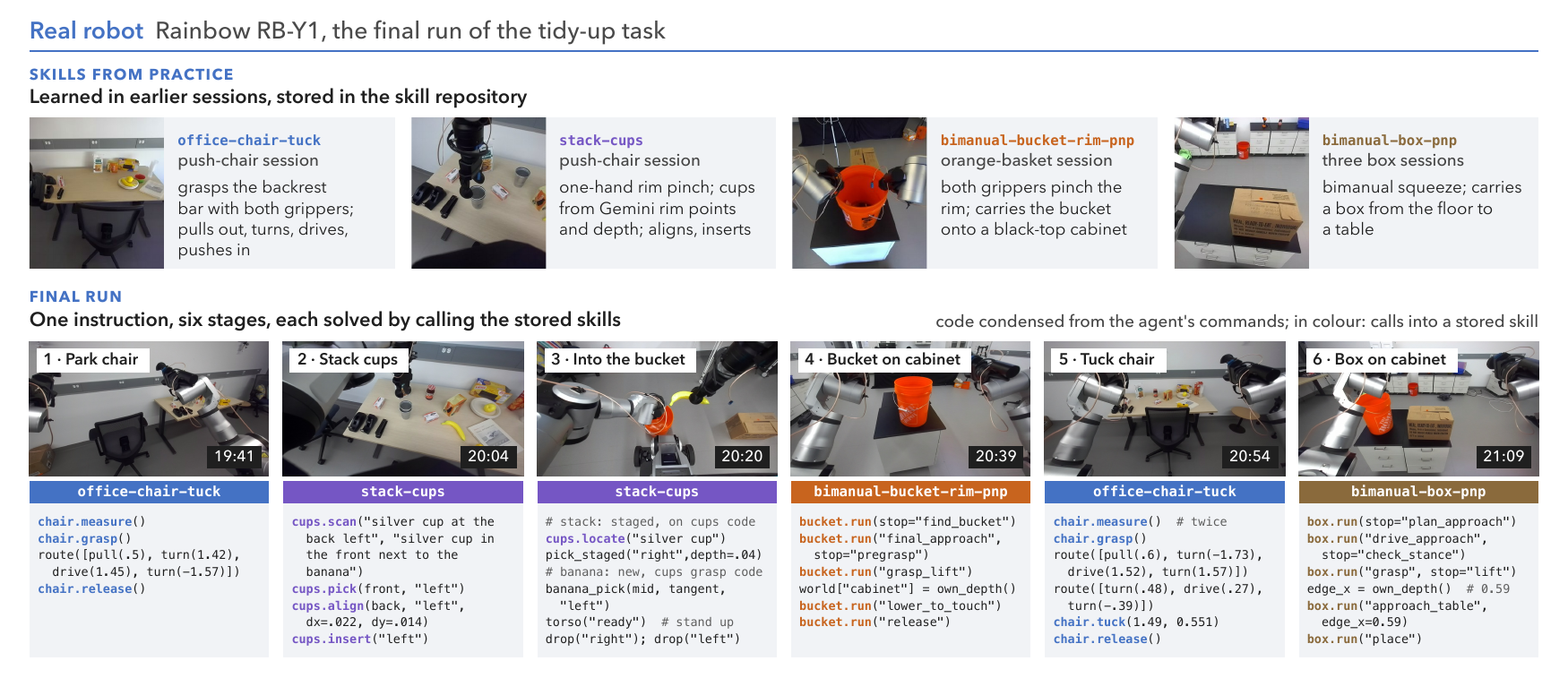}
    \caption{\textbf{Continual learning on the real-world RB-Y1 mobile manipulator.} \textbf{Top:} Modular skills acquired during prior exploration sessions and stored in the skill repository, complete with descriptions from their respective skill files. \textbf{Middle:} The final execution of a complex tidy-up task---a single high-level instruction carried out across six sequential stages, annotated with completion timestamps and invoked skills. \textbf{Bottom:} The agent's generated code for each stage (condensed from execution logs with passed arguments). Code invoking pre-stored skills is color-coded by skill (\texttt{chair}, \texttt{cups}, \texttt{bucket}, and \texttt{box} correspond to each skill's \texttt{run.py}), while ad-hoc control logic written directly by the agent is shown in black.}
    \label{fig:continual}
\end{figure}

\subsubsection{Human--Robot Interaction}
\label{sec:exp-hri}
The agent reaches the operator through a request tool and uses it with purpose (\Cref{fig:hri}). On
the ManipulationNet board, after the board shifted during its attempts at tight fits, it asked the
operator to tape the board to the table. On the air fryer (\Cref{app:real-airfry}), it refused to pick a banana placed in front
of the camera it had hit earlier, because its forearm would pass within about 1\,cm of the camera, and
asked for the banana to be moved. After unattended successes, it asked the operator to move and rotate
the fryer, in effect designing its own tests of generalization: across runs 6 to 10, the fryer moved
by up to 7\,cm and its handle turned by up to 18$^\circ$, and the program succeeded unattended in every
new placement but one, where it needed help and which exposed two errors in its program that the
agent then fixed.

\subsection{Continual Learning}
\label{sec:exp-continual}
In this section, we demonstrate how \rax achieves continual learning with robot programs: by reusing skills learned in earlier exploration sessions, and by rapidly adapting its programs when the environment or the task changes.

\subsubsection{Reusing Skills}
\label{sec:exp-reuse}
On the RB-Y1 mobile manipulator, \rax acquires skills through isolated exploration sessions and reuses them to solve long-horizon compositional tasks (\Cref{fig:continual}). In initial sessions, the agent independently learned to park and tuck an office chair, stack nested cups, transport a bucket by its rim onto a cabinet, and pick and place a box. Each skill was committed to the repository only after achieving successful end-to-end execution, stored as a versioned program bound to explicit precondition checks. The multi-stage tidy-up task (\Cref{fig:rby1}) evaluates the composition of these skills under a single high-level goal: park the chair, stack the cups, place the cups and a banana inside the bucket, transport the bucket onto the cabinet, tuck the chair back under the desk, and finally pick the box onto the other empty cabinet.

\paragraph{Skills are called, not relearned} In the final execution of the tidy-up task, every stage is resolved by composing stored skills from the repository (\Cref{fig:continual}). The agent executes the chair skill sequentially: it estimates the chair geometry, plans a collision-free motion path, grasps the backrest, navigates along the path, and releases the target. It invokes the cup-stacking skill using semantic visual grounding from the current scene, applying fine-grained closed-loop offsets prior to insertion. For the bucket and box skills, the agent parameterizes execution using real-time perception data, such as the estimated cabinet boundary derived from depth sensing. To bridge gaps between stored skills, the agent dynamically synthesizes ad-hoc code---such as extending the grasp planner to handle a banana, commanding torso elevation prior to transit, and re-estimating desk pose before final chair insertion.

\paragraph{Skills are refined in use} Across three iterations of the tidy-up task, the agent iteratively updated program versions rather than rewriting entire control loops. The bucket and box skills required minor programmatic adjustments ($\sim$25 out of more than 1200 lines) to integrate dynamic cabinet-edge measurements from onboard depth perception. Skills encountering greater environmental domain shift underwent larger revisions: following two failed grasp attempts, the agent updated the chair skill to grasp 4 cm deeper, while extending the cup-stacking skill to support picking additional object classes.

\begin{figure}[htb]
    \centering
    \includegraphics[width=\linewidth]{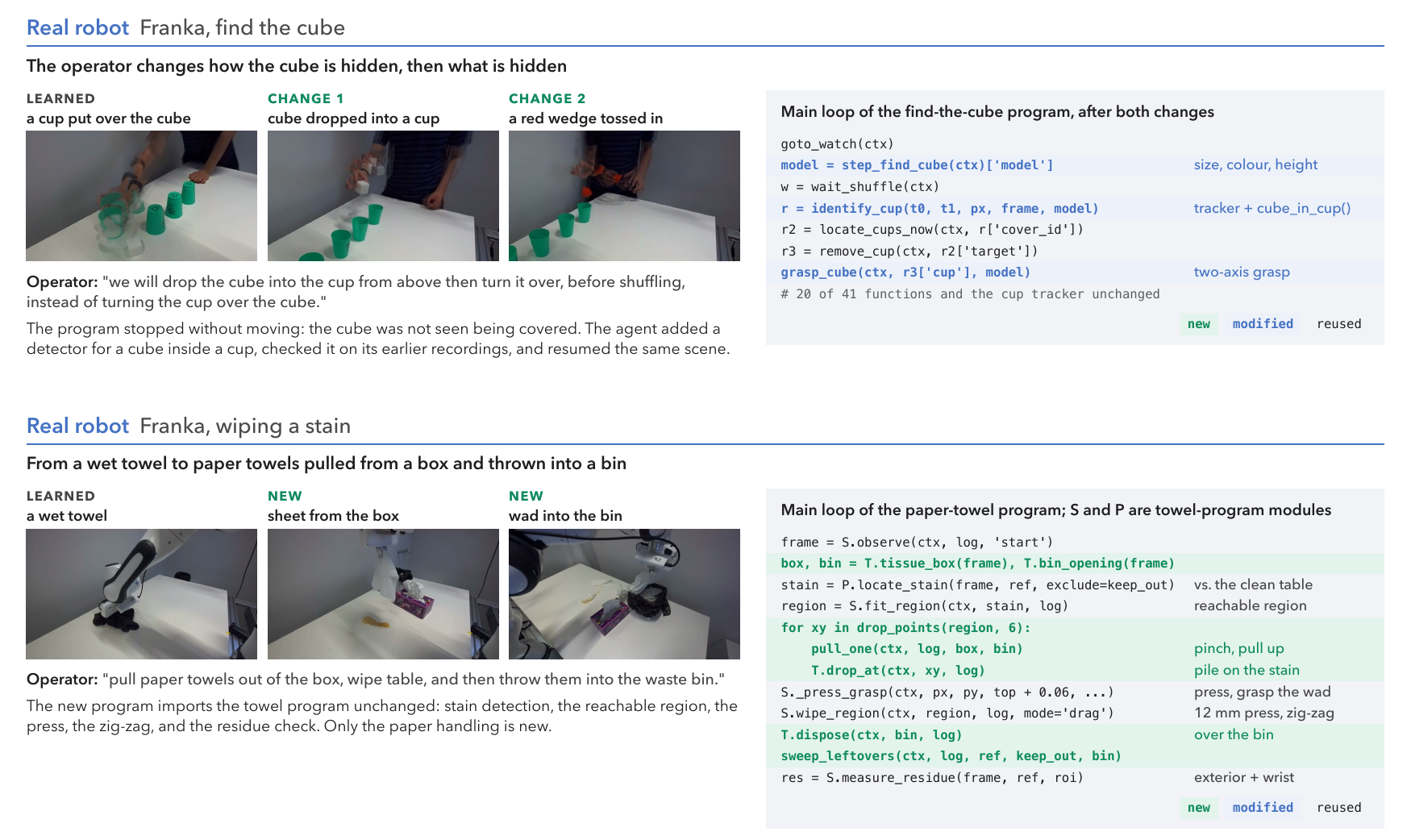}
    \caption{\textbf{Adapting skills and programs on the real Franka.} Each row is one program that already worked,
    with the task it learned (left image) and the operator's change (green). Right: the program's entry
    point after the change, condensed from the agent's code, with new lines in green, modified lines in
    blue, and reused lines in black. Top: find the cube, where each image overlays frames from the left
    camera a quarter of a second apart, earlier ones fainter, to show how the object is hidden; the cube
    is first dropped into a cup and then replaced by a red wedge. Bottom: wiping with paper towels from a box; \texttt{S} and \texttt{P} are
    the towel program's modules, imported unchanged.}
    \label{fig:adapt}
\end{figure}

\paragraph{Repository expansion} Intermediary control code synthesized by the agent to bridge task stages is modularized and saved as new skills. For instance, a synthesized \texttt{hand-reach} primitive was automatically abstracted, committed to the repository, and subsequently reused across both the chair-manipulation and cup-stacking routines.

\subsubsection{Rapid Adaptation of Programs}
\label{sec:exp-adapt}
When task conditions vary post-deployment, \rax preserves valid program subroutines and selectively updates only the modules invalidated by environmental shifts (\Cref{fig:adapt}). On the Franka manipulator, two end-to-end autonomous programs were adapted to novel task variants within the same session.

\paragraph{Find the cube} In a hidden-object task, the physical occlusion sequence was modified: rather than directly inverting a cup over the target cube, the cube was placed inside an upright cup prior to inversion. The original program detected no covering cup and safely halted execution via exception handling without initiating ungrounded motion. The agent extended its perception pipeline with a detector for cubes contained within upright cups, validated the new logic against current and historical sensor logs, and resumed execution to complete the task (\Cref{fig:real-cube}). Crucially, the cup-tracking, execution-monitoring, and removal routines were reused without modification. When substituting the cube with a red wedge, the agent generalized its geometric object representation from a fixed cube primitive to an object parameterized by estimated dimensions, color, and height, adjusting grasp synthesis accordingly while preserving 20 of the original 41 program functions (including the tracking pipeline; \Cref{fig:real-cube-wedge}).

\paragraph{Wiping with paper towels} Following the acquisition of a reliable surface-wiping skill using a wet cloth, the task was extended to require pulling disposable paper towels from a dispenser box, executing the wiping motion, and discarding used towels into a receptacle. The agent synthesized a composite program that imported the original wiping module as a dependency. The new program reused existing subroutines for stain segmentation against the table background, reachable workspace estimation, pressing by position control ($12\,\text{mm}$ below the measured surface contact, along a serpentine trajectory), and post-wipe residue verification. Only paper-handling subroutines were added: locating the dispenser box and disposal bin, extracting individual sheets onto the stain, discarding the crumpled towel, and clearing loose paper from the workspace. The stain was removed in both wet runs, although leftover sheets still had to be cleared with supervised scripts (\Cref{app:real-wipe}, \Cref{fig:real-wipe-paper}).

\section{Discussion}
\label{sec:discussion}

Progress in robot learning has largely been driven by collecting more demonstrations and
training larger models. Our results show that a capable agent, when allowed to explore, continues to
improve with the experience it accumulates, without teleoperation and without updates to its
parameters. Because this form of learning relies on general-purpose agents whose capabilities are
advancing rapidly in other domains, every improvement in those agents translates directly into
improved robot learning.

The knowledge acquired by \rax agents takes the form of perception and control programs,
skills, and notes. In this form, knowledge executes at the rate of the robot, can be inspected
and verified, composes into longer behaviors, and transfers across tasks, and we
expect code to play the same role for learning from experience that network parameters play
for learning from data. Since the instructions, skills, and memory structures of \rax are
themselves code and text, a natural next step is to let agents improve the architecture itself
from experience, and to share verified skills across fleets of robots, so that a skill learned
by one robot becomes available to all.

\section{Acknowledgements}
We thank Qianqian Wang for the valuable discussions and help in mobile manipulation robot experiments.
\newpage
\bibliography{references}
\bibliographystyle{arxiv-numbered}

\clearpage
\beginappendix
\section{Interface Contracts}
\label{app:interface}

On the real robots the agent acts only through tools that wrap each vendor's control SDK, and its
own programs call the same tools. Every command is checked before the robot moves; workspace, joint
limit, or IK violations are refused, and motion works only while the operator has armed the robot.
\Cref{tab:real-tools} summarizes the tools; both robots also offer \texttt{status} and
\texttt{request\_operator}.

\begin{table}[htb]
    \centering
    \small
    \caption{\textbf{Tools on the real robots.}}
    \label{tab:real-tools}
    \begin{tabularx}{\linewidth}{@{}lXX@{}}
        \toprule
        & \textbf{Franka FR3} & \textbf{Rainbow RB-Y1} \\
        \midrule
        Perception & RGB-D images from two exterior and one wrist camera; pixel to 3D point
              & Head RGB-D images and gaze control; pixel to 3D point; a vision-language model \\
        Motion & Straight-line end-effector moves; joint moves with path checks
              & Joint, end-effector, and torso targets; IK over torso and both arms; base drive \\
        Gripper & Grasp to a width with a set force
              & Close with a current limit, or set an opening \\
        Model & None (the agent writes its own kinematics)
              & Forward kinematics; dry runs for collision, joint load, and center of mass \\
        \bottomrule
    \end{tabularx}
\end{table}

\section{Agent Instructions and Skills}
\label{app:skills}

Every session starts from the same workspace template: one instruction file and a folder of skills,
written once and shared by all platforms (\texttt{auto\_research\_agent/} and \texttt{agent\_skills/}
in our code). Robot-specific facts, such as tool names, cameras, and control rates, are filled in when
the workspace is built, so the agent reads the same instructions on every robot.

\paragraph{Instructions.} The instruction file fixes the order of work rather than how to solve any
task. The agent first reads the task text and its own notes from earlier sessions, then (1) solves the
task by controlling the robot directly through tools until it confirms success, (2) turns what it did
into a program whose steps check their preconditions, verify their results, and report states they
were not built for, (3) runs that program on every further exploration episode, stepping in where it falls
short and fixing the code, and (4) keeps its notes, code, and a hand-off note current, since only
those carry over to later sessions. In benchmark sessions it then freezes the program, rehearses it
once end to end without help, and submits it. The file also states the budget rules: new layouts
in every episode, so scene geometry is measured again rather than assumed, and a blocker is
reported rather than bypassed.

\paragraph{Skills.} Skills are short documents that the agent opens when it reaches the step they
cover. They are grouped by axis, task strategy, coordination, perception, and robot control, and a
session receives only the layers that match its robot and observation mode. The task and
coordination skills are robot-agnostic (\Cref{tab:skills}); the perception and control layers give
each robot's interface under the same names, so a task skill can point to, for example, the
robot's interface or its free-space motion without knowing which robot it runs on.

\begin{table}[htb]
    \centering
    \small
    \caption{\textbf{Robot-agnostic skills} given to every session.}
    \label{tab:skills}
    \begin{tabularx}{\linewidth}{@{}lX@{}}
        \toprule
        \textbf{Skill} & \textbf{What it asks of the agent} \\
        \midrule
        Manual robot control & Look first, explore with small moves, choose each action from the
            current observation, and confirm every condition of the task before handing off. \\
        In-context action learning & When images of a completed task are supplied, infer the goal
            and useful strategy from them and adapt it to the current layout, without copying actions. \\
        Experience memory & Record what worked right away, with the observable conditions under
            which it worked; retrieve it for similar situations and distill it as evidence grows. \\
        Autonomous control & Start from a confirmed manual success, turn it into observation-driven
            code that can be supervised and resumed, and test it across layouts. \\
        Transfer hand-off & Before exploration ends, write how each subtask and each object is
            handled, for a later session on new layouts. \\
        Multi-environment exploration & Run parallel environments, track their scenes separately,
            and share one episode budget. \\
        Multi-agent exploration & Decide whether to delegate; give each subagent its own
            environment and a complementary goal, and combine their findings and code. \\
        \bottomrule
    \end{tabularx}
\end{table}

\clearpage
\section{Benchmarks and Tasks}
\label{app:tasks}

This section gives the protocol that all simulation experiments share and then the setting of
each benchmark and its tasks. The real-robot tasks are described in \Cref{app:real}.

\subsection{Sessions}
\label{app:sessions}

\paragraph{Agents.} Every experiment is a session of one coding agent in its own command-line
harness: Claude Code with Claude Opus 5.5 or, on four RoboDojo tasks and one BEHAVIOR-1K task,
Codex with a GPT-6 model, at the xhigh reasoning effort unless stated otherwise. A session
provides one or more simulated environments; with more than one, the agent may start subagents
of the same model, which share its workspace and its budget. The weights of the model are never
updated.

\paragraph{Workspace.} The agent starts from an empty workspace. Its folders for code, notes, and
outputs are empty, and it holds only the instructions of the harness, the task description, the
contract of the tool interface, and the skills of the platform, composed for the embodiment and
the observation mode of the session (\Cref{app:skills}); the agent keeps no memory from earlier
sessions. It runs in a container without network access, with 16\,GB of memory and four CPUs,
and reaches the simulator only through its tools. Every program it registers runs in a fresh
container with 8\,GB of memory, four CPUs, and no network access.

\paragraph{Task information.} The task description is generated from the public definition of the
task; its content differs between benchmarks and is given below. Images of the completed task
are frames of successful official demonstrations, either the final frame of each or a sequence of
keyframes along each. They contain no actions and no states, and only the exploring agent sees
them, never its programs. During an interactive episode, the agent may ask the benchmark's
evaluator whether the task is complete.

\paragraph{Budget and scenes.} The exploration budget counts exploration episodes. A single ledger
charges every episode start and every rehearsal, across all environments and subagents, and the
agent sees how many episodes remain. The budget is an upper bound, since submission ends
exploration. Exploration scenes come in a fixed order, which the agent cannot choose, from the
training distribution of the benchmark, and evaluation scenes from its official evaluation set,
which exploration never uses.

\paragraph{Limits.} Each episode has the native step limit of the benchmark and a wall-clock limit,
which during an interactive episode also bounds the agent's own deliberation. The agent and its
programs may call a vision-language model, Gemini 3.8 Flash, with a budget of \$10 per session
that exploration and evaluation share.

\subsection{Rehearsal and Evaluation}
\label{app:evaluation}

\paragraph{Rehearsal.} When the agent considers its program ready, it freezes a copy of its
code with a registered entry point into a bundle and requests a rehearsal. The bundle is executed
alone in a fresh isolated environment on a new exploration scene while the agent is paused, and the
rehearsal consumes one exploration episode, like an episode in which the agent acts itself.
Afterwards, the agent reads the outcome and the logs of the rehearsal and may revise the bundle
and rehearse again. Only a bundle that has completed a rehearsal without error can be submitted,
and submission ends exploration.

\paragraph{Fully automated program.} Evaluation executes the submitted bundle on each unseen
scene in a fresh isolated environment, without the agent and without edits or retries. The
results are never revealed to the agent. Because the bundle calls the same tools that the agent
called during exploration, the program that is evaluated is the program that was rehearsed.

\paragraph{Agent-in-the-loop program.} At the end of exploration, the workspace is frozen,
including the code, the notes, the handoff, and a compressed version of the conversation. For
each evaluation scene, a fresh copy of the agent starts from this frozen workspace and attempts
the scene, running its code, observing, and acting through the same tools as during exploration.

\paragraph{Scoring.} Both forms are scored by the benchmark's native evaluator, with partial
credit wherever the benchmark defines it. An error, a timeout, or an early exit of the program
counts as a failure, and scores are averaged over all scheduled evaluation episodes.

\subsection{RoboDojo}
\label{app:robodojo}

\paragraph{Tasks and interface.} We use seven tasks of RoboDojo-Sim~\citep{chen2026robodojo} with the
dual-arm ARX X5 (\Cref{tab:robodojo-tasks}). The agent commands absolute joint targets at 25\,Hz,
six arm joints and a gripper opening for each arm, with between 1 and 50 rows per call. It may
also command bounded end-effector steps, plan collision-free motions of one arm against the robot
and the table, lift pixels to 3D points, and preview a motion on the robot alone. It observes the
scene through a high camera and a camera on each wrist at 640$\times$480; the benchmark renders
RGB, and the harness adds metric depth, the intrinsics, and the pose of every camera.

\paragraph{Task information.} The task description is the public scoring rubric of the task, which
for the randomized variants is that of the base task, with its score table. For make kong, pour
by language, and swap blocks it adds a short clarification of the conditions, and for plug in
charger and pour by language a summary of the official task code without its numeric thresholds.
The agent also receives one image of the completed task, the last frame of the official video of
the task.

\paragraph{Exploration and evaluation.} A single agent explores a single environment; on plug in
charger and pour by language it may start one subagent. Exploration scenes are the layouts of one
official collection in order, which repeat once the collection is exhausted, and the program is
evaluated on the first layouts of another collection (\Cref{tab:robodojo-tasks}); on hang mugs
random, both come from a pooled collection, with evaluation on layouts 0 to 49 and exploration on
layouts 50 to 89. An episode is limited to 6\,s of wall-clock time per allowed step. A score is
100 on native success and the native partial score otherwise. The parallel experiment
(\Cref{tab:parallel}) uses make toast with a description that includes the numeric thresholds,
keyframes of a completed run, and 32 evaluation layouts from a third collection; the RGB-only
sessions of \Cref{tab:depth} receive no depth, calibration, or pixel lifting.

\begin{table}[htb]
    \centering
    \small
    \caption{\textbf{RoboDojo tasks.} Step limit at 25\,Hz; partial credit: native score
    tiers; evaluation: number of official layouts.}
    \label{tab:robodojo-tasks}
    \begin{tabular}{@{}llcccl@{}}
        \toprule
        \textbf{Task} & \textbf{Benchmark id} & \textbf{Step limit} & \textbf{Partial credit}
            & \textbf{Evaluation} & \textbf{Agent} \\
        \midrule
        Make toast random   & \texttt{make\_toast\_random}   & 1400 & 25/50/100 & 45 & Codex \\
        Hang mugs random    & \texttt{hang\_mugs\_random}    & 800  & 15/40/100 & 50 & Claude Code \\
        Sweep blocks random & \texttt{sweep\_blocks\_random} & 1000 & --        & 30 & Codex \\
        Plug in charger     & \texttt{plug\_in\_charger}     & 400  & --        & 50 & Claude Code \\
        Swap blocks         & \texttt{swap\_blocks}          & 700  & --        & 50 & Codex \\
        Make kong           & \texttt{make\_kong}            & 600  & --        & 50 & Codex \\
        Pour by language    & \texttt{pour\_by\_language}    & 800  & 20/50/100 & 50 & Claude Code \\
        \bottomrule
    \end{tabular}
\end{table}

\subsection{BEHAVIOR-1K}
\label{app:behavior}

\paragraph{Tasks and interface.} We use six tasks of the 2025 BEHAVIOR
Challenge~\citep{li2023behavior} with the Galaxea R1Pro, in the official evaluation environment
of BEHAVIOR-1K (\Cref{tab:behavior-tasks}). The agent commands the native action at 30\,Hz, base
velocity together with absolute joint targets of the four trunk joints and the two seven-joint
arms and a command for each gripper, with between 1 and 150 rows per call. It may also move the
base in closed loop, step an end effector through inverse kinematics, plan collision-free arm
motions, lift pixels to 3D points, and preview a motion on the robot alone. It observes the scene
through the head camera at 720$\times$720 and a camera on each wrist at 480$\times$480, all RGB-D
with intrinsics and poses relative to the base, together with the official proprioception and an
odometry estimate integrated from the base velocity. The step limit of an episode is 1.5 times the
mean length of the human demonstrations of the task, as in the 2026 evaluation, whereas the 2025
challenge allowed twice the mean.

\paragraph{Task information.} The task description contains the official instruction, the goal and
initial conditions of the task's BDDL definition, the scene and its rooms, and the limits of an
episode. Except on boxing books, the agent also receives keyframes of successful official
demonstrations from the head camera: the first frame, the end of each annotated skill, the final
frame, and enough frames in between that no gap exceeds 10\,s.

\paragraph{Exploration and evaluation.} Exploration scenes are the 300 training instances of each task,
and evaluation uses the test instances 301 to 310 (\Cref{tab:behavior-tasks}). An episode is
limited to 900\,s plus 0.5\,s per allowed step. Boxing books is evaluated as a fully automated
program and the other five tasks as agent-in-the-loop programs, with one fresh copy of the agent
per test instance. On sorting items, clearing food, and clean up desk, each copy also receives the
reports that a review model, Gemini 3.8 Flash, wrote on the fully automated program's runs on the
other nine test instances, built from the final status of each goal condition, the partial score,
the program's log, and camera frames, and referring to objects only by their task names. Episodes
are scored by the Q-score of the challenge: 1 on success and otherwise the fraction of goal
conditions that become satisfied during the episode, maximized over the alternative groundings of
the goal and averaged over the ten instances.

\begin{table}[htb]
    \centering
    \small
    \setlength{\tabcolsep}{3.5pt}
    \caption{\textbf{BEHAVIOR-1K tasks and sessions.} Step limit at 30\,Hz; budget: exploration
    episodes allowed; envs: parallel environments; program: the program evaluated in \Cref{tab:behavior}. $^\ast$Maximum reasoning effort. $^\dagger$Without
    demonstration keyframes or the vision-language model; 32 episodes used.}
    \label{tab:behavior-tasks}
    \resizebox{\linewidth}{!}{%
    \begin{tabular}{@{}lllcccl@{}}
        \toprule
        \textbf{Task} & \textbf{Benchmark id} & \textbf{Agent} & \textbf{Step limit}
            & \textbf{Budget} & \textbf{Envs} & \textbf{Program} \\
        \midrule
        Boxing books      & \texttt{boxing\_books\_up\_for\_storage}          & Codex$^\dagger$ & 36{,}341 & 128 & 6 & fully automated \\
        Getting organized & \texttt{getting\_organized\_for\_work}            & Claude Code$^\ast$ & 23{,}506 & 128 & 2 & agent-in-the-loop \\
        Sorting items     & \texttt{sorting\_household\_items}                & Claude Code & 23{,}711 & 128 & 8 & agent-in-the-loop \\
        Clearing food     & \texttt{clearing\_food\_from\_table\_into\_fridge} & Claude Code & 19{,}602 & 128 & 4 & agent-in-the-loop \\
        Clean up desk     & \texttt{clean\_up\_your\_desk}                    & Claude Code & 32{,}126 & 128 & 4 & agent-in-the-loop \\
        Loading the car   & \texttt{loading\_the\_car}                        & Claude Code$^\ast$ & 28{,}839 & 128 & 2 & agent-in-the-loop \\
        \bottomrule
    \end{tabular}}
\end{table}

\subsection{HumanoidArena}
\label{app:humanoidarena}

\paragraph{Tasks and interface.} We use the seven tasks of HumanoidArena~\citep{wang2026humanoidarena}
that have official test episodes, with a Unitree G1 that has 29 body joints and Dex3 hands
(\Cref{tab:humanoidarena-tasks}). The agent commands the official 40-dimensional whole-body
reference at 50\,Hz, with between 1 and 100 frames per call: the root translation and height,
which the SONIC tracker~\citep{luo2026sonic} does not follow, since the robot translates only by
stepping, the root orientation relative to the initial heading, 29 joint targets, and an open or
close command for each hand. It may also plan collision-free motions of the arms, optionally with
the waist, and lift pixels to 3D points in the pelvis frame and in a gravity-aligned frame. It
observes the scene through the head camera at 640$\times$480 in RGB-D, with the intrinsics and the
camera pose in the pelvis frame, together with the root orientation, the joint positions and
velocities, and the state of the hands. An episode ends when the robot falls, by the rule of the
benchmark.

\paragraph{Task information.} The agent can read three teleoperated demonstrations of the task
 from the official raw dataset. Each is a complete episode with the 40-dimensional
references, joint states, root poses, and the head camera's RGB video and depth. The
demonstrations are mounted read-only for the exploring agent only; its programs may contain
motion segments that the agent cut from them (\Cref{app:doubledesk}).

\paragraph{Exploration and evaluation.} We use Claude Opus 5.5 as our agent. Exploration scenes use seeds derived from
a hash that excludes every official seed. Sessions with a budget above 32 explore two environments
and may start two subagents. An episode is limited to 600\,s plus 1\,s per allowed step. The
program is evaluated on the 60 official test episodes of each task, seeds 0, 1, and 2 with 20
episodes each, derived as in the official evaluation; every episode runs in a fresh simulator and
container, four in parallel, and the simulator advances only when the program sends actions.
Success is binary, and a fall or an error counts as a failure.

\begin{table}[htb]
    \centering
    \small
    \setlength{\tabcolsep}{4pt}
    \caption{\textbf{HumanoidArena tasks.} Step limit at 50\,Hz; budget: largest exploration
    budget in \Cref{tab:humanoidarena}. HOI and HSI: human--object and human--scene interaction.}
    \label{tab:humanoidarena-tasks}
    \begin{tabular}{@{}llcp{0.52\linewidth}c@{}}
        \toprule
        \textbf{Task} & \textbf{Type} & \textbf{Step limit} & \textbf{Success condition} & \textbf{Budget} \\
        \midrule
        Football   & HOI & 2000 & the ball passes the goal line between the posts & 32 \\
        DoubleDesk & HOI & 2000 & the hammer rests inside the basket on the second desk & 128 \\
        P\&PBox    & HOI & 1450 & the box rests on the shelf, inside its outline & 128 \\
        OpenDoor   & HSI & 1800 & the latch is released and the door is open by at least 60$^\circ$, robot standing & 32 \\
        SitSofa    & HSI & 2000 & the pelvis is over the seat and below 0.66\,m & 32 \\
        Boxing     & HSI & 900  & a palm comes within 0.18\,m of the target on the bag & 32 \\
        VisNavi    & HSI & 1800 & both ankles are inside the destination, robot standing & 32 \\
        \bottomrule
    \end{tabular}
\end{table}

\subsection{DexJoCo}
\label{app:dexjoco}

\paragraph{Tasks and interface.} We use the eleven tasks of DexJoCo~\citep{wang2026dexjoco} in its
rand-obj setting, in which the placement of the objects and the height of the table are
randomized at every reset (\Cref{tab:dexjoco-tasks}). Single-arm tasks use one Franka Panda arm
with a 16-degree-of-freedom Allegro hand, and bimanual tasks use two. The agent commands the
official policy action of the benchmark: for each arm, an absolute target for the flange
position and orientation in the world frame, the orientation as a rotation vector, together with
sixteen hand joint targets, which gives a 22-dimensional row for one arm and a 44-dimensional
row for two. The official operational-space controller pulls the flange toward its target and
position actuators drive the hand joints, without a motion planner. Each row advances the
simulation by 20\,ms, and one call to the interface executes between 1 and 150 rows.
Observations are 640$\times$640 RGB-D images from the official cameras of each task, with depth
in millimeters and the intrinsics and world pose of every camera, together with the official
proprioceptive state, the flange pose of each arm and the joint positions of each hand. No
force, torque, contact, or tactile signal is exposed, and the object poses and table height in
the simulator's state never reach the agent. The interface also offers a dry run of a motion on
the robot alone, without the table and objects, that does not advance the episode, and helpers
for rotation arithmetic and for lifting a pixel with its depth to a world position.

\begin{table}[htb]
    \centering
    \small
    \caption{\textbf{DexJoCo tasks, rand-obj setting.} All cameras are RGB-D. $^\ast$Including 30
    steps before control is handed over.}
    \label{tab:dexjoco-tasks}
    \begin{tabular}{@{}llllc@{}}
        \toprule
        \textbf{Task} & \textbf{Benchmark id} & \textbf{Arms} & \textbf{Cameras} & \textbf{Step limit} \\
        \midrule
        Hammer    & \texttt{hammer\_nail}             & one & front, wrist & 1000 \\
        Mouse     & \texttt{click\_mouse}             & one & ego\_right, wrist & 1200$^\ast$ \\
        Bucket    & \texttt{pick\_bucket}             & one & front, wrist & 1200 \\
        Tongs     & \texttt{pinch\_tongs}             & one & front, wrist & 1000 \\
        Glasses   & \texttt{fold\_glasses}            & one & front, wrist & 1200 \\
        Water     & \texttt{water\_plant}             & one & front, wrist & 1000 \\
        iPad      & \texttt{bimanual\_unlock\_ipad}    & two & ego, wrist\_left, wrist\_right & 1200 \\
        Hanoi     & \texttt{bimanual\_hanoi}           & two & ego, wrist\_left, wrist\_right & 1500 \\
        Assembly  & \texttt{bimanual\_assembly}        & two & ego, wrist\_left, wrist\_right & 1500 \\
        Microwave & \texttt{bimanual\_microwave\_cook} & two & ego, wrist\_left, wrist\_right & 1100 \\
        Photo     & \texttt{bimanual\_photograph}      & two & ego, wrist\_left, wrist\_right & 1000 \\
        \bottomrule
    \end{tabular}
\end{table}

\paragraph{Task information.} The agent receives a task description generated from the
task definition: the official instruction, the robot and its cameras, the step limit and
wall-clock limit of an episode, the fact that object placement and table height change between
episodes, and the binary success criterion without partial credit. It contains no detail of the
success check. The agent also receives three images of the completed task, the final frames of
three successful human demonstrations from the official dataset, each in a different layout,
taken from the task's overview camera. They contain no actions and no states. The three
demonstrations are the first that replay successfully in the official environment from initial
layouts that differ from all 150 evaluation layouts.

\paragraph{Exploration.} A session gives one agent, Claude Code with Claude Opus 5.5 at the xhigh
effort setting and without subagents, one simulated environment, and a budget of $N \in
\{2, 4, 8, 16, 32\}$ exploration episodes. The agent starts from an empty workspace (\Cref{app:sessions}). A exploration episode is either interactive, in
which the agent drives the robot itself through the interface or through its own scripts, or a
rehearsal, in which its registered program runs alone in a fresh container on a fresh episode
while the agent is paused; both consume the budget. The agent ends exploration by submitting a
program that has completed a rehearsal without error, so the budget is an upper bound, and over
the 55 sessions the agents used 592 of the 682 budgeted episodes (\Cref{tab:dexjoco-sessions}).
Exploration layout $i$ is the first reset of the official rand-obj environment with seed
$1000+i$, the same randomization as in evaluation, whose layouts come from seeds 0 to 2. An episode is
limited to the official number of steps and to 600\,s plus 5\,s per allowed step of wall-clock
time, a limit that no episode reached. The agent and its programs may call a vision-language
model, Gemini 3.8 Flash, with a budget of \$10 per session.

\paragraph{Evaluation.} Once the agent has exited, its workspace is frozen, and the submitted
program is evaluated alone on 150 episodes that reproduce the official evaluation of DexJoCo
layout for layout: seeds 0, 1, and 2, with 50 consecutive resets each. Every episode runs in a
fresh simulator process and a fresh container for the program, four episodes in parallel, under
the official step limit and success check, and an error or an early exit of the program counts
as a failure. The simulator advances only when the program sends actions, so evaluation is
synchronous and free of inference latency, whereas the official client of the benchmark is
asynchronous.

\begin{table}[htb]
    \centering
    \small
    \caption{\textbf{DexJoCo sessions by exploration budget.} One session per task. Episodes are
    summed over the eleven sessions; agent time runs from the start of the agent to evaluation and
    is the median over the eleven sessions, with the range in parentheses.}
    \label{tab:dexjoco-sessions}
    \begin{tabular}{@{}cccccc@{}}
        \toprule
        \textbf{Budget} & \textbf{Interactive} & \textbf{Rehearsal} & \textbf{Used / budgeted}
            & \textbf{Submitted early} & \textbf{Agent time (h)} \\
        \midrule
        2  & 11  & 11  & 22 / 22   & 0 of 11 & 0.70 (0.44--1.28) \\
        4  & 26  & 18  & 44 / 44   & 0 of 11 & 1.05 (0.51--2.46) \\
        8  & 57  & 29  & 86 / 88   & 2 of 11 & 1.34 (0.59--3.02) \\
        16 & 93  & 67  & 160 / 176 & 5 of 11 & 1.52 (0.72--2.89) \\
        32 & 179 & 101 & 280 / 352 & 8 of 11 & 1.76 (0.98--3.89) \\
        \bottomrule
    \end{tabular}
\end{table}

\subsection{LIBERO and LIBERO-Pro}
\label{app:libero}

\paragraph{Tasks and interface.} We use the 40 tasks of the Spatial, Object, Goal, and Long suites of
LIBERO~\citep{liu2023libero} in the official environment, numbered from 0 in the official order,
and the position-perturbed variants of LIBERO-Pro~\citep{zhou2025libero}, in which objects
exchange their initial regions. The agent commands the official action at 20\,Hz, increments of
the end-effector position and rotation, scaled so that 1 corresponds to 5\,cm or 0.5\,rad, and a
gripper command, with between 1 and 100 rows per call and no motion planner. It observes the scene
through the agent-view and wrist cameras at 256$\times$256 in RGB-D, with the intrinsics and the
pose of each camera, together with the end-effector pose and the finger positions, and it may lift
pixels to 3D points. The step limit of an episode is 220 on Spatial, 280 on Object, 300 on Goal, and
520 on Long, after ten initial steps without motion.

\paragraph{Task information.} The task description contains the official instruction, the robot
and its cameras, the limits of an episode, and the binary success criterion without the detail of
its check. The agent also receives three images of the
completed task, the final agent-view frames of three official human demonstrations that replay
successfully.

\paragraph{Exploration and evaluation.} We use Claude Opus 5.5 as our agent in two environments. The budget is
up to 20 exploration episodes on the official placement distribution, followed by at most 5 optional
episodes of adaptation on the perturbed distribution, which the agent may use when it judges that
its program will not transfer (\Cref{app:adaptation}). Exploration layout $i$ is a fresh
reset of the official environment with seed $1000+i$, a sample of the official placement
distribution and never a stored initial state. An episode is limited to 600\,s plus 5\,s per allowed
step. The program is evaluated on the 50 official initial states of each task and on the 50
position-perturbed states of LIBERO-Pro, each episode in a fresh container, synchronously, and with
binary success. Adaptation episodes are fresh resets of the perturbed task with seeds from 2000.

\clearpage
\section{Simulation Case Studies}
\label{app:sim-cases}

This section follows single sessions on the simulation benchmarks in detail: what the agent
explored, the program it wrote, and how that program behaves on unseen scenes.

\subsection{Re-Perception Under Perturbation: Two Cans into a Basket}
\label{app:libero-basket}

\begin{figure}[htb]
    \centering
    \includegraphics[width=\linewidth]{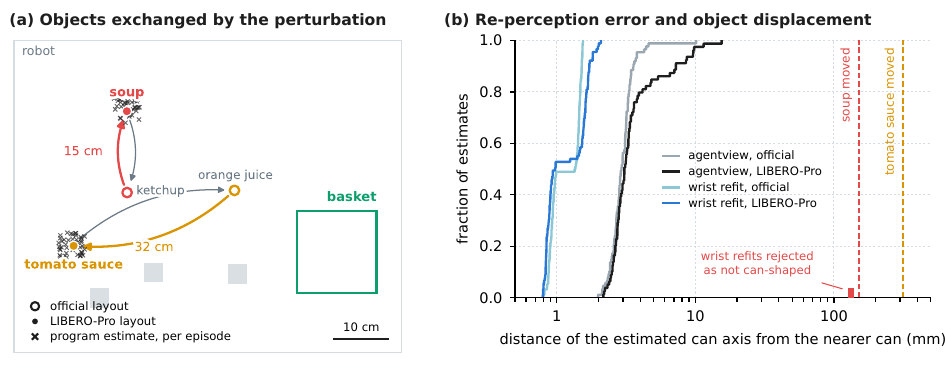}
    \caption{\textbf{Why the program transfers to LIBERO-Pro.} (a) Mean object positions over the 50
    official and the 50 perturbed layouts, seen from above with the robot at the top; the
    perturbation exchanges the soup with the ketchup and the tomato sauce with the orange juice.
    Crosses mark the program's first estimate of each can in every perturbed episode. (b) The
    distance between the program's estimate of a can's axis and the nearer can in the simulator,
    for the first agent-view estimate and for the wrist refit before each grasp, against the
    displacement of the two cans. The task takes the cans in either order, and in two perturbed
    episodes Gemini exchanged their labels; the rejected refits are those on the ketchup bottle.}
    \label{fig:libero-transfer}
\end{figure}

The first task of the LIBERO-Long suite asks the robot to put both the alphabet soup and the
tomato sauce in the basket, on a table that also holds a ketchup bottle, a carton of orange
juice, milk, butter, and cream cheese. The agent explored the task for 15 episodes in about an
hour, ten of them interactive and five rehearsals of its program, and wrote a program of 1,362
lines that succeeds in all 50 official initial states and in 44 of the 50 perturbed states of
LIBERO-Pro. The perturbation exchanges the soup with the ketchup and the tomato sauce with the
orange juice, which moves the two cans by 15 and 32\,cm (\Cref{fig:libero-transfer}a). This
section shows what the program perceives, how it plans, and why its behavior transfers.

\begin{figure}[htb]
    \centering
    \includegraphics[width=\linewidth]{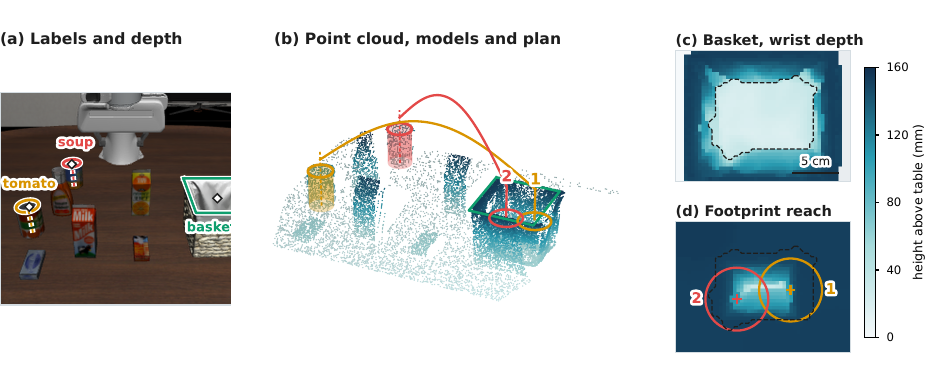}
    \caption{\textbf{Perception and planning on a LIBERO-Pro layout,} re-run with the agent's code
    on the stored frames of one evaluation episode. (a) The agent-view image with Gemini's points
    (diamonds) and the geometry that the program derives from depth: the top and axis of each can
    and the basket rim. (b) The point cloud of the agent-view camera, completed over the basket by
    the wrist camera and coloured by height above the table on the scale of (c) and (d), with the
    fitted cans, the basket rim, and the two planned placements, numbered in the order of
    execution. (c) The basket measured by the wrist camera in 4\,mm cells,
    with the interior outlined. (d) Footprint reach, the highest point under the footprint of a can,
    and the two spots that minimize it; the circles show the footprint.}
    \label{fig:libero-perception}
\end{figure}

\paragraph{Perception.} The program perceives the scene anew in every round, and it divides the
work between a vision-language model and depth (\Cref{fig:libero-perception}). Gemini only names
the objects: it returns a pixel on each can and in the basket, together with a short description
of each can's state. Every geometric quantity comes from depth. The program lifts the agent-view
depth into the world frame, takes the table as the mode of the point heights, and, starting from
each Gemini pixel, segments the object by a flood fill over 3D-connected points; the top height
is a high percentile of the segment, and the axis is the centroid of its top 1.2\,cm. A segment
counts as a can only if its top is 4.5 to 8\,cm wide and 5 to 30\,cm above the table, bounds
that the agent set after measuring every object on the table and that reject the taller and
narrower ketchup bottle. A second detector, which uses depth alone, finds flat, round plateaus
of the size of a lid and adds any can that Gemini missed. Before grasping, the program hovers
7\,cm above the lid and refits the lid disk from the wrist camera; a top that is not can-shaped
is blacklisted, and the grasp yaw is the first of four candidates whose finger zones are clear
of other objects.

\paragraph{Placing by footprint reach.} The agent-view camera cannot see the floor of the basket,
which its near wall hides, and the agent found during exploration that this flat floor is only about
7 by 10\,cm: drop spots chosen by clearance put the second can's footprint over the sloped walls,
where the can stalled. The program therefore maps the interior from the wrist camera while it
holds the first can, as a height map with 4\,mm cells, and computes for every cell the highest
point under the footprint of a can, its \emph{footprint reach} (\Cref{fig:libero-perception}c,d).
It chooses the pair of spots, at least 6.8\,cm apart, that minimizes the larger of their two
reaches, releases each can just above its reach, and remaps the basket after the first release.
During exploration, the agent also arranged perturbed layouts of its own,
stacking the tomato sauce on the soup, standing the soup on the cream-cheese box, placing the cans
side by side, and pushing the basket by 8\,cm and rotating it by 20$^\circ$, and it ran its program
on each of them.

\paragraph{Why the program transfers.} Nothing in the program refers to where the objects stood
during exploration, so the perturbation changes its inputs but not its procedure. Under
perturbation, its first estimate of each can's axis lies within a median of 3.1\,mm of the ground
truth, against 2.9\,mm on the official layouts, and the wrist refit within a median of 1.0\,mm,
while the cans themselves moved by 15 and 32\,cm (\Cref{fig:libero-transfer}b); a policy that
reproduced its official motions would miss each can by that displacement. The checks also absorb
labelling errors. When Gemini placed the soup's label on the ketchup bottle, which now stands
where the soup stood in the official layouts, the wrist refit measured a top 3.7\,cm across, and
the program rejected it as not can-shaped, blacklisted its position, and found the soup in the
next round. All six failures share one cause: the soup is not visible from the agent-view camera
at the start, and the program, which re-perceives only from its home pose, stops with about 300
steps left. The listings below condense the program's main loop and give the agent's code for
the footprint reach and the choice of drop spots, without docstrings and one log line.

\begin{raxpseudo}{run(), condensed from controller.py}
def run(robot):
    table_z = mode of the agent-view point heights
    repeat up to 3 rounds:                          # perceive anew in every round
        lids   = flat, round plateaus in depth      # depth alone
        labels = Gemini(agent view)                 # one pixel per can and the basket
        cans   = [flood_fill(p) for p in labels if is_can(top width, height)]
        cans  += lids that no label explains, minus blacklisted positions
        for can in order(cans):                     # stacked cans: the higher first
            hover 7 cm above the lid; refit the lid disk from wrist depth
            if not can_shaped(disk): blacklist(can); continue
            yaw = first of {0, 90, 45, -45} deg with clear finger zones
            descend to top - 3 cm; close; lift
            if not 3.5 cm < finger opening < 7.4 cm: continue
            if no basket map: map the basket from the wrist, holding the can
            spot = plan_spots(map) for the first can, else lowest_spot(remap)
            lower to the release height above the footprint reach; open; lift
            if task complete: return
\end{raxpseudo}

\begin{raxcode}[firstnumber=148]{basket.py\raxcodesep footprint reach and drop-spot planning}
def footprint_max(b, r=0.036):
    H = np.where(np.isfinite(b["H"]), b["H"], b["rim_z"])   # unobserved cells count as rim
    res = b["grid"][2]
    k = int(np.ceil(r / res))
    yy, xx = np.mgrid[-k:k + 1, -k:k + 1]
    fp = (xx ** 2 + yy ** 2) * res ** 2 <= r * r           # disk of the can's footprint
    return ndimage.maximum_filter(H, footprint=fp, mode="constant", cval=b["rim_z"])

def plan_spots(b, r=0.036, min_sep=0.068):
    F = footprint_max(b, r)
    ii, jj = np.nonzero(b["interior"])
    if len(ii) < 2:
        return None
    f = F[ii, jj]
    keep = np.argsort(f)[:400]
    ii, jj, f = ii[keep], jj[keep], f[keep]
    res = b["grid"][2]
    D = np.hypot(ii[:, None] - ii[None], jj[:, None] - jj[None]) * res
    cost = np.maximum(f[:, None], f[None]) + 0.5 * np.maximum(0.0, 0.08 - D)
    cost[D < min_sep] = np.inf
    a, c = np.unravel_index(np.argmin(cost), cost.shape)
    if not np.isfinite(cost[a, c]):
        return None
    s1, s2 = cell_xy(b["grid"], ii[a], jj[a]), cell_xy(b["grid"], ii[c], jj[c])
    return [(s1, float(f[a])), (s2, float(f[c]))]
\end{raxcode}

\clearpage
\subsection{Self-Calibration and Grasp Design: Tower of Hanoi with Two Dexterous Hands}
\label{app:hanoi}

The bimanual Tower of Hanoi task of DexJoCo asks for the final two moves of a three-level tower:
the medium red disk goes from the middle peg onto the large blue disk on the right peg, and the
small yellow disk from the left peg on top of it. Two Franka arms carry 16-degree-of-freedom
Allegro hands, and the disks slide over pegs that stand 13\,cm above them, so a grasp has to hold
a disk level as it travels along its peg without touching the peg. The agent received no grasp
and no model of the hand, only the joint semantics and a few generic postures in its skills. It
wrote a program that completed all 150 formal episodes
on unseen layouts, in which the board shifts by up to 4.9\,cm and the table height varies by as
much.

\begin{figure}[htb]
    \centering
    \includegraphics[width=\linewidth]{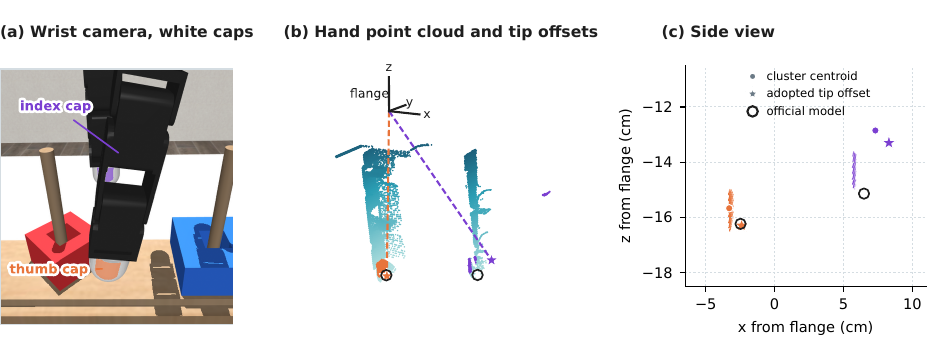}
    \caption{\textbf{Measuring the fingertips with the wrist camera,} in the agent's first episode,
    re-run on its stored frames. (a) The right wrist camera with the hand in its grasp posture;
    the agent segments the white fingertip caps and splits them into two clusters. (b) The hand's
    point cloud in the frame of the measured flange, with the two clusters and the tip offsets
    the agent adopted (stars). (c) A side view of the clusters, their centroids and the adopted
    offsets, with the fingertips of the official hand model, which the agent never saw, for
    comparison.}
    \label{fig:hanoi-calibration}
\end{figure}

\paragraph{Measuring its own fingertips.} The agent spent its first episode on its own hand,
without touching a disk. It moved single joints of the right hand and lifted the wrist cameras'
depth into world points to see what each joint moves, finding that the first thumb joint turns
the thumb from pointing sideways to pointing straight down and that the last two flex its tip
forward. It then held the posture it meant to grasp with, the index finger and the thumb both
pointing down, segmented the white fingertip caps in the right wrist camera, lifted them to 3D,
and split them into two clusters (\Cref{fig:hanoi-calibration}). Relative to the measured flange,
the thumb cluster lies at $(-3.3, 4.1, -15.7)$\,cm and the index cluster at $(7.3, 3.1,
-12.9)$\,cm; the agent adopted tip offsets of $(-2.4, 3.8, -16.3)$ and $(8.3, 3.6, -13.3)$\,cm,
about 1\,cm toward the centre of each tip, and every grasp of its program is built on them, as
the first listing at the end of this section shows. The agent never saw a model of the hand, yet its thumb offset lies within
1.1\,mm of the fingertip of the official model. Its index offset lies 27\,mm from the model's,
because the camera sees only the upper part of that cap, but the error shifts the planned pinch
line by only about 8\,mm across a 7.7\,cm disk. The same measurements placed the wrist camera at
$(-14.0, -3.0, 4.1)$\,cm from the flange, exactly where the official model puts it.

\begin{figure}[htb]
    \centering
    \includegraphics[width=\linewidth]{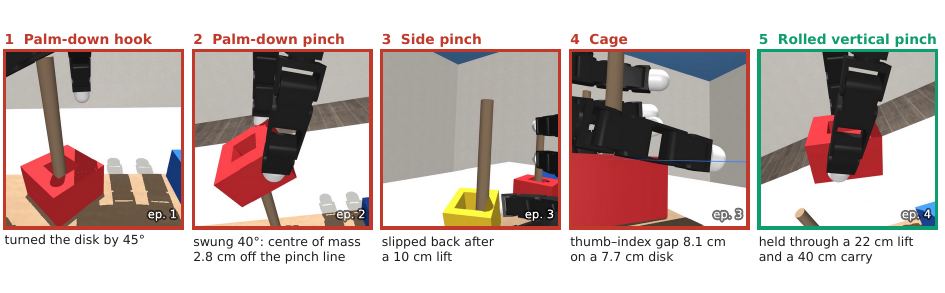}
    \caption{\textbf{The grasps the agent tried,} seen by the wrist camera of the grasping hand in
    its exploration episodes, with panels 3 and 4 rotated so that the peg stands upright. The first
    four failed for the reasons given below them; the rolled vertical pinch carried both disks and
    became the grasp of the program.}
    \label{fig:hanoi-grasps}
\end{figure}

\paragraph{Designing the grasp.} The agent then designed its grasp by trying candidates on the red
and yellow disks (\Cref{fig:hanoi-grasps}). A palm-down hook pressed the palm onto the disk, and
flexing the thumb turned the disk by 45$^\circ$ about its peg. A palm-down pinch at the wall of
the disk lifted it, but the disk swung by about 40$^\circ$ about the pinch line, which the agent
measured to pass 2.8\,cm from the disk's centre of mass, and a firmer squeeze dropped it. A side
pinch lifted the yellow disk by 10\,cm before it slipped back onto its peg, and a cage grasp left
only 4\,mm between the hand's fixed thumb--index gap of 8.1\,cm and the 7.7\,cm red disk, so that
only the thumb touched it. From these failures, the agent derived that rolling the hand by an
angle $\theta$ moves the index edge of the palm $0.10\sin\theta - 0.013\cos\theta$\,m away from the
thumb--index line, so that a roll of about 25$^\circ$ clears the peg and lets the pinch line pass
through the centre of the disk. The resulting rolled vertical pinch, with a roll of 0.44\,rad,
held both disks through a 22\,cm lift and a 40\,cm carry in the fourth episode.

\paragraph{The program.} On every layout, the program measures the disks and pegs from the
egocentric depth image, pinches both disks at once, and moves the red disk first while the left
hand waits. It verifies every lift by measuring the held disk from three
cameras, and it opens, backs off, and pinches again when the disk rose less than half of the lift.
Because each arm keeps converging to its command after a move, the agent measured the tracking
lag, which decays by a factor of 0.78 every five steps, and the program aligns the disk with the
peg on its predicted final position rather than on its current one. Before submitting, the agent
injected alignment errors of 8 and 15\,mm and a missed pinch into its own runs, and the program
recovered from all of them; its five versions completed all of their autonomous runs.

\begin{figure}[htb]
    \centering
    \includegraphics[width=\linewidth]{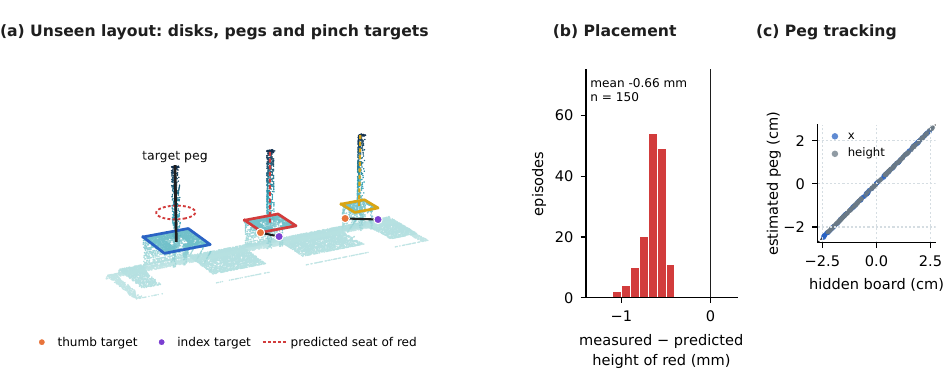}
    \caption{\textbf{Perception on an unseen layout and precision over the 150 formal episodes,}
    re-run with the agent's code. (a) The egocentric point cloud, coloured by height, with the
    measured disks, the peg axes, the target peg, the thumb and index targets of both pinches, and
    the predicted seat of the red disk. (b) The height of the red disk measured after placement,
    minus the height the program predicted. (c) The program's estimate of the target peg against
    the hidden pose of the board, both relative to their means over the 150 layouts.}
    \label{fig:hanoi-scene}
\end{figure}

\paragraph{Precision on unseen layouts.} The program's estimate of the target peg follows the hidden
pose of the board across all 150 formal layouts, with residual standard deviations of 0.24, 0.05,
and 0.18\,mm in $x$, $y$, and height (\Cref{fig:hanoi-scene}). After placing the red disk, the
program measures its top again and finds it on average 0.66\,mm, and at most 1.0\,mm, below the
height it predicted from the blue disk, the counterpart of the 6.5\,mm check on the air fryer
(\Cref{app:perception}).

\begin{raxcode}[firstnumber=10]{task.py\raxcodesep tip offsets and the rolled vertical pinch}
TH_TIP = np.array([-0.024, 0.038, -0.163])   # default-frame right-hand thumb tip (th0=1.396, straight)
IX_TIP = np.array([0.083, 0.036, -0.133])    # index tip, x1=1.57 straight
R_T = 0.012
ROLL = 0.44
DISK_H = 0.05
PEG_ABOVE_TOP = 0.135                        # peg top above top of a disk resting on the board
PRESHAPE = dict(ff=(0, 1.3, 0, 0), th=(1.396, 0, 0, 0))
SQUEEZE = dict(ff=(0, 1.75, 0.5, 0.3), th=(1.396, 0, 0.7, 0.4))
RELEASE = dict(ff=(0, 1.2, 0, 0), th=(1.396, 0, 0, 0))
PARK = {"right": np.array([-0.45, -0.40, 1.42]), "left": np.array([-0.45, 0.40, 1.42])}


def pinch_pose(side, c, half, yaw, mid, gap_t=0.006, dz=-0.010):
    """Flange pose so thumb tip is gap_t outside the near face and the thumb-index line crosses the centre."""
    c = np.asarray(c, float)
    th_off = world_offset(side, TH_TIP, ROLL, yaw)
    ix_off = world_offset(side, IX_TIP, ROLL, yaw)
    u = np.array([np.cos(yaw), np.sin(yaw)]); v = np.array([-np.sin(yaw), np.cos(yaw)])
    lat_t = (th_off[:2] @ v - ix_off[:2] @ v) / 2
    t_xy = c + (-half - R_T - gap_t) * u + lat_t * v
    t = np.array([t_xy[0], t_xy[1], mid + dz])
    flange = t - th_off
    rv = rotvec_near(hand_rot(side, ROLL, yaw), np.array([-np.pi, 0, 0]))
    return flange, rv
\end{raxcode}

\begin{raxpseudo}{run(), condensed from controller.py}
def run():
    disks  = colour and depth boxes from the egocentric camera   # measured on every layout
    pegs   = depth column above each disk; target = fuse(blue disk, blue peg)
    pinch(red, right hand) and pinch(yellow, left hand) at once     # rolled vertical pinch
    for hand, disk in [(right, red), (left, yellow)]:
        repeat up to 3 times:
            lift until the disk clears the peg top
            if the held disk rose less than half the lift:          # three cameras
                open, back off, measure again, pinch again          # GRASP FAILED
        carry: flange = target - (disk centre - measured flange)
        repeat up to 3 times:                       # settle, then predict the final pose
            lag = measured flange - command
            err = target - (disk centre - lag);  stop if |err| < 4 mm
        lower onto the support; check the bottom; open; lift by 15 cm
        if disk is red: measure its top, predicted as the blue top + 5 cm
\end{raxpseudo}

\clearpage
\subsection{From Exploration to Code: Carrying a Hammer over a Bar on a Humanoid}
\label{app:doubledesk}

In the DoubleDesk task of HumanoidArena, a Unitree G1 humanoid has to take a hammer from one desk,
carry it over a bar on the floor, and drop it into a basket on a second desk. The agent commands
the robot through whole-body kinematic references that a learned tracker, SONIC, executes; it
was given the tracker, three teleoperated demonstrations, a forward-kinematic model of the robot,
and a tool that returns the level frame of the pelvis. Over 125 exploration episodes and 10.7 hours,
it wrote a program of 3,639 lines in eleven files, of which the 637-line kinematic model is the
one it was given. The program completed 34 of the 60 formal episodes; before submitting it, the
agent predicted ``roughly half to two-thirds''.

\begin{figure}[htb]
    \centering
    \includegraphics[width=\linewidth]{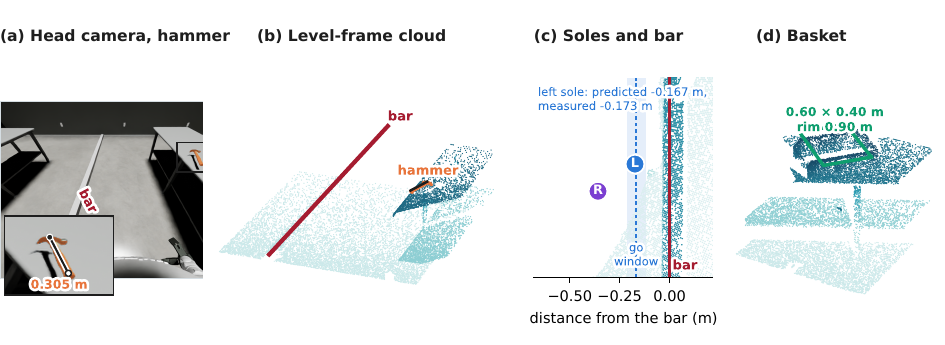}
    \caption{\textbf{Perception on an unseen DoubleDesk scene,} re-run with the agent's code on
    the stored frames of one formal episode. (a) The head camera at the start, with the hammer the
    program finds and, enlarged, its axis and length. (b) The point cloud in the level frame,
    coloured by height, with the hammer and the bar fitted to it. (c) The last stance before the
    crossing, seen from above in the frame of the bar: the soles, the window from which the program
    starts a crossing, and the position of the left sole that the stance model predicted. (d) The
    basket found at the placing stance.}
    \label{fig:dd-perception}
\end{figure}

\paragraph{Perception.} Every quantity the program uses comes from the depth image of the head
camera, lifted into a point cloud in the level frame of the pelvis (\Cref{fig:dd-perception}).
The rotation into that frame comes from the provided tool, and the agent's own fit of the floor
plane by random sample consensus agrees with it to within 0.07$^\circ$. On this cloud, the agent's
code finds the floor as the mode of the point heights, the bar as the dominant line among points 5
to 11\,cm above the floor, the desk tops and their near edges among upward-facing points, and the
back wall, and it localizes the robot from the bar, the desks, and the wall against a map whose
constants it measured itself. It finds the hammer by its orange handle, takes the handle's
principal axis, decides which end carries the head from the dark points beyond the handle, and
measures its length, using the 0.302 to 0.310\,m it found during exploration to place the centre when
part of the hammer is hidden; on the unseen scene of \Cref{fig:dd-perception}, the length is
0.305\,m. It finds the basket as the largest cluster of
points above a desk and fits a rectangle of 0.60 by 0.40\,m with its rim 0.90\,m above the floor,
and it measures the distance of each sole from the bar by combining the fitted bar with the
kinematic model. The program perceives anew at every stage and before every step at the bar, and
after a grasp it confirms that the hammer is held by measuring the height of the handle again.

\begin{figure}[htb]
    \centering
    \includegraphics[width=\linewidth]{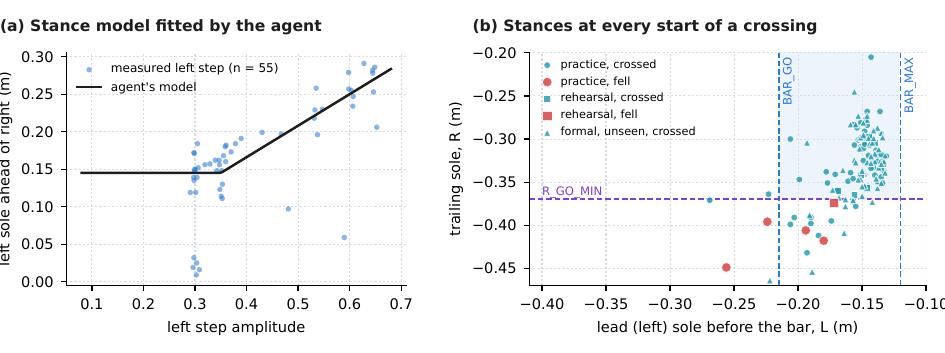}
    \caption{\textbf{The stance model and the crossing window.} (a) How far a left step from a
    squared stance places the left sole ahead of the right one, against the step amplitude: the
    agent's measurements and the piecewise-linear model it wrote. (b) The positions of the two
    soles before the bar at every start of a crossing in exploration, rehearsal, and formal episodes,
    with the window that the program uses.}
    \label{fig:dd-stance}
\end{figure}

\paragraph{Locomotion and the bar.} The agent's skills recommended walking by replaying a cycle cut
from a teleoperated recording. The agent cut a symmetric 66-step cycle from one demonstration,
looped it with a ramp in heading, and measured that its amplitude does not control the length of
a single step, which varied by $\pm$10\,cm; it therefore calibrated multi-step walks against its
own localization and used them as a table of distances. Sidestepping moved the robot by 1 to 2\,cm
per cycle, so lateral placement turns and then walks. The crossing replays the step over the bar
from one demonstration: with the recorded pitch, the pelvis did not travel and the lead foot
landed on the bar, so the agent scaled the pitch to zero; perpendicular approaches landed the lead
foot on the edge of the bar, so it approaches at 64$^\circ$, where the swinging foot also reaches
sideways; and it adds a trailing step when the robot ends astride the bar, after learning never
to turn in that position. Whether a crossing succeeds depends on the stance from which it starts,
and the agent built a model of how each step moves the two soles relative to the bar, from its
own measurements of 102 transitions (\Cref{fig:dd-stance}a). The model predicts that a left step
places the left sole 14.5\,cm ahead of the right one up to an amplitude of 0.35, and up to 25\,cm
at 0.6; a planner searches three actions ahead to bring the left sole into the window from which
the program crosses. The bounds of that window are measured too: crossings started with the left
sole within 21\,cm of the bar succeeded in 17 of 17 trials, which set the far edge of the window
at 21.5\,cm, and the trailing sole is kept within
37\,cm of it, since the agent found that crossings with the trailing sole closer than 35.5\,cm had
succeeded 27 times out of 27, and those with it between 35.5 and 42\,cm had fallen 4 times out of
13. The agent then wrote its own Monte Carlo simulators of the approach, which replay the stance
model with its measured errors, and used them to choose that margin, which lowered the simulated
risk of a fall from 9.3 to about 6\% at a cost of 0.3 actions. Over all 137 crossings in its logs,
every crossing started with the trailing sole within 35.5\,cm of the bar succeeded, and in the formal
evaluation the robot crossed the bar in all 59 logged episodes without a fall
(\Cref{fig:dd-stance}b). In the episode of \Cref{fig:dd-perception}c, the model predicted that the
left step would bring the left sole to 16.7\,cm before the bar, and the point cloud measured it at
17.3\,cm.

\begin{figure}[htb]
    \centering
    \includegraphics[width=\linewidth]{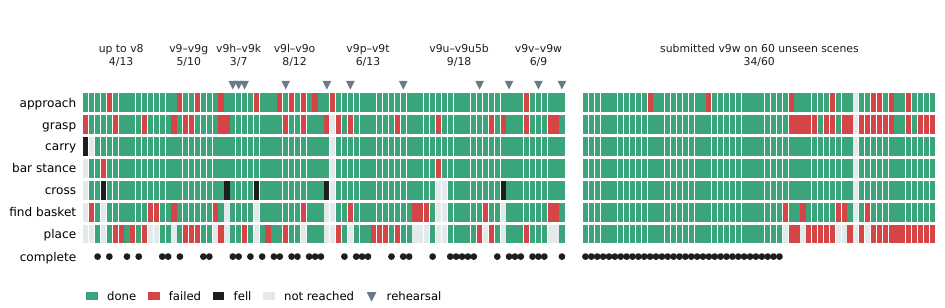}
    \caption{\textbf{The outcome of every stage in every program run,} in interactive and rehearsal
    episodes grouped by version (left), and for the submitted version on the 60 formal episodes,
    ordered by outcome (right). Each column is one episode; the counts give completed episodes.}
    \label{fig:dd-strip}
\end{figure}

\paragraph{The program and its versions.} The program runs seven stages, each with its own
checks: it approaches the hammer until the handle lies at about 60$^\circ$ to its heading, which
the agent found to give grasps that hold; it grasps with up to two attempts, treating a closure
without torque as a miss; it carries the hammer to the bar while correcting its odometry from the
bar and the basket; it plans its stance at the bar and crosses only from the window; it verifies
after the crossing that both soles are past the bar; and it finds the basket and places the hammer
inside its fitted rectangle. Each stage has a deadline in steps, measured from the durations of
earlier runs. The agent kept its own table of every stage in every episode, linked each failure to
the change it made, and grew the program over more than thirty versions
(\Cref{tab:dd-versions,fig:dd-strip}). Its interactive and rehearsal runs completed 41 of 82
episodes, rising to 6 of 9 for the final versions, and in the formal evaluation most failures
occurred at the grasp, which held in 40 of the 59 logged episodes.

\begin{table}[htb]
    \centering
    \small
    \caption{\textbf{Versions of the DoubleDesk program.} From the agent's own records.
    Completed counts interactive and rehearsal runs of each group of versions.}
    \label{tab:dd-versions}
    \begin{tabularx}{\linewidth}{@{}l>{\raggedright\arraybackslash}X>{\raggedright\arraybackslash}Xc@{}}
        \toprule
        \textbf{Versions} & \textbf{Main changes} & \textbf{Failures that motivated them} & \textbf{Completed} \\
        \midrule
        up to v8 & First full chain; localization from the bar and the basket; crossing window and
            stance planner; grasp retries; basket approach by turning, then walking & Odometry off by
            0.55\,m; a walk onto the bar; a fall from 25.6\,cm; a false miss that dropped the hammer
            & 4/13 \\
        v9--v9g & Left step aimed by the stance model; handle at 60$^\circ$; deadlines; a subagent's
            grasp measured at 3/13 against 13/22 and reverted & Late grasps; sweeps of the desk
            & 5/10 \\
        v9h--v9k & Looser carrying posture; limited fine steps and backing off; window back to
            21.5\,cm & The arm over the head; the hammer lost; a fall from 22.4\,cm & 3/7 \\
        v9l--v9o & Limits on the trailing sole; stance model refitted on 60 transitions with a
            calibration episode; correction of the handle axis & A fall with the trailing sole
            41.8\,cm back; a handle at 85$^\circ$; a flipped hammer & 8/12 \\
        v9p--v9t & Trailing-sole margin from the 27/27 against 4/13 record; search turn; stall
            guard; placing in a partly seen basket & A fall in rehearsal; stalled steps; a basket
            not seen & 6/13 \\
        v9u--v9u5b & Re-approach; correction of lateral drift (9.5\,cm, $n=11$); cycle guard; a
            faster place near the step limit & Three to six actions at the bar; a rehearsal cut off
            with 49 steps left & 9/18 \\
        v9v--v9w & Turning toward the basket and looking before walking; position tolerance of
            9.5\,cm & Walks back at the basket; an overshoot of 35\,cm & 6/9 \\
        \bottomrule
    \end{tabularx}
\end{table}

\begin{raxcode}[firstnumber=58]{phases.py\raxcodesep the crossing window and its evidence}
BAR_GO = -0.215         # crossings from L >= -0.21 succeeded 17/17 (ep26-57); -0.232 (x1), -0.224 (x2) and -0.223
                        # (ep43) crossed but -0.224 with R -0.396 fell (ep59), as did -0.256/-0.257
BAR_MAX = -0.12
R_GO_MIN = -0.37        # trailing (right) sole at the go: R > -0.355 crossed 27/27; R -0.42..-0.355 fell 4/13
                        # (ep72 -0.364, R5 -0.374, ep59 -0.396, ep64 -0.418)
R_PLAN_MIN = -0.355     # planned stances keep the right sole at least here (sim: fall risk 9
\end{raxcode}

\begin{raxcode}[firstnumber=252]{phases.py\raxcodesep crossing with verification}
def cross_bar(r, log, f=None):
    """Two high steps over the bar (demo01 replay, expert roll, no pitch). Returns True when both feet are past."""
    if f is None:
        f = C.measure_feet(r)
    lead = 'L'   # diagonal approach (heading = normal - 26 deg): the left foot leads
    r.step(C.cross_frames(r.base_joints(), r.base_yaw(), r.hands, mirror=(lead == 'R'), rp=1.0, pitch=0.0))
    r.hold(20)
    f2 = C.measure_feet(r)
    log(f"[bar] crossed lead {lead}: after {None if f2 is None else (round(f2['L'], 3), round(f2['R'], 3))}")
    if f2 is not None and min(f2['L'], f2['R']) < 0.0 < max(f2['L'], f2['R']):
        trail = 'L' if f2['L'] < f2['R'] else 'R'
        log(f"[bar] straddle; trail step {trail}")
        r.step(C.cross_frames(r.base_joints(), r.base_yaw(), r.hands, 905, 965, mirror=(trail == 'L'), rp=1.0,
                              pitch=0.0))
        r.hold(20)
        f2 = C.measure_feet(r)
    return f2 is None or min(f2['L'], f2['R']) > 0.0
\end{raxcode}

\begin{raxpseudo}{run() and deliver(), condensed from controller.py}
def run():
    approach the hammer until the handle lies at about 60 deg   # search turns if unseen
    grasp with up to 2 attempts                  # a closure without torque is a miss
    while not held and step < 1000:              # deadlines measured from earlier runs
        re-approach if the hammer moved out of reach; grasp again
    deliver()

def deliver():
    carry: locate from the bar and the basket; turn; walk a calibrated distance; face 64 deg
    repeat up to 3 times:
        repeat up to 8 times:                    # stance planning at the bar
            measure L, R from depth and kinematics
            action = bar_action(L, R)            # go, left step, right step, back, or walk
            if action is go: break
        if cross_bar(): break                    # both soles past the bar
    raise the hand; find the basket (up to 6 looks); place inside the fitted rectangle
\end{raxpseudo}

\clearpage
\subsection{Adaptation Under Perturbation on LIBERO-Pro}
\label{app:adaptation}

After its exploration on unperturbed layouts, \rax may adapt its program to the perturbed
distribution of LIBERO-Pro. A subagent reads the program, the description of the perturbed
task, and the first frame of a perturbed layout, and judges whether the program will transfer.
When it judges that the program will not, the agent resumes from its workspace, with the notes
and code of its exploration, and continues for at most 5 episodes on perturbed layouts before the
program is evaluated again on its own. In this mode, \rax runs as an agent-in-the-loop program
across the change of distribution: the program acts alone, and the agent steps in only to
revise it, so that adaptation exercises the continual learning of \Cref{sec:continual}. The
agent adapted its program on five of the 40 tasks, tasks 3, 8, and 9 of the Long suite and
tasks 1 and 5 of the Goal suite (\Cref{tab:adaptation}).

\begin{table}[htb]
    \centering
    \small
    \caption{\textbf{Tasks adapted to LIBERO-Pro.} Success rate (\%) on the 50 position-perturbed
    states before and after adaptation, with the exploration episodes it used.}
    \label{tab:adaptation}
    \begin{tabular}{@{}llcc@{}}
        \toprule
        \textbf{Task} & \textbf{Instruction} & \textbf{Episodes} & \textbf{Before $\rightarrow$ after} \\
        \midrule
        Long 3 & put the black bowl in the bottom drawer of the cabinet and close it & 5 & 0 $\rightarrow$ 100 \\
        Long 8 & put both moka pots on the stove & 5 & 0 $\rightarrow$ 100 \\
        Long 9 & put the yellow and white mug in the microwave and close it & 4 & 0 $\rightarrow$ 2 \\
        Goal 1 & put the bowl on the stove & 5 & 2 $\rightarrow$ 96 \\
        Goal 5 & push the plate to the front of the stove & 4 & 2 $\rightarrow$ 44 \\
        \bottomrule
    \end{tabular}
\end{table}

\begin{figure}[htb]
    \centering
    \includegraphics[width=\linewidth]{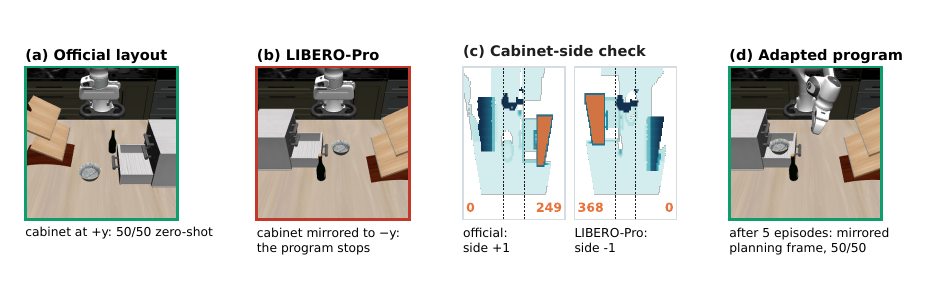}
    \caption{\textbf{Adapting to a mirrored cabinet.} (a) An official layout of the third Long task
    and (b) the same layout under the LIBERO-Pro perturbation, which moves the cabinet to the other
    side of the table. (c) The adapted program's check of the side, re-run on both first frames:
    level cells of the cabinet top above 1.08\,m (orange) on either side of the robot, counted on
    the height map of the agent-view point cloud. (d) The adapted program on the perturbed layout,
    with the bowl in the drawer.}
    \label{fig:adaptation}
\end{figure}

\paragraph{A mirrored cabinet.} The third task of the Long suite asks the robot to put the black bowl
in the bottom drawer of the cabinet and close it. The program solved all 50 official states, but
LIBERO-Pro places the cabinet on the other side of the table, facing the other way
(\Cref{fig:adaptation}a,b), and the program, whose code assumed a drawer that opens toward
$-y$, failed in all 50 perturbed states. Its checks kept it from acting blindly: in 38 of them, it
found no drawer panel where it expected one and stopped at the first step. Resuming from its
workspace, the agent identified the cause in its first episode on the perturbed layouts and,
rather than rewriting the program, added a 36-line layer that mirrors the frame in which it
plans: points, orientations, and action rows are reflected across the plane $y = 0$ whenever the
cabinet stands at $-y$ (the listing below). The program decides the side itself, from the flat top
of the cabinet in the agent-view point cloud (\Cref{fig:adaptation}c). In a stress test, the agent
moved the wine bottle in front of the drawer and found that its detector of the drawer panel took
the bottle for the panel; it narrowed the detector's window, and with its five episodes spent, the
adapted program completed all 50 perturbed states (\Cref{fig:adaptation}d).

\begin{raxcode}{frame.py\raxcodesep the mirror layer added during adaptation (excerpt)}
"""World <-> planning frame. The controller plans in a frame where the cabinet stands at +y and its drawer opens
toward -y (the layouts it was developed on). A cabinet at -y facing +y (swapped layouts) is handled by mirroring
y: points (x, -y, z), rotation vectors (-wx, wy, -wz) (pseudovector under the reflection), action rows likewise.
The mirror is an involution, so the same functions map both ways."""
MIRROR = False

def pts(P):
    if not MIRROR:
        return P
    P = np.array(P, dtype=float, copy=True)
    P[..., 1] *= -1
    return P

# state(s) mirrors the 8-D state in the same way

def row(r):
    """7-D action row [dx, dy, dz, drx, dry, drz, grip]."""
    r = list(r)
    if MIRROR:
        r[1], r[3], r[5] = -r[1], -r[3], -r[5]
    return r
\end{raxcode}

\clearpage
\section{Exploration Budget of ASPIRE on LIBERO-Pro}
\label{app:aspire}

ASPIRE~\citep{lu2026aspire} writes robot programs in CaP-X~\citep{fu2026cap} with Claude Code
and Claude Opus 4.6. Its programs call the CaP-X interfaces for perception, including
text-prompted segmentation and vision-language pointing, for grasp planning, including a
top-down grasp selector, for inverse kinematics, and for joint-space motion, and they may not
read the simulator's ground-truth state. On LIBERO-Pro, the unit of development is a task under
one perturbation: each of the 30 tasks of the Spatial, Object, and Goal suites is developed
separately for the position and the task perturbation, giving 60 programs, each written for
its own perturbed suite.

For each task and perturbation, ASPIRE executes an initial program on a debugging split of 15
environment seeds (51--65), where a seed fixes one instance of the task, including object
poses, distractors, and the initial state of the robot. After each execution, the coding agent
reads a trace of every primitive call, with its inputs, outputs, and return status, the camera
frames before and after the call, grasp candidates, object poses, and motion-planning results.
It diagnoses the failure, patches the program, and replays the debugging seeds, with at most
three replays per seed. On some tasks, an evolutionary search additionally proposes $K$
candidate programs per iteration, conditioned on the three best programs and their failure
traces, and executes each candidate on the debugging split; the reported searches run for up
to four iterations, and $K$ is not reported. The final program is executed once on a
validation split of 15 further seeds (66--80) and then evaluated on seeds 1--50. A coordinator
promotes validated repairs into a shared library of skills, each stored with its failure
signature, its condition of use, its repair strategy, and a code sketch, which later sessions
read as in-context guidance. With the same loop on Claude Opus 5, \citet{kang2026encore} report
a median of 0.99 hours and 91 thousand output tokens per task and perturbation.

The debugging and validation seeds index instances of the same perturbed suite as the
evaluation seeds, so every rollout before the final evaluation is drawn from the distribution
on which the program is evaluated. This is also required by the task perturbation, which
changes the goal of the task: a program developed on the original task would pursue the
original goal. ASPIRE's result on the Long suite comes from a separate setting, in which a
skill library built on the 90 tasks of LIBERO-90 is applied to the Long tasks without any
debugging on them. ASPIRE does not report the total number of rollouts per task;
\Cref{tab:aspire-budget} gives the budget that its protocol implies, alongside that of \rax.
With at most 5 rollouts from the evaluation distribution on the five tasks that adapted to the
perturbation (\Cref{app:adaptation}), and none on the other 35, \rax exceeds ASPIRE on every suite
under position perturbation, by 28.2 points on Spatial, 2.0 on Object, 6.0 on Goal, and 48.8 on
Long (\Cref{tab:libero}).

\begin{table}[htb]
    \centering
    \small
    \caption{\textbf{Exploration before evaluation under position perturbation.} ASPIRE's counts
    are its debugging and validation splits for one task and perturbation; evolutionary search adds
    $K$ rollouts per debugging instance and iteration.}
    \label{tab:aspire-budget}
    \begin{tabular}{@{}lcc@{}}
        \toprule
        & ASPIRE & \rax \\
        \midrule
        Programs per task & one per perturbation & one, adapted on 5 of 40 tasks \\
        Instances from the evaluation distribution & 30 & 0, or up to 5 when adapted \\
        Rollouts on these instances & up to 75, plus evolutionary search & 0, or up to 5 when adapted \\
        Exploration on unperturbed layouts & -- & up to 20 episodes \\
        Evaluation under perturbation & after development on it & zero-shot, or after adaptation \\
        \bottomrule
    \end{tabular}
\end{table}

\section{Summary Results of \texorpdfstring{\Cref{fig:teaser}}{Figure~\ref{fig:teaser}}}
\label{app:teaser}

The two simulation charts of \Cref{fig:teaser} summarize RoboDojo, BEHAVIOR-1K, HumanoidArena,
DexJoCo, and LIBERO-Pro, and compare \rax with the strongest prior model, that is, with
learned policies such as VLAs and world action models rather than with other agents. For each
benchmark, the right chart pairs the average success rate of \rax, with each task at its largest
budget, with the average of the strongest prior model on the same tasks: VPP2 on RoboDojo, the
highest-ranked challenge entry on BEHAVIOR-1K, WB-WAM on HumanoidArena, ME-Dex on DexJoCo, and
VPP2 on LIBERO-Pro. On LIBERO-Pro, the chart therefore compares with VPP2 (42.3\%), the strongest
model under position perturbation, and not with the coding agent ASPIRE (63.2\%), which we compare with in \Cref{sec:exp-libero}. On DexJoCo,
the chart includes only the five bimanual tasks, iPad, Hanoi, Assembly, Microwave, and Photo.
The six single-arm tasks are too easy to show the effect of scaling, since \rax solves each of
them within a few exploration episodes, and the bimanual tasks are the harder part of the
benchmark: with 32 exploration episodes, \rax averages 94.3\% on them, against 48.8\% for ME-Dex.
The average margin of 37.4 points stated in the abstract and the introduction is the mean of the
five margins of this chart, 53.4 on RoboDojo, 38.3 on BEHAVIOR-1K, 7.9 on HumanoidArena, 45.5 on
the bimanual tasks of DexJoCo, and 42.1 on LIBERO-Pro.

The per-benchmark margins listed in the introduction instead compare with the strongest prior
result of any kind on all tasks of each benchmark. On DexJoCo, this is the average of all eleven
tasks at 32 exploration episodes, 96.7\%, against ME-Dex at 65.3\% (+31.4), and on LIBERO-Pro it
is ASPIRE (+21.2). \Cref{sec:exp-dexjoco} further reports the best session of each task across
budgets, which averages 99.5\%. The left chart averages, at each budget, the success rates of the
tasks within each benchmark and then the five benchmarks with equal weight. A task without a
result at a budget shown counts with its result at the largest smaller budget. On LIBERO-Pro, 37
of the 40 tasks have a budget of 15 exploration episodes and use 11 to 15, Goal 1, 5, and 9 have
a budget of 20 and use 18, 20, and 19, and the 40 tasks use 14.2 on average (13.5 on Spatial,
14.0 on Object, 15.1 on Goal, and 14.0 on Long), so their results are included at 16. The dashed line, 41.2\%,
is the mean of the strongest prior models of the right chart.

\clearpage
\section{Real-Robot Tasks}
\label{app:real}

This section describes the setup of every real-robot experiment: the task, the robot and its
sensors, what the agent was given, what the operator did, how many trials the agent ran, how
success was judged, and the final outcome. Five tasks use a Franka FR3 arm with a parallel
gripper, two exterior RGB-D cameras (left and right of the table), and an RGB-D camera on the
wrist; the sixth uses a Rainbow RB-Y1 mobile manipulator with an RGB-D camera on its head. The
control interfaces are those of \Cref{sec:control}. In every task the agent received only the task
text quoted below, typed by the operator, and the robot's tools; it received no demonstrations,
no object models, and no trained policy. Each session ran Claude Code with maximum reasoning
effort. The agent could send requests to the operator through its tools, and the operator could
reply in a chat; we report every reset and intervention recorded in the agent's notes. Times are
local wall-clock times taken from the agent's notes and the camera recordings, which are timestamped
frame by frame. The key-frame figures show raw camera frames without overlays.

\subsection{Find the Cube}
\label{app:real-cube}

\paragraph{Task.} The operator hides a white cube of about 3\,cm under one of four inverted green
cups (about 9.3\,cm tall) and shuffles the cups by hand; the robot must then lift the cup that
holds the cube away and grasp and lift the cube. The task text was: \emph{``First wait for the
operator to place a cube under a green cup and watch it permute the positions of the cups. after it
finishes, you should identify the cup with the cube insides and lift the cup away, then grasp the
cube and lift it in the gripper, as success.''} The agent had to watch the shuffle while it
happened: it held the arm still at a watch pose above the cups and recorded the wrist camera's
stream, then identified the cup from that recording.

\paragraph{Operator.} The operator hid the cube, shuffled the cups, and reset the scene between
trials when the agent asked. During the session the operator made the game harder in three steps.
At first the cube was covered by a cup and the cups were slid around. In later trials the operator
also lifted the cube and moved it, or set a decoy cup on its old spot. From 21:35 the operator
dropped the cube into an upright cup, not necessarily the nearest, turned that cup over, and then
shuffled. From 21:50 the cube was replaced by a red wedge of about 4\,cm.

\begin{figure}[htbp]
    \centering
    \includegraphics[width=\linewidth]{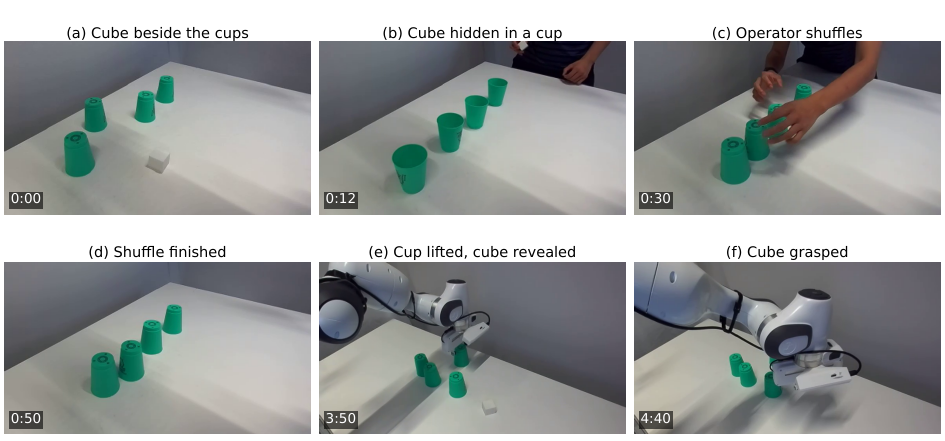}
    \caption{\textbf{Find the cube, trial 9: adapting to a new way of hiding the cube.} (a)--(d)
    Left exterior camera: the operator drops the cube into an upright cup, turns the cup over, and
    shuffles the four cups. (e)--(f) Right exterior camera: after the agent added a detector for this
    case, the robot sets the identified cup aside and grasps the cube. The identification comes from
    the updated detector run on the recorded wrist stream, and the run resumed from locating the
    cups. Time is from the first frame.}
    \label{fig:real-cube}
\end{figure}

\begin{figure}[htbp]
    \centering
    \includegraphics[width=\linewidth]{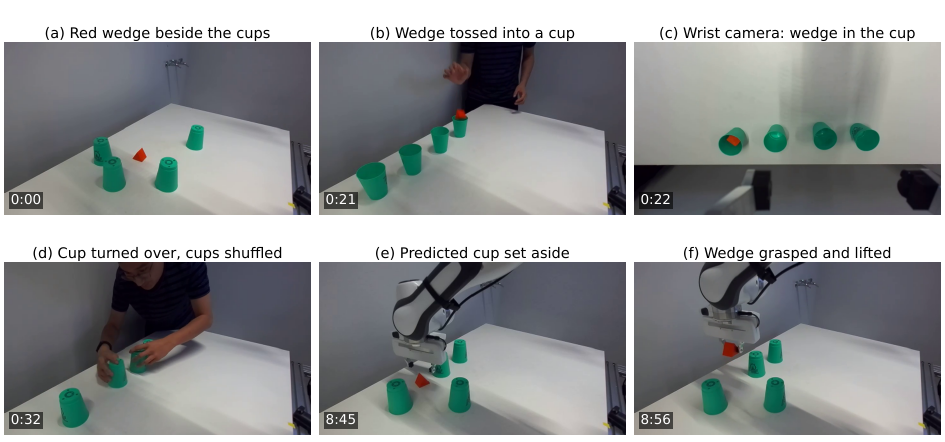}
    \caption{\textbf{Find the cube, trial 11: a red wedge instead of the cube.} (a)--(d) The operator
    tosses the wedge into an upright cup, turns the cup over, and shuffles; (c) is the wrist camera,
    which the agent's detector watches. (e)--(f) On the resumed run, the robot sets the predicted cup
    aside and grasps the wedge on the first attempt. Left exterior camera except (c); time is from the
    first frame.}
    \label{fig:real-cube-wedge}
\end{figure}

\begin{figure}[htbp]
    \centering
    \includegraphics[width=\linewidth]{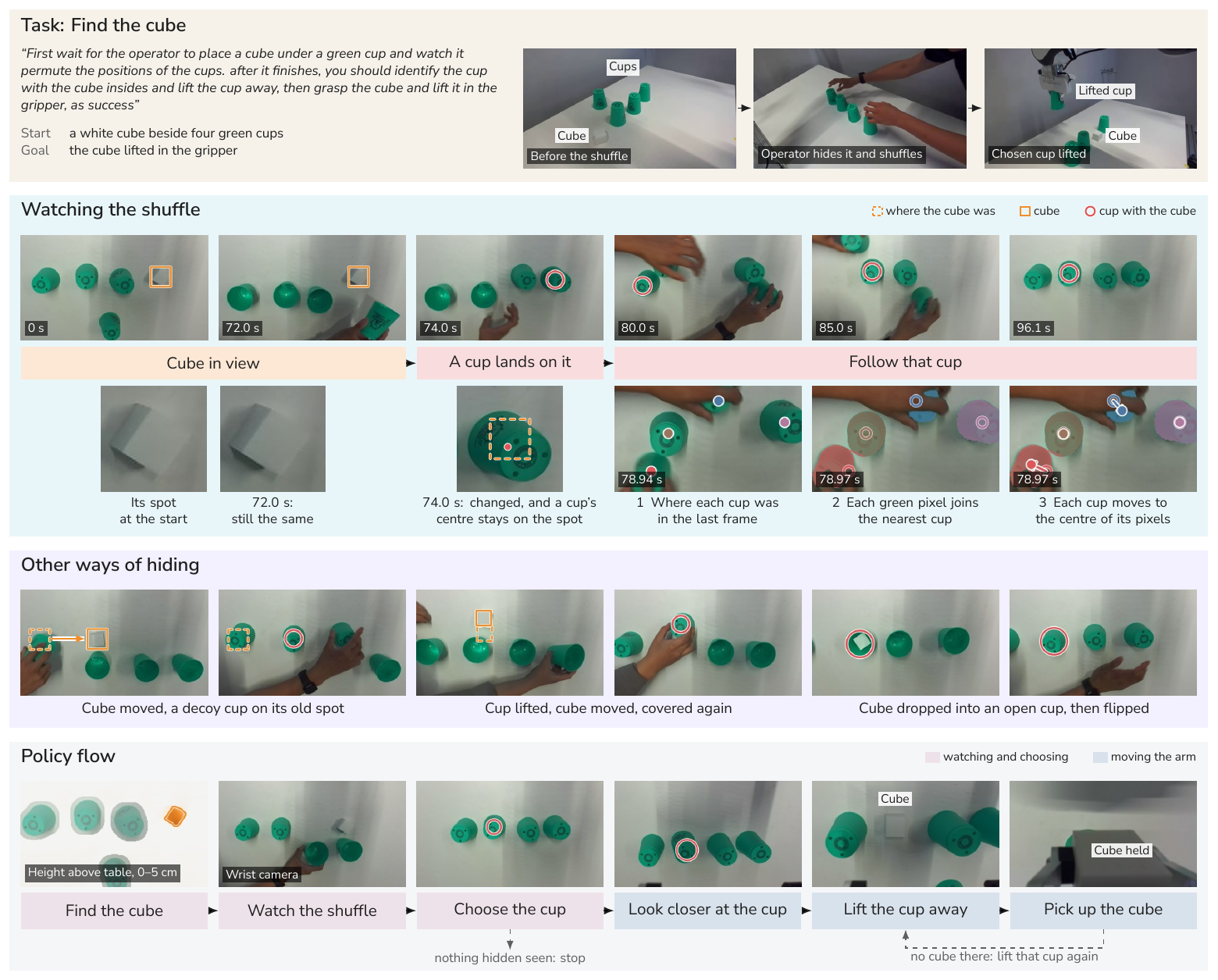}
    \caption{\textbf{The find-the-cube program.} During the shuffle it follows every cup on the wrist camera
    and remembers the cup that covered the cube (top); it also catches a cube that is moved or dropped into
    a cup (middle). It then lifts the cup that hides the cube and picks the cube up (bottom).}
    \label{fig:real-cube-program}
\end{figure}

\paragraph{Trials and outcome.} The agent ran 12 trials between 20:08 and 22:35, and recorded each
in its notes; trials that stopped partway were resumed from the step that failed. In the first
trial the shuffle started before the arm reached the watch pose. In the next two the agent
finished parts by hand-issued commands, and in the third its tracker assumed a static cube and
chose the decoy. Trials 4, 5, and 6 were three consecutive end-to-end successes of the agent's
program without intervention, in 166, 231, and 194\,s. Trial 7 missed the moment the cup
covered the cube; trial 8 identified the cup correctly but pinched the cube through the cup wall
and carried both, after which the agent grasped cups higher and with less force. Trial 9 introduced a
new way of hiding the cube: dropped into an upright cup that was then turned over. The program
recognized no hiding event it knew and stopped without moving the robot. The agent adapted on the
spot: it added a detector for a cube lying inside an upright cup, checked it on the recorded stream
and on all seven recordings so far, and resumed the same scene one minute later, when the robot
lifted the predicted cup and grasped the cube in 98\,s (\Cref{fig:real-cube}). \Cref{fig:real-cube-program} shows how its program tracks the cups. With the red wedge, the cup was identified correctly in trials 10 and 11;
the wedge was grasped after several fixes and manual help in trial 10, and on the first attempt in
the resumed trial 11 (\Cref{fig:real-cube-wedge}). For the wedge the agent kept the cup tracker and
20 of the program's 41 functions unchanged, generalized its model of the hidden object to a
measured size, color, and height, and changed the grasp to close on the wedge along two axes. Success was judged from the gripper width after the grasp and from the
wrist camera's view of the object in the gripper.

\clearpage
\subsection{Air Fryer}
\label{app:real-airfry}

\paragraph{Task.} The robot must open the drawer of an air fryer by its handle, put a banana into
the basket, and close the drawer. The task text was: \emph{``You should open the air fry pan, put
the banana inside, and close it. Reminder: the plate is really heavy and you should figure out
where to grasp and the force applied. Also if the position of the air fry pan is out of range you
may ask for operator reset. Be careful about the forces and maybe not pull out the plate entirely so
we do not have to lift it in everytime.''} The agent's perception, its articulated model of the
fryer, and its verification are described in \Cref{app:perception}.

\paragraph{Operator.} The operator reset the scene between runs; after the first run, for example, the operator took the
banana out of the basket, which the robot cannot reach into. Between runs the fryer was moved (by
about 6\,cm and 4\,cm in the two table directions before run 7) and rotated it (so that the handle
faced the robot's base, before run 8). Before the final run the operator moved the banana away
from the left camera at the agent's request and added an orange ball as a second object.

\begin{figure}[htbp]
    \centering
    \includegraphics[width=\linewidth]{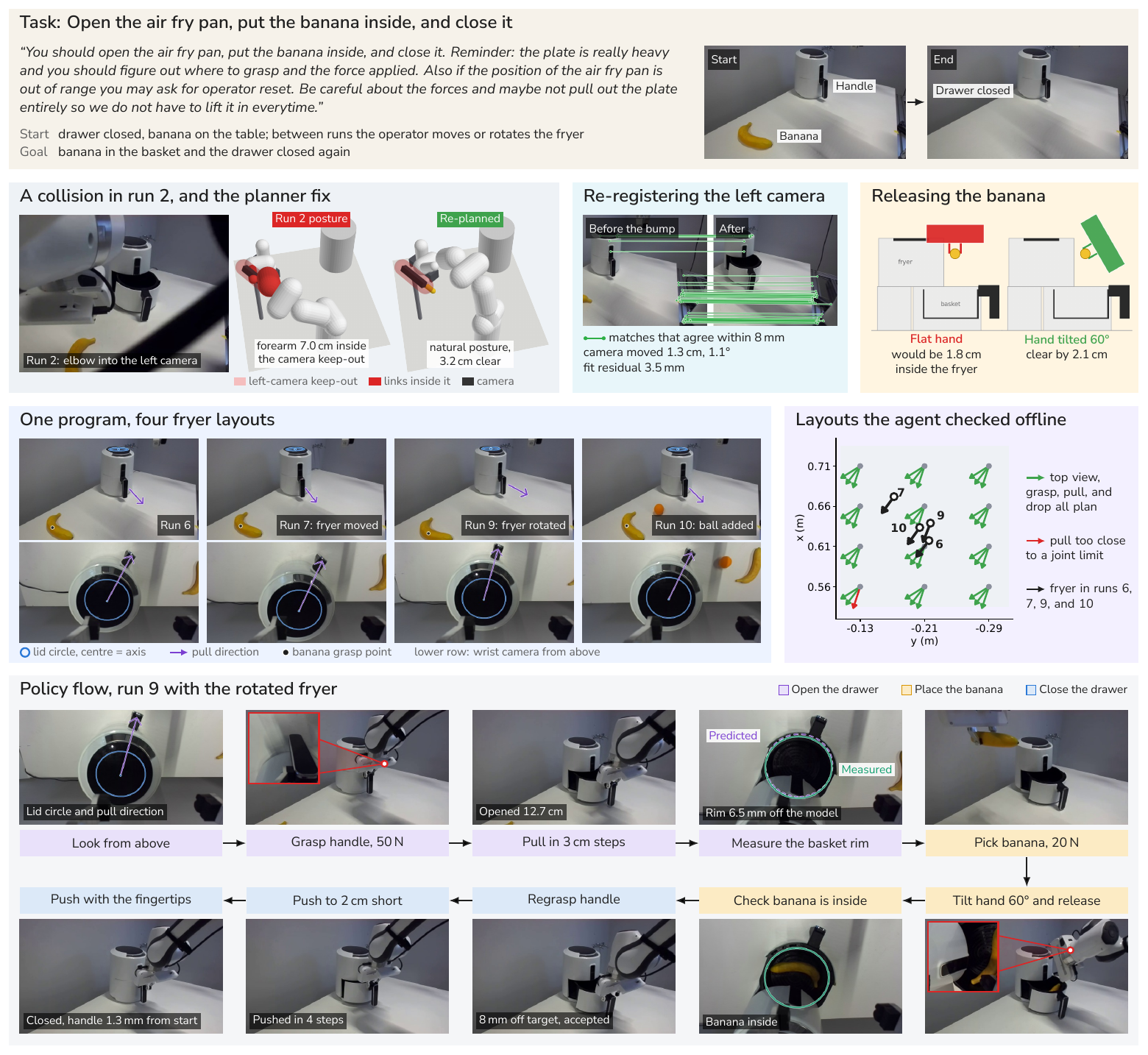}
    \caption{\textbf{The air-fryer program.} After its elbow hit the left camera in run 2, the agent
    re-registered the camera and planned only postures clear of both cameras; it releases the banana with
    the hand tilted 60$^\circ$. One program handles four fryer layouts; the bottom row follows the run with
    the fryer rotated.}
    \label{fig:real-airfry-program}
\end{figure}

\paragraph{Runs and outcome.} The session lasted from 09:18 to 12:05. The agent first completed the
task once by direct tool calls (run 1, about 75 minutes including exploration), then wrote a
program and ran it nine times (runs 2--10). Run 2 ended with a safety stop when the elbow hit the
left camera; the collision moved the right camera, which the agent stopped using. Runs 4, 5, and 8
needed resumes after a false-negative check, a descent timeout, and refused long moves, and run 8
also needed recovery commands issued by the agent. Runs 6, 7, 9, and 10 were clean unattended
successes, including run 7 on the moved fryer and run 9 on the rotated one. In the final run the
program opened the drawer by 13.1\,cm, placed the banana and then the ball in the basket, and
closed the drawer, in 210\,s, leaving the handle 3.5\,mm from its initial position;
\Cref{fig:airfry-stills} shows it. \Cref{fig:real-airfry-program} shows how its program handles the fryer's layout, the camera bump, and the release. Success was judged by the agent's own measurements: the drawer
opening, the objects' points inside the basket rim, and the handle's final position.

\subsubsection{Agent-Written 3D Perception}
\label{app:perception}

This section details the perception code behind \Cref{fig:perception}. The task is to open the
drawer of an air fryer by its handle, put a banana into the basket, and close the drawer, on a
real Franka FR3 with a left exterior and a wrist RGB-D camera. The agent wrote the perception
and the program itself, in about 1,500 lines of Python; they use no learned detector and no
vision-language model. All numbers below are from the agent's final run.

\paragraph{Lifting to a metric frame.} Every depth pixel is back-projected with the camera
intrinsics and the camera-to-base pose into the robot's base frame, and the views of the
exterior and wrist cameras are fused in that frame. All later computations operate on these
points; estimates are projected back into the images only to check them.

\paragraph{Segmentation and geometric primitives.} Each part of the scene is selected by a
colour threshold combined with a height band and a workspace region: the dark lid on top of the
fryer, its white body, the dark handle outside the body, the yellow banana, and an orange ball
just above the table. Geometric primitives are then fitted to the selected points, each with a
plausibility check on its size:
\begin{itemize}
    \item the fryer body, by a robust circle fit to its side (radius 8 to 16\,cm accepted), and
    the lid, seen from the wrist camera above, by a second circle (4.5 to 8\,cm), whose centre
    gives the fryer's vertical axis $c$;
    \item the handle, by the direction from the axis to its horizontal arm, which gives the pull
    direction $d$, the outer face of its grip in 1.5\,cm height slices, the height of its arm,
    and its lateral offset;
    \item the banana, by its principal axis and a quadratic centre line, which give its arc
    middle, the normal on its convex side, its chord (13 to 21\,cm accepted; 18.0\,cm here), and
    its height;
    \item the basket, once the drawer is open, by a circle fitted to its rim.
\end{itemize}
The pull direction from the top view must agree with the one from the side view within
12$^\circ$; the side view alone was biased by about $4^\circ$, which is why the agent added the
top view.

\paragraph{Task frame and articulated model.} All targets are expressed in a frame anchored on
the fryer: the axis $c$, the pull direction $d$, and $t = z \times d$. The agent also wrote an
articulated model of the fryer: a fixed upper cylinder and the back of the lower body, and a
drawer cylinder with box models of the handle's grip and arm, which translate together by the
drawer's opening along $d$. The model provides a signed distance from any point, and the
program instantiates it at the drawer's current opening to check every planned motion for
clearance between the arm, hand, fingers, wrist camera, and held object, and the fryer, the
table, and keep-out regions around the exterior cameras.

\paragraph{How the agent built the model.} No model of the fryer was provided. The agent first
completed the task once by hand, controlling the robot step by step, and recorded what it measured
in its notes: the fryer's top at a height of 0.415\,m and the seam between body and drawer at
0.245\,m, a handle with a horizontal arm and a hanging grip, and the basket centre relative to the
handle once the drawer was out. It also noted that the drawer slides freely along the pull
direction and tips slightly when released. About 90 minutes into the session it turned these
observations into the model above, in the first version of its program, and revised it in later
versions. \Cref{tab:fryer-model} separates what the agent estimates in each new scene from what
it fixed from these measurements.

\begin{table}[htb]
    \centering
    \small
    \caption{\textbf{What the agent's fryer model takes from each scene.}}
    \label{tab:fryer-model}
    \begin{tabularx}{\linewidth}{@{}>{\raggedright\arraybackslash}p{4.6cm}>{\raggedright\arraybackslash}p{2.6cm}X@{}}
        \toprule
        \textbf{Part of the model} & \textbf{In each scene} & \textbf{Source} \\
        \midrule
        Position and orientation (axis $c$, pull direction $d$) & estimated & lid circle and handle direction in the point cloud \\
        Handle face, arm height, lateral offset & estimated & handle points in the point cloud \\
        Drawer opening (joint state) & measured & shift of the handle face after the pull \\
        Radii, heights, handle dimensions & fixed & the agent's measurements from its first manual run \\
        \bottomrule
    \end{tabularx}
\end{table}

\begin{figure}[htbp]
    \centering
    \includegraphics[width=\linewidth]{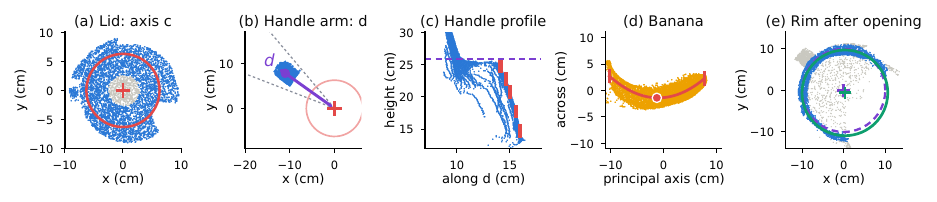}
    \caption{\textbf{The agent's estimators on its final run}, computed by its own code. (a) Lid
    points from the wrist camera above the fryer; blue points are kept and grey points discarded by
    the trimmed least-squares circle fit (red), whose centre is the axis $c$. (b) Handle-arm points
    seen from above, around the lid circle; points outside the $\pm 0.25$\,rad window (dashed) are
    discarded, and $d$ points to the median of the rest. (c) Handle points in the plane along $d$:
    the outer face in each 1.5\,cm height slice (red bars) and the top of the arm (dashed). (d)
    Banana points along their principal axis, with the fitted centre line, its ends, and its middle
    (the grasp point). (e) Rim points after opening around the model's prediction (dashed), and the
    fitted rim (green), 6.5\,mm apart. Distances are in centimetres.}
    \label{fig:fryer-fitting}
\end{figure}

\paragraph{Estimators.} The agent does not register the whole model to the point cloud at once.
Instead, each parameter of the model comes from a small, robust estimate on the points of one
part of the scene, illustrated on its final run in \Cref{fig:fryer-fitting}:
\begin{enumerate}
    \item \textbf{Axis $c$} (\Cref{fig:fryer-fitting}a). From the wrist camera above the fryer, the
    dark points at the height of the lid are fitted with a circle by least squares: writing the
    circle as $x^2 + y^2 = 2ax + 2by + k$ turns the fit into a linear system for the centre
    $(a, b)$ and $k$, with radius $\sqrt{k + a^2 + b^2}$. The fit is repeated six times, each time
    discarding the points that lie farthest from the current circle, so that the dark inner part
    of the lid and stray points do not pull the centre away. The centre is the fryer's axis; the
    radius, 6.3\,cm here, must lie between 4.5 and 8\,cm.
    \item \textbf{Pull direction $d$} (\Cref{fig:fryer-fitting}b). Dark points at the height of
    the handle and just outside the lid are the handle's arm seen from above. Points more than
    0.25\,rad from their median bearing around the axis are dropped, and $d$ points from the axis
    to the median of the rest; a median ignores the few points of other dark objects.
    \item \textbf{Handle face and arm height} (\Cref{fig:fryer-fitting}c). In the plane through the
    axis along $d$, the handle's points are cut into 1.5\,cm height slices. In each slice, the
    97th percentile of the distance along $d$ marks the handle's outer face, the surface the
    fingers and the closing push rest on, while ignoring a few noisy points beyond it; the 95th
    percentile of height gives the top of the arm, which sets the grasp height.
    \item \textbf{Banana} (\Cref{fig:fryer-fitting}d). The banana's yellow points are expressed
    along their principal axis, found by singular value decomposition, and a quadratic curve
    fitted by least squares gives its centre line. The middle of that curve is the grasp point,
    its bulge gives the convex side that sets the hand's orientation, and its ends give the chord,
    18.0\,cm here, which must lie between 13 and 21\,cm.
    \item \textbf{Basket rim} (\Cref{fig:fryer-fitting}e). After opening, the dark rim points are
    searched for around the rim predicted by the model and fitted with a circle in the same way as
    the lid. The measured centre lies 6.5\,mm from the prediction, and the release poses use it.
\end{enumerate}
The drawer's opening is the shift of the handle face between the measurements before and after
the pull, and the camera pose after the bump is a rigid transform fitted by least squares to
feature matches that agree with each other within 8\,mm.

\begin{figure}[p]
    \centering
    \includegraphics[width=\linewidth]{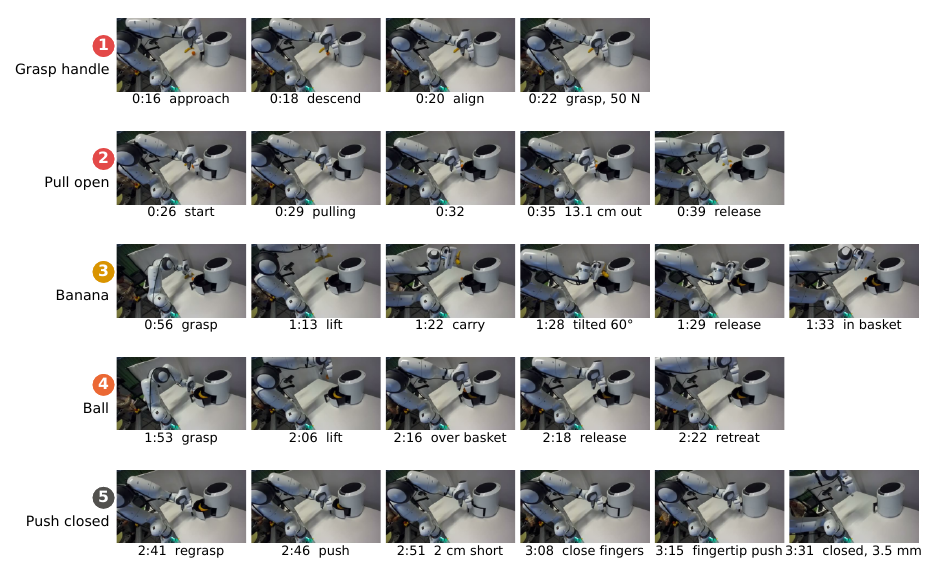}
    \caption{\textbf{The air fryer task on the real robot.} The agent's final run, from the
    continuous recording of an exterior camera, with the time since the start of the run under each
    frame. Each row is one planned action, numbered as in \Cref{fig:perception}(b): grasping the
    handle, pulling the drawer open by 13.1\,cm in 3\,cm steps, carrying the banana and releasing it
    with the hand tilted $60^\circ$ over the basket, releasing the ball, and closing the drawer by a
    push that stops about 2\,cm short followed by a push with closed fingers, which leaves the
    handle within 3.5\,mm of its initial position. This camera was moved by a collision earlier in
    the session and was not used for perception.}
    \label{fig:airfry-stills}
\end{figure}

\paragraph{Planning on the model.} \Cref{fig:airfry-stills} shows each planned action on the real
robot, numbered as in \Cref{fig:perception}(b). The handle grasp lies 1.2\,cm inside the measured grip face
along $d$ and 9\,mm below the top of the arm, with the hand tilted $5^\circ$ toward the fryer.
The drawer is pulled 12.5\,cm along $d$ in absolute 3\,cm steps at a fixed height, since relative
steps sagged under the drawer's weight. The basket centre is predicted from the handle face and
the opening, with an offset measured in the first manual run; the banana is released with its
middle 1.5\,cm beyond the basket centre and the hand tilted $60^\circ$ about the banana's axis, and
the ball slightly further along $d$. The drawer is closed by pushing along $-d$, finishing with
the fingertips on the grip face.

\paragraph{Verification.} Each step is checked by a new measurement. The opening is the shift
of the handle face (13.1\,cm here), and the step fails if it falls more than 3\,cm short of the
command. The rim is searched for around the model's prediction and found 6.5\,mm from it, and the
release poses use the measured rim. The banana counts as placed when its points lie inside the
rim and below its height: 99.99\% of them, with its top at 17.9\,cm against a rim at 23.5\,cm.
Before closing, the handle is measured again, since the released drawer tips by about 2.5\,cm and
shifts sideways by about 1.5\,cm; after closing, its face lies 3.5\,mm from its initial position.

\paragraph{Sensor limits and calibration.} The agent recorded two limits of its sensors and
coded around them: the wrist camera's depth is unreliable closer than about 30\,cm, so metric
targets come from the exterior camera; and side-view pixels on the edge of the banana take the
depth of the rim behind it, so heights are measured from the top-down wrist view. When the arm
bumped the left camera, whose reported pose stayed unchanged, the agent re-estimated that pose
by matching image features against frames captured before the bump, lifting the matches to 3D
with both depth maps, and fitting a rigid transform, with a residual of 3.5\,mm and fixed
landmarks within 2 to 9\,mm of their references. It applies the correction only to frames
captured after the bump, and it stopped using the right camera, which had moved too far to
correct.

\clearpage
\subsection{Wiping a Stain}
\label{app:real-wipe}

\begin{figure}[htbp]
    \centering
    \includegraphics[width=\linewidth]{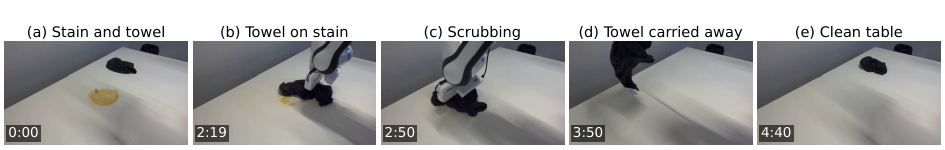}
    \caption{\textbf{Wiping a stain, trial 7b,} from the left exterior camera. (a) The poured stain
    and the towel heap; (b) the grasped wad pressed on the stain; (c) scrubbing; (d) the towel is
    carried back; (e) the table after one pass. Time is from the first frame.}
    \label{fig:real-wipe}
\end{figure}

\begin{figure}[htbp]
    \centering
    \includegraphics[width=\linewidth]{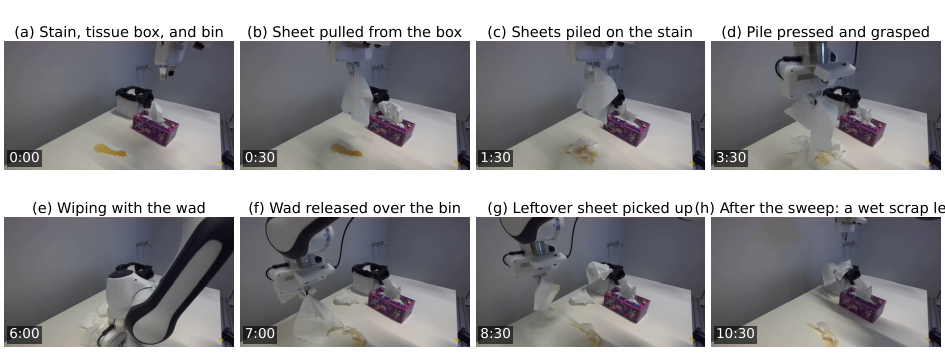}
    \caption{\textbf{Wiping with paper towels, second wet run,} from the left exterior camera. (a) The
    poured stain, the tissue box, and the bin behind it; (b) a sheet pulled from the box; (c) six sheets
    piled on the stain; (d) the pile pressed and grasped as a wad; (e) wiping, with the towel program's
    zig-zag; (f) the wad released over the bin; (g) a leftover sheet picked up by the sweep; (h) the
    table afterwards, with a flat wet scrap the sweep missed. Time is from the first frame.}
    \label{fig:real-wipe-paper}
\end{figure}

\begin{figure}[htbp]
    \centering
    \includegraphics[width=\linewidth]{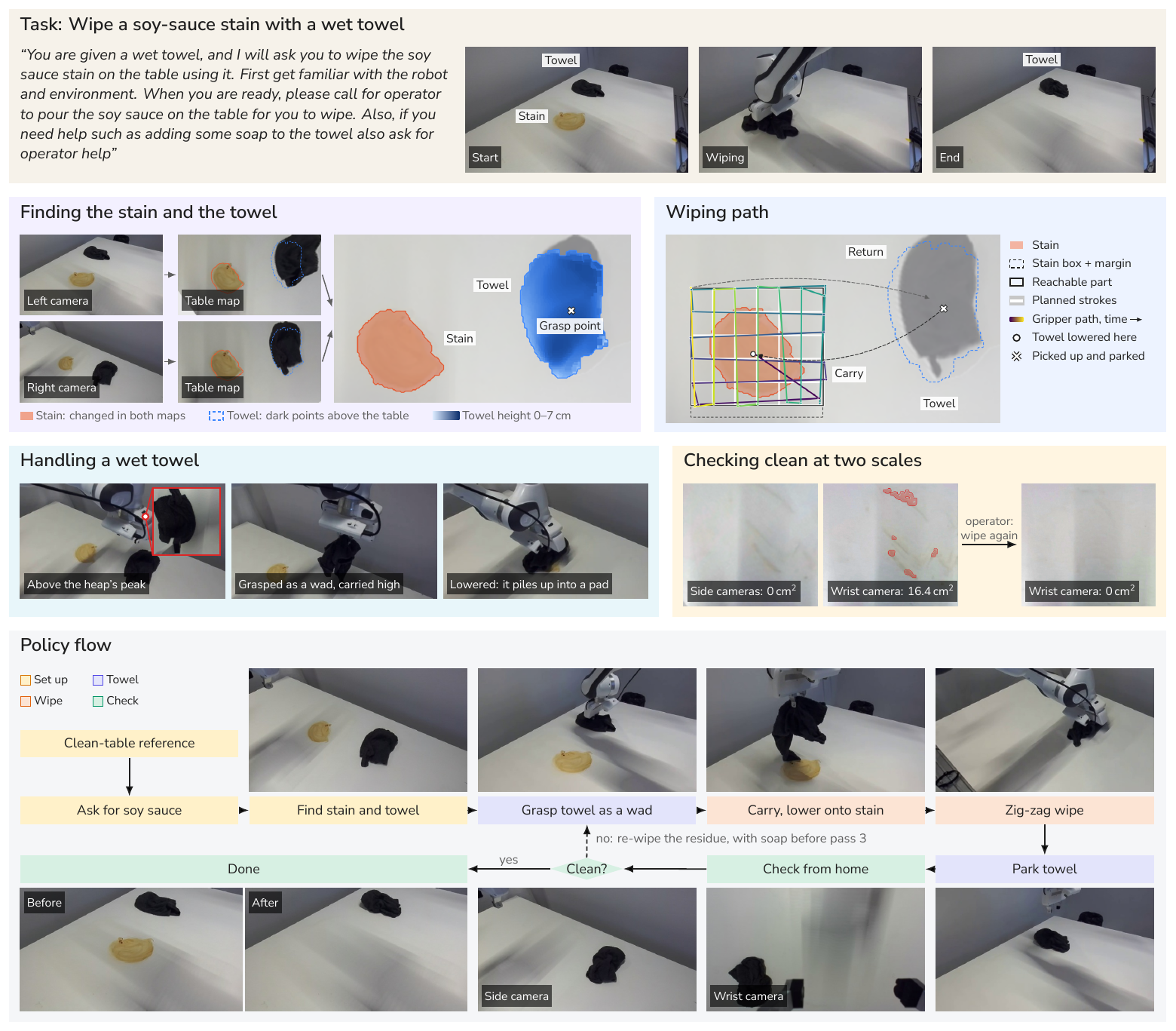}
    \caption{\textbf{The wiping program.} It finds the stain by comparing both exterior cameras with a clean
    reference and the towel heap in depth, grasps the heap as a wad, and wipes a zig-zag fitted to the
    reachable area. The wrist camera catches thin streaks that the exterior cameras miss, so the table is
    checked at both scales.}
    \label{fig:real-wipe-program}
\end{figure}

\paragraph{Task.} The robot must grasp a wet towel and wipe a soy-sauce stain off the table. The
task text was: \emph{``You are given a wet towel, and I will ask you to wipe the soy sauce stain on
the table using it. First get familiar with the robot and environment. When you are ready, please
call for operator to pour the soy sauce on the table for you to wipe. Also, if you need help such as
adding some soap to the towel also ask for operator help.''} The towel is deformable and has no
fixed shape on the table. The agent did not use force sensing: it pressed the towel onto the table
by commanding a height 12\,mm below the contact height it had measured.

\paragraph{Operator.} The operator poured soy sauce when the agent asked, washed the towel and
added soap to it between trials, and returned it to the table as a loose heap. The stains
differed from trial to trial in size (the agent measured 10 to 311\,cm$^2$), shape, and position,
including one next to the left camera at the edge of the reachable workspace.

\paragraph{Trials and outcome.} The session lasted from 09:21 to 12:46. The agent first wiped a
stain by direct tool calls (two passes; the first left a smeared film, and the second, after the
operator added soap, cleaned it), then wrote a program that locates the towel heap, grasps a wad
of it, locates the stain against a clean reference image, scrubs it in a back-and-forth pattern,
parks the towel, and checks for residue. It ran the program in 13 recorded trials and resumes (trials 1 to 7b). Failures
were a refused far reach, an operator reply the program did not expect, one false ``clean''
verdict on thin streaks (the operator asked for a second wipe; the agent then added a check with
the wrist camera at 1\,mm resolution), stains next to the left camera that required fitting the
wiping region to the reachable area, and a refused approach move. The last two trials, 6 and 7b,
cleaned the stain in one pass without intervention; trial 7b reused the scene of trial 7 after the
agent fixed the approach move (\Cref{fig:real-wipe}). \Cref{fig:real-wipe-program} shows how its program finds, wipes, and checks the stain. Success was judged by the agent's residue
measurement against the clean reference in the exterior and wrist cameras (0\,cm$^2$ in both for
trials 6 and 7b) and by eye in the three camera views. 

\paragraph{Adaptation to paper towels.} Afterwards the operator asked for a variant: \emph{``pull
paper towels out of the box, wipe table, and then throw them into the waste bin.''} The agent wrote
a new program that imports the towel program unchanged and calls its parts: locating the stain
against the clean reference, fitting the wiping region to the reachable area, the press-then-grasp
of a wad, the zig-zag wipe 12\,mm below the measured contact, and the residue check. The new code
handles the paper: it finds the box and the bin, pinches the top of a protruding sheet and pulls it
out, piles six sheets on the stain, releases the wad over the bin and scrapes paper stuck to the
fingers, and sweeps up sheets left on the table. After three dry runs, the stain was removed in both
wet runs (\Cref{fig:real-wipe-paper}), but leftover sheets, and in the second run a flat wet scrap
that the sweep missed, were cleaned up with supervised scripts.

\clearpage
\subsection{Peg Insertion: the ManipulationNet Peg-in-Hole Challenge}
\label{app:real-peg}

\begin{figure}[htbp]
    \centering
    \includegraphics[width=\linewidth]{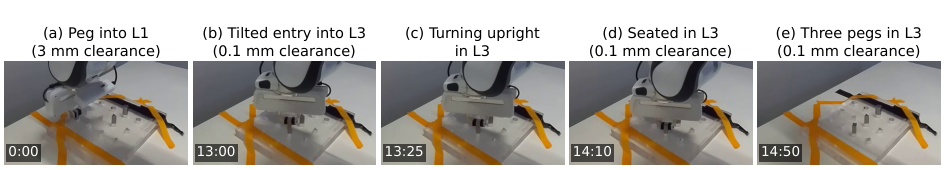}
    \caption{\textbf{The ManipulationNet peg-in-hole challenge,} final run, from a crop of the left
    exterior camera. (a) A peg inserted into L1 (3\,mm clearance); (b)--(d) the rectangle peg
    entering L3 (0.1\,mm clearance) corner first, turned upright, and seated; (e) the three pegs seated in L3.
    Time is from the first frame.}
    \label{fig:real-peg}
\end{figure}

\begin{figure}[htbp]
    \centering
    \includegraphics[width=\linewidth]{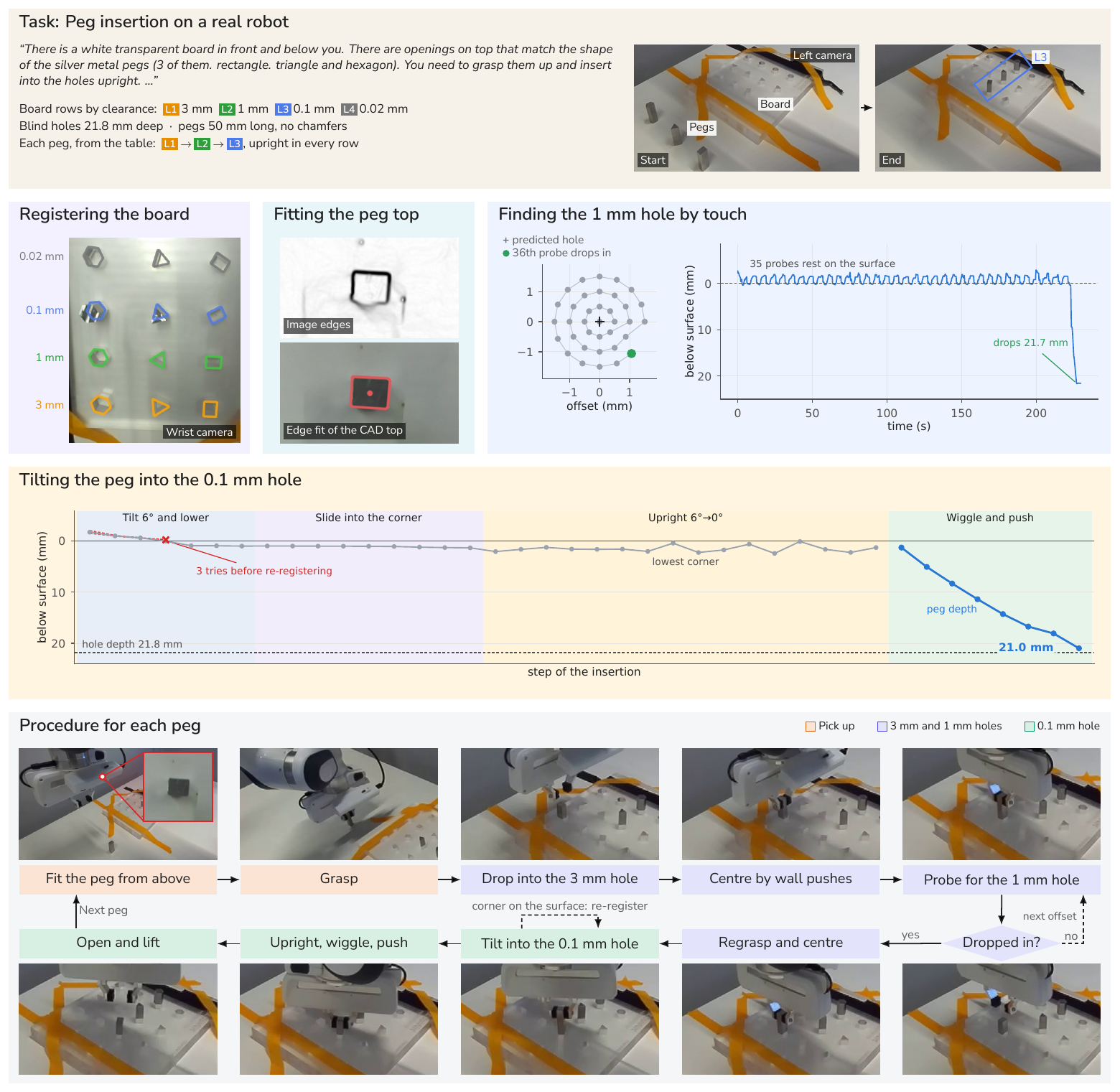}
    \caption{\textbf{The peg-insertion procedure.} The agent fits the board's CAD model and each peg's top
    outline to the wrist camera's image edges. It drops the peg into the 3\,mm row, probes by touch for the
    1\,mm hole, and tilts a corner into the 0.1\,mm hole before turning the peg upright and pushing it down.}
    \label{fig:real-peg-program}
\end{figure}

\paragraph{Task.} This is the peg-in-hole challenge of ManipulationNet~\citep{chen2026manipulationnet}.
Its board is a transparent acrylic plate of $320\times250\times25$\,mm with blind holes 21.8\,mm
deep for five peg shapes, each at four clearance levels, rows L1 to L4 with 3, 1, 0.1, and 0.02\,mm clearance.
Neither the steel pegs, 50\,mm long, nor the holes are chamfered. We used three pegs, the
rectangle, the triangle, and the hexagon. The task text was: \emph{``There is a white transparent
board in front and below you. There are openings on top that match the shape of the silver metal
pegs (3 of them. rectangle. triangle and hexagon). You need to grasp them up and insert into the
holes upright. Reminder: the depth of the wrist is not accurate at close distance, so you can locate
approximately and probe the position and yaw based on force feedback, to succeed in the fine
insertion. If you need help or if the position is of bad IK you can ask for operator to move it.''}
The agent converted the public CAD files of the board and the pegs into a
model of the hole positions, which it registered to the wrist camera's image. During the session
the user asked it to work toward the tighter rows, from L2 to L3. L4 was not attempted.

\paragraph{Operator.} The board rests on the table unfixed. After the agent dragged it while lifting
a jammed peg with a closed gripper, the operator taped the board to the table (at 01:30). The
operator re-seated a hexagon peg that had toppled in L2 after a failed L3 attempt. The L3 fits are
tight enough that the robot could not pull a seated peg out (15\,N with rocking); before the final
run the operator pulled the three pegs out of L3 and stood them upright on the table. The user also
gave advice in the chat, for example to lower the peg further before turning it upright.

\paragraph{Exploration and outcome.} The agent worked by direct tool calls and its own scripts,
supervised by the user; it never ran an unattended program in this task. Between 23:13 and 01:46
it inserted each peg into L1, L2, and L3 in turn (the triangle and the hexagon each failed once in L3
before succeeding). The user then asked for a full run, and between 02:28 and
03:22 the agent took each peg from the table and inserted it into L1, then L2, then L3, showing it
upright in each row (\Cref{fig:real-peg}). \Cref{fig:real-peg-program} shows how it registers the board, finds the holes by touch, and tilts the peg in. L3 needed a tilted entry with a corner of the peg first,
followed by small rocking pushes while the peg was turned upright. All three pegs were seated in L3
at the end, at depths of 20.8\,mm (hexagon), 19.7\,mm (triangle), and 21.0\,mm (rectangle), out of
21.8\,mm. The full run took about 55 minutes including the operator's extraction of the pegs from
the earlier L3 attempts. Success was judged by the insertion depth the agent measured with the arm
and by images of the upright pegs.

\clearpage
\subsection{Bandu Tower}
\label{app:real-bandu}

\begin{figure}[htbp]
    \centering
    \includegraphics[width=\linewidth]{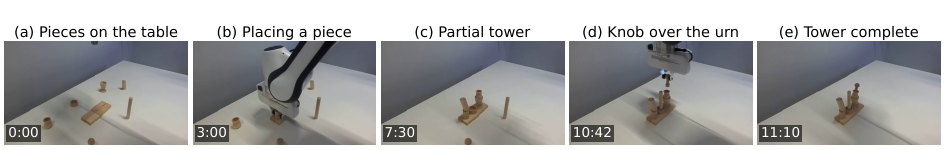}
    \caption{\textbf{Bandu tower, final fully automatic assembly,} from the left exterior camera.
    (a) The pieces after disassembly; (b)--(c) pieces are placed on the base; (d) the knob is held
    over the urn for visual alignment; (e) the completed tower. Time is from the first frame.}
    \label{fig:real-bandu}
\end{figure}

\begin{figure}[htbp]
    \centering
    \includegraphics[width=\linewidth]{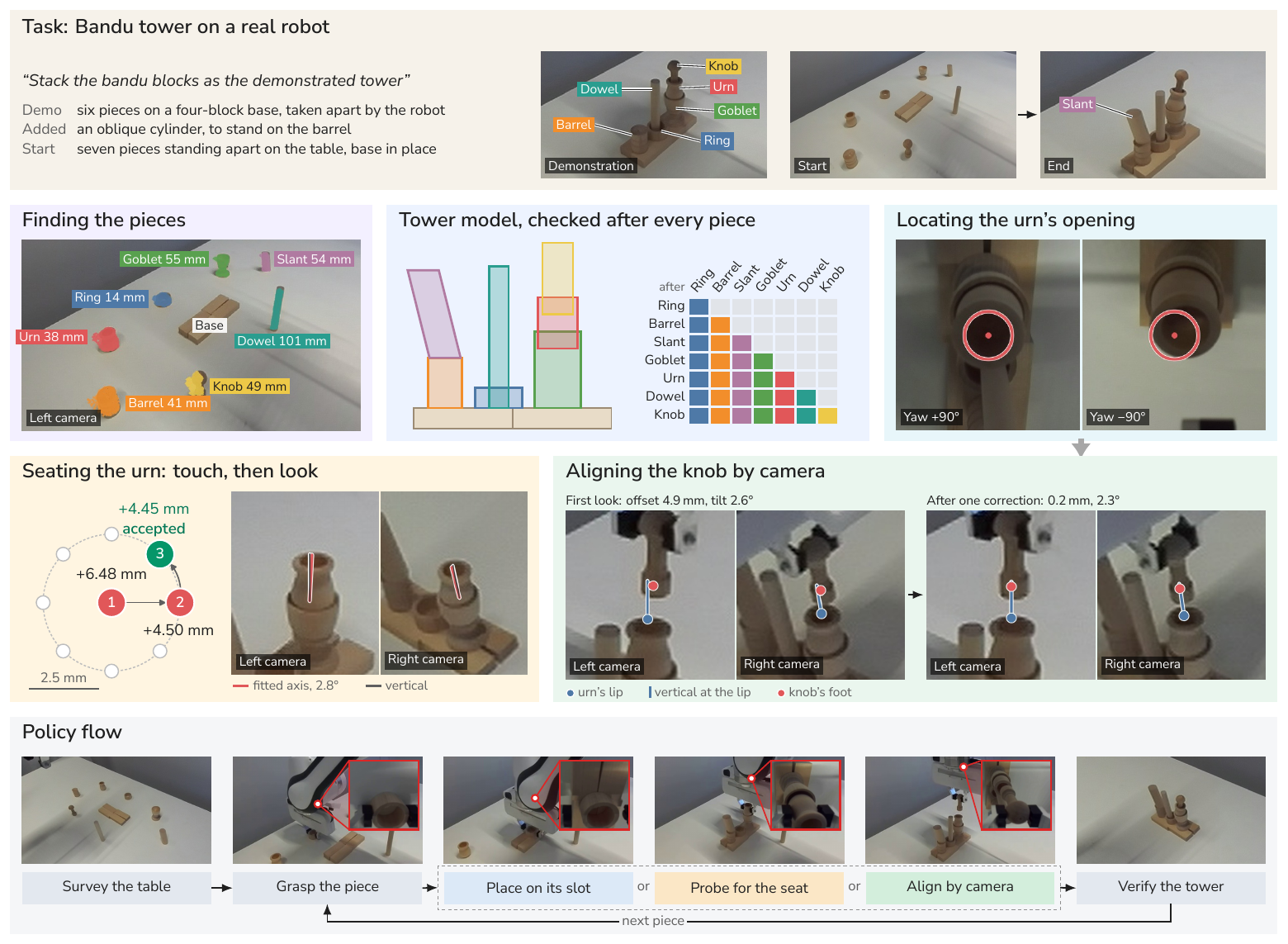}
    \caption{\textbf{The Bandu assembly program,} running fully automatically. It finds the pieces with both
    exterior cameras and Gemini, keeps a slot model of the demonstrated tower, and places each piece by a
    guarded place, a tactile search for the seat (with a tilt check for the urn), or a camera alignment of the
    knob over the urn's opening.}
    \label{fig:real-bandu-program}
\end{figure}

\paragraph{Task.} Bandu is a stacking game with irregular wooden pieces. The task text was a single
sentence: \emph{``Stack the bandu blocks as the demonstrated tower.''} The ``demonstration'' was a
tower standing on the table when the session began; the agent was not shown how it was built. The tower stands on a base of four rectangular blocks laid in a $2\times2$
pattern and consists of seven pieces: a ring with a dowel standing in it, a barrel, a goblet with an
urn seated in its cup and a knob seated in the urn, and an oblique cylinder that the user added
at 13:42, asking for it on top of the barrel. Several of these placements leave little play, and
the knob's opening clearance was about 1\,mm by the agent's estimate. The user told the agent to
take the tower apart and rebuild it, and to repair its own disturbances rather than rely on
human resets.

\paragraph{Operator.} The operator reset the pieces once after the robot's hand swept the tower
(at 15:06, during a sideways pick the user had asked for), restarted or re-homed the gripper twice, and moved
one piece the arm could not reach. The user asked for visual alignment instead of a contact search
for the knob (``align with at most 1\,mm opening error'').

\paragraph{Exploration and outcome.} The session lasted from 11:13 to 17:20. The agent first took the
tower apart and rebuilt it by direct tool calls, then wrote a program with a disassembly and an
assembly mode and alternated them over runs 1 to 15, fixing a failure after each stop. In the
final exploration cycle the program disassembled the tower (7/7 pieces, 459\,s) and then rebuilt
it fully automatically, without intervention: 7/7 pieces in 663\,s, the knob placed by visual
alignment from both exterior cameras to an offset of 0.19\,mm before insertion (\Cref{fig:real-bandu}). \Cref{fig:real-bandu-program} shows how its program perceives, places, and checks each piece. Each
placement was checked by the measured seat height and, for the urn, an uprightness check in both
cameras; the final tower was compared with the demonstrated one in both exterior views.

\clearpage
\subsection{Long-Horizon Tidying on the RB-Y1}
\label{app:real-rby1}

\paragraph{Task.} The RB-Y1 must tidy a lab corner in six chained subtasks given in one
instruction: \emph{``I want you to complete a series of tasks and you should reuse the skills while
still monitor and control the process. First, you should move the chair in front of you out and
park it to the left of the yellow table in front of you to clear it out of the way. then you should
stack the two silver cups together and carry the stacked cups and banana over to the orange basket
to your right and place them in, and pick up the orange basket on to the cabinet with dark top.
Then you should move the chair back and tuck it to the table again. Finally you should pick up the
box on your right and put it also on the cabinet. Be agile and reactive.''} The robot perceives
the scene only through its head camera and must move between the table, the basket, and the cabinet
with its base.

\paragraph{Skills from earlier sessions.} The agent started from a shared skill library written in
five earlier RB-Y1 sessions: carrying a box between the floor and a table (\texttt{bimanual-box-pnp},
from three sessions), lifting the basket onto the cabinet by its rim with both grippers
(\texttt{bimanual-bucket-rim-pnp}), and moving and tucking an office chair (\texttt{office-chair-tuck})
and nesting two cups (\texttt{stack-cups}), both from one chair session. Carrying the cups and the
banana into the basket had no skill. During the session the agent added a skill for reaching hand
moves (\texttt{oc-hand-reach}) and a composition skill for the whole task, and updated the
library after each trial.

\paragraph{Operator.} The operator reset the scene before each trial, stayed in the loop
throughout, and gave advice in the chat, such as ``straighten the arms'', ``lift first'', and
``check the torso and height against previous skills''. In trial 1 the robot's base hit the
cabinet during a hand-planned turn, after which the agent checked the base's swept footprint
before every base move. In trial 3 the operator re-armed the robot after controller faults while the cups
were being nested.

\paragraph{Trials and outcome.} The agent ran the whole sequence three times; all six subtasks
were completed in each trial (trials 1 and 2 overnight on the first day, trial 3 on the
next evening). Trial 3, shown in \Cref{fig:rby1}, ran from 19:31 to 21:09 (98 minutes), with the agent at
every checkpoint of the skills. Success of each subtask was judged by the agent from head-camera
images (for example, two rims visible for nested cups, both objects inside the basket, the basket
and the box on the cabinet).

\clearpage


\section{Compute and Cost}
\label{app:cost}

Across all experiments in this paper, simulated and real, the agents' model calls consumed over
100k US dollars of API credits.

\end{document}